\documentclass[10pt,twocolumn,letterpaper]{article}

\usepackage[pagenumbers]{cvpr} % To force page numbers, e.g. for an arXiv version

\usepackage{indentfirst}

\definecolor{cvprblue}{rgb}{0.21,0.49,0.74}
\usepackage[pagebackref,breaklinks,colorlinks,citecolor=cvprblue]{hyperref}
\usepackage[accsupp]{axessibility}  % Improves PDF readability for those with disabilities.

\definecolor{lightgray}{rgb}{0.85,0.85,0.85}
\definecolor{lightlightgray}{rgb}{0.9,0.9,0.9}
\definecolor{verylightBG}{rgb}{0.9,0.99,0.99}
\definecolor{darkgreen}{rgb}{0., 0.7, 0.2}

\usepackage[utf8]{inputenc} % allow utf-8 input
\usepackage[T1]{fontenc}    % use 8-bit T1 fonts
\usepackage{hyperref}       % hyperlinks
\usepackage{url}            % simple URL typesetting
\usepackage{booktabs}       % professional-quality tables
\usepackage{amsfonts}       % blackboard math symbols
\usepackage{nicefrac}       % compact symbols for 1/2, etc.
\usepackage{microtype}      % microtypography
\usepackage{colortbl}
\usepackage[dvipsnames]{xcolor}
\usepackage{graphicx}
\usepackage{multirow}
\usepackage{caption}
\usepackage{makecell}
\usepackage{arydshln}
\usepackage{array, comment}
\usepackage{nicematrix}
\usepackage{sidecap}

\title{Few-Shot Recognition via Stage-Wise Retrieval-Augmented Finetuning}

\author{
  Tian Liu$^{1}$
  \quad Huixin Zhang$^{1}$ \quad Shubham Parashar$^{1}$ 
  \quad Shu Kong$^{2,3,}\thanks{Corresponding author.}$ \\ 
  {\small 
  $^1$Texas A\&M University \quad $^2$University of Macau \quad $^3$Institute of Collaborative Innovation}  
  \\
  {\small \em website and code: \url{https://tian1327.github.io/SWAT}}
  \vspace{-4mm}
}

\begin{document}
\maketitle

\begin{abstract}
Few-shot recognition (FSR) aims to train a classification model with only a few labeled examples of each concept concerned by a downstream task, where data annotation cost can be prohibitively high. We develop methods to solve FSR by leveraging a pretrained Vision-Language Model (VLM). We particularly explore retrieval-augmented learning (RAL), which retrieves open data, e.g., the VLM's pretraining dataset, to learn models for better serving downstream tasks. RAL has been studied in zero-shot recognition but remains under-explored in FSR. Although applying RAL to FSR may seem straightforward, we observe interesting and novel challenges and opportunities. First, somewhat surprisingly, finetuning a VLM on a large amount of retrieved data underperforms state-of-the-art zero-shot methods. This is due to the imbalanced distribution of retrieved data and its domain gaps with the few-shot examples in the downstream task. Second, more surprisingly, we find that simply finetuning a VLM solely on few-shot examples significantly outperforms previous FSR methods, and finetuning on the mix of retrieved and few-shot data yields even better results. Third, to mitigate the imbalanced distribution and domain gap issues, we propose \textbf{S}tage-\textbf{W}ise retrieval-\textbf{A}ugmented fine\textbf{T}uning (SWAT), which involves end-to-end finetuning on mixed data in the first stage and retraining the classifier on the few-shot data in the second stage. Extensive experiments on nine popular benchmarks demonstrate that SWAT significantly outperforms previous methods by $>$6\% accuracy. 

\end{abstract}

\section{Introduction}
\label{sec:intro}

\begin{figure}[]
  \centering
  \includegraphics[width=0.99\linewidth, clip=true,trim = 0mm 0mm 0mm 0mm]{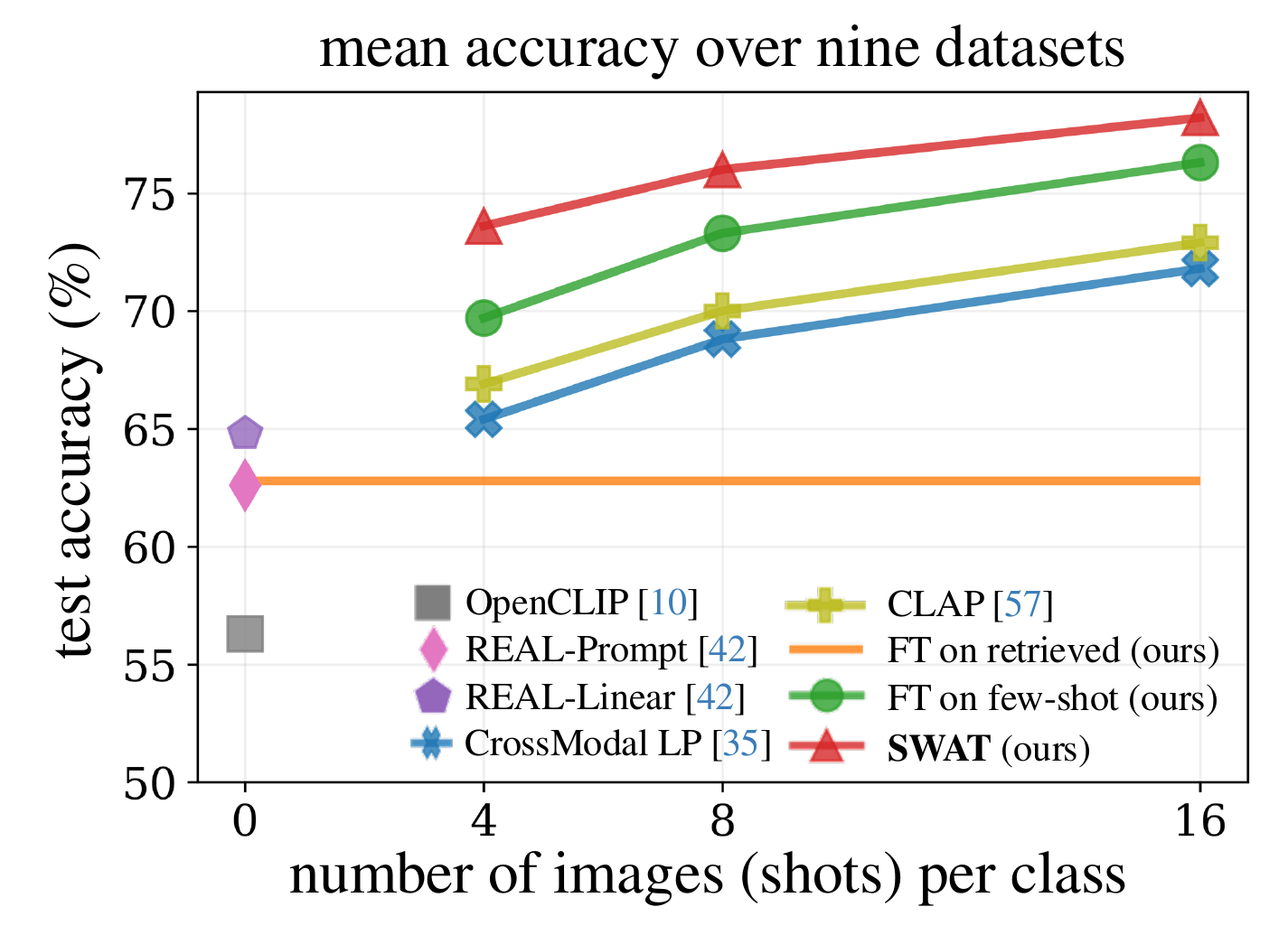}
  % \vspace{-3mm}
  \caption{\small
  {\bf A summary of few-shot recognition (FSR) benchmarking results over nine datasets}.
  Somewhat surprisingly, although underexplored in the literature, finetuning the entire visual encoder solely on few-shot annotated data
  (\textcolor{darkgreen}{green line}) 
  already outperforms previous methods~\cite{lin2023multimodality, clap24} by >3\% accuracy!
  Yet, finetuning only on retrieved data by retrieval augmented learning (RAL, \textcolor{orange}{orange line})
  underperforms the state-of-the-art zero-shot methods~\cite{parashar2024neglected}.
  This is due to that the retrieved data follows an imbalanced distribution and has domain gaps with the few-shot data (Fig.~\ref{fig:domain_gap}).
  By addressing these issues, our {\bf SWAT} performs the best (\textcolor{red}{red line}),  achieving $>$6\% accuracy better than previous methods. Refer to Appendix Fig.~\ref{fig:compare_sota} for detailed results on each of the nine datasets.
  }
  \label{fig:ft_retrieve}
% \vspace{-2mm}
\end{figure}

Few-shot recognition (FSR) aims to train a model with only a few examples
per concept provided by a  downstream task,
where data annotation can be costly.
One motivational application is to train machines for automated data annotation 
by learning from a few examples of each concept provided by \emph{data annotation guidelines}.
Motivated by this data annotation application,
we focus on solving FSR, particularly by leveraging a foundational Vision-Language Model (VLM) and its web-scale pretraining data.

\begin{figure*}[t]
    \centering
    \small
    \includegraphics[width=1\linewidth, clip=true,trim = 0mm 0mm 0mm 0mm]{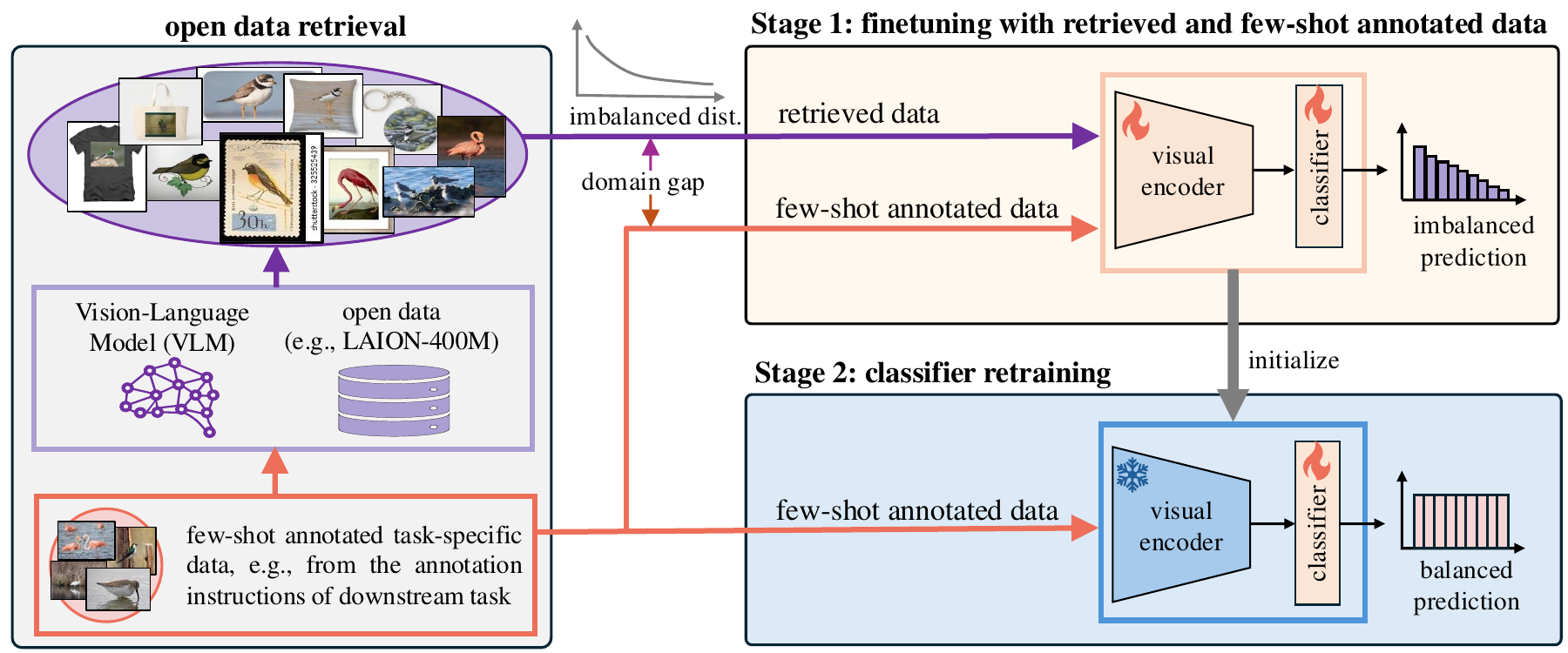}
    % \vspace{-1mm}
    \caption{\small 
    {\bf Overview of our \emph{S}tage-\emph{W}ise retrieval-\emph{A}ugmented fine-\emph{T}uning (\emph{SWAT})  for few-shot recognition (FSR).} 
    Consider the scenario where one wants to train a model on a few examples per concept concerned in data annotation guidelines.
    SWAT exploits a pretrained Vision-Language Model (VLM) and retrieves open data, e.g., the VLM's pretraining data relevant to the concepts of interest.
    We observe that the retrieved data follows an imbalanced distribution and has domain gaps from the few-shot examples (Fig.~\ref{fig:domain_gap}).
    SWAT addresses the two issues jointly by first end-to-end finetuning the VLM's visual encoder on mixed retrieved and few-shot annotated data, then re-training the classifier only using the few-shot examples.
    Over nine FSR benchmarks, our SWAT achieves state-of-the-art performance, significantly outperforming previous methods by $>$6\% accuracy (Fig.~\ref{fig:ft_retrieve}).
    } 
% \vspace{-1mm}
\label{fig:teaser}
\end{figure*}

{\bf Status Quo.}
In the literature,
FSR has served as a proxy task to study parameter-efficient finetuning (PEFT)~\cite{schmidt2009meaning, snell2017prototypical, liu2023learning, wang2018low},
and the robustness and generalization of pretrained models~\cite{maple, zhou2022learning, zhu2023prompt}, etc.
Consequently,
most FSR methods focus on learning a few parameters,
such as a classifier head~\cite{clipadapter, tipadapter, taskres, lin2023multimodality, clap24, tang2024amu} or prompt tokens~\cite{zhou2022learning, chen2022plot} over a frozen pretrained backbone (e.g., VLM's visual encoder).
However, emphasizing learning fewer parameters 
without prioritizing recognition accuracy limits the performance of these methods in real-world applications such as automated data annotation~\cite{madan2024revisiting}.
Importantly, recent works~\cite{lin2023multimodality, clap24} show that the setup for a majority of FSR methods unreasonably assumes access to a much larger validation set for hyperparameter tuning, hindering their practical utility.
Our work \emph{solves} FSR with the realistic motivation of automated data annotation, adopting a rigorous setup without access to an unrealistic large val-set and optimizing for higher accuracy without limiting to learning only a few parameters.

{\bf Motivation.}
Different motivations drive different FSR methods. 
Many existing FSR methods are factually motivated to study PEFT~\cite{schmidt2009meaning, snell2017prototypical, liu2023learning, wang2018low} or learned models' robustness and generalization~\cite{zhou2022learning, zhu2023prompt, maple}. 
This motivation has led to a \emph{flawed} setup, as noted in \cite{lin2023multimodality, clap24}, where an unrealistically large validation set is used for hyperparameter tuning.
In contrast,
we solve FSR motivated by {\bf \em data annotation}~\cite{madan2024revisiting}, 
avoiding large validation sets and prioritizing accuracy over learning a small number of parameters.
This data annotation perspective expands FSR research
beyond the prevailing PEFT~\cite{clipadapter, tipadapter,  lin2023multimodality, clap24, tang2024amu}. 
Interestingly, 
we find that finetuning a pretrained visual encoder on few-shot annotated examples already surpasses prior FSR methods without overfitting issues (Fig.~\ref{fig:ft_retrieve})!

Furthermore,
most FSR methods focus on adapting an open-source pretrained Vision-Language Model (VLM)~\cite{lin2023multimodality, clap24, tang2024amu}.
Differently, we also embrace VLM's pretraining data as \emph{open data}, exploiting retrieval-augmented learning (RAL) \cite{li2023internet, liu2023learning, wallingford2023neural, iscen2023retrieval,  saha2024improved, parashar2024neglected} to address the challenge of limited labeled data.
RAL shines in zero-shot recognition,
where state-of-the-art zero-shot methods adopt a VLM to retrieve relevant pretraining data for the concepts of interest and learn a classification model over such retrieved data~\cite{parashar2024neglected}.
Our work, for the first time, extends RAL to FSR. 
Despite its simplicity, we reveal interesting and novel challenges and opportunities. We elaborate on them below.

{\bf Challenges and Insights.}
Unlike prior FSR methods that learn a small number of parameters, 
we first explore full finetuning of the entire pretrained model (i.e., VLM’s visual encoder) on the few-shot examples.
Somewhat surprisingly, this simple method significantly outperforms previous FSR methods (Fig.~\ref{fig:ft_retrieve}) with quite affordable computation costs
(Table~\ref{tab:compute_cost}), 
owning to the small scale of few-shot data.
In another line of research, state-of-the-art methods of zero-shot recognition (ZSR) adopt a strategy called retrieval-augmented learning (RAL), which retrieves relevant examples from VLM's pretraining set and learns a classifier over them~\cite{parashar2024neglected}.
We test RAL for FSR by only using the retrieved data: both classifier learning (i.e., REAL-Linear~\cite{parashar2024neglected}) and full finetuning barely surpass the non-RAL ZSR method REAL-Prompt~\cite{parashar2024neglected} and underperform other FSR methods~\cite{clap24, lin2023multimodality}, 
even trained with far more retrieved data.
We identify two culprits (Fig.~\ref{fig:domain_gap}): 
(1) imbalanced distribution of retrieved data, and (2) domain gaps between the retrieved data and few-shot examples. 
To address the two issues, we propose a simple method,
\textbf{S}tage-\textbf{W}ise retrieval-\textbf{A}ugmented fine\textbf{T}uning (SWAT, cf. Fig.~\ref{fig:teaser}).
SWAT first finetunes the VLM’s visual encoder on mixed retrieved and few-shot data, then re-trains the classifier solely on few-shot examples.
This stage-wise training of our SWAT aligns with the decoupling learning strategy~\cite{kang2019decoupling}, which addresses the imbalanced distribution issue by first learning feature representations using all imbalanced data and then learning the classifier with balancing techniques.
SWAT also mitigates the domain gap in a transfer learning manner, i.e., pretraining on larger source-domain data followed by finetuning on the target-domain data~\cite{Li_2016_CVPR, Inoue_2018_CVPR, Hsu_2020_WACV}.

{\bf Contributions.}
We make three major contributions:
\begin{enumerate}
    \item    
    To solve few-shot recognition (FSR) for real-world applications, e.g., automated data annotation, we develop FSR methods focusing on better accuracy without 
    restricting the number of learned parameters.
    This broadens FSR research space. 
    For example,
    we find that finetuning a VLM's visual encoder on few-shot examples already outperforms previous FSR methods by $>$3\% (Fig.~\ref{fig:ft_retrieve}).

    \item 
    For the first time, we explore retrieval-augmented (RAL) learning for FSR, a technique well-studied in zero-shot recognition (ZSR). Despite its simple extension from ZSR to FSR, we identify novel and interesting challenges: the retrieved data follows an imbalanced distribution and has domain gaps with the few-shot examples.
    
    \item
    We develop a simple method Stage-Wise retrieval-Augmented fineTuning (SWAT) that effectively mitigates the aforementioned imbalanced distribution issue and domain gaps, resulting in significantly better performance (>6\%) than prior arts on nine benchmarks.
    Fig.~\ref{fig:teaser} illustrates our SWAT and Fig.~\ref{fig:ft_retrieve} summarizes our results.

\end{enumerate}

\section{Related Work}

{\bf Few-Shot Recognition (FSR).}
Early FSR methods leverage metric learning~\cite{snell2017prototypical, vinyals2016matching, bateni2020improved}, meta learning~\cite{finn2017model, ravi2016optimization}, transductive learning ~\cite{zhu2023transductive, joachims1999transductive, boudiaf2020information} and graph neural networks (GNNs)~\cite{gidaris2019generating, satorras2018few}. 
These approaches pretrain a model on a large meta-training set, then finetune it on task-specific few-shot examples, aiming for strong generalization. 
Recent works exploit Vision-Language Models (VLMs) pretrained on web-scale data~\cite{clip, align, alayrac2022flamingo} for FSR.
With a frozen visual encoder, 
these methods harness the extraordinary zero-shot transfer capabilities of VLMs by learning better prompts~\cite{zhou2022learning,zhou2022conditional,chen2022plot, maple, xing2023dual, yao2023visual, zhu2023prompt} or lightweight adapters~\cite{clipadapter, tipadapter, taskres, lin2023multimodality, clap24, tang2024amu, zhang2023prompt}.
While utilizing pretrained VLMs achieve state-of-the-art performance on various benchmark datasets, these studies prioritize parameter-efficient finetuning (PEFT) with frozen backbones, even though finetuning more layers can yield better results~\cite{lin2023multimodality, clap24}.
Furthermore, recent studies~\cite{lin2023multimodality, clap24} point out that these works unrealistically use large validation sets for hyperparameter tuning, limiting their real-world applicability.
This focus on FSR as a proxy for PEFT~\cite{schmidt2009meaning, snell2017prototypical, liu2023learning, wang2018low} or model robustness/generalizability~\cite{maple, zhou2022learning, zhu2023prompt} has shaped the current ``flawed setup'' and their limitations.

Contrasting to these methods,
our work is driven by the real-world application of data annotation,
prioritizing recognition accuracy over PEFT constraints.
We leverage both VLMs and their pretraining data while strictly adhering to a validation-free setting. This realistic setup broadens FSR research. Notably, we find that simply finetuning a VLM’s entire visual encoder on few-shot examples already significantly outperforms previous FSR methods.

{\bf Vision-Language Models (VLMs) and Retrieval-Augmented Learning (RAL).}
VLMs, such as CLIP~\cite{clip} and ALIGN~\cite{align}, are pre-trained on web-scale data sampled from the open world.
They learn to map image and text data into a common embedding space via contrastive learning on millions of image-text pairs. 
VLMs have demonstrated impressive zero-shot transfer capability across various downstream tasks. 
To further improve VLMs' zero-shot accuracy,
recent works~\cite{parashar2024neglected, liu2023learning, wallingford2023neural, iscen2023retrieval, li2023internet} explore RAL,
which retrieves open data (e.g., from the VLM's pretraining dataset) relevant to the downstream task and then uses such data to finetune the VLM or learn a classifier. 
Inspired by RAL's success in achieving state-of-the-art zero-shot performance, 
for the first time, we extend RAL to few-shot recognition.
We identify not only opportunities but also interesting challenges. 
We present a novel FSR method that addresses these challenges, achieving state-of-the-art few-shot recognition performance on standard benchmark datasets.

\begin{figure*}[t]
    \centering
    \small
    \includegraphics[width=0.99\linewidth, clip=true,trim = 5mm 0mm 0mm 0mm]{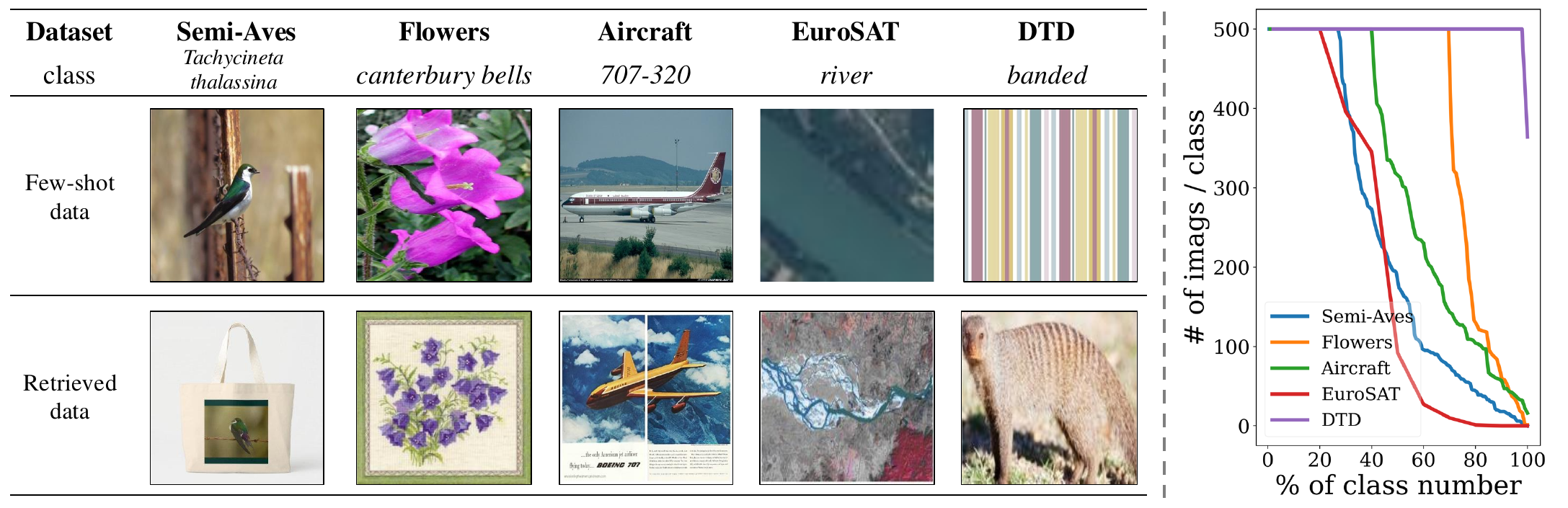}
    % \vspace{-4mm}
    \caption{\small {\bf Retrieved data shows domain gaps with downstream few-shot data and follows an imbalanced distribution.}
    Left: we compare retrieved and few-shot annotated images for random categories from five benchmark datasets. 
    The two sets of images exhibit clear domain gaps regarding image styles, background content, and even semantics, e.g., the animal with banded stripes in the DTD dataset.
    Right: retrieved data follows imbalanced distributions w.r.t concepts defined in different downstream tasks, as the VLM's pretraining set does not contain sufficient examples for certain classes.
    Due to these two issues, leveraging the retrieved data to improve FSR presents significant challenges.
    Refer to Table~\ref{tab:dataset_acc} for quantitative justification of domain gaps, and Appendix Fig.~\ref{fig:imbalanced_all} and \ref{fig:retrived_imgs_more} for additional examples of more datasets.
    }
% \vspace{-3mm}
\label{fig:domain_gap}
\end{figure*}

{\bf Data Issues in the Open World.}
Real open-world data poses various challenges to a machine-learned model.
Two well-known challenges relevant to our work are
\emph{imbalanced distribution} of training data, and {\em domain gaps} between training and testing data.
Imbalanced learning has been extensively studied through long-tailed learning~\cite{zhang2023deep, kang2019decoupling, cao2019learning, yang2020rethinking, alshammari2022long};
recent work \cite{parashar2024neglected} points out that pretrained VLMs also suffer from imbalanced distribution, as pretraining data naturally follows a long-tailed distribution in the real open world.
Indeed, we confirm that the retrieved data also follows an imbalanced distribution w.r.t concepts concerned in a downstream task (cf. Fig.~\ref{fig:domain_gap}).
Among various methods that address the imbalanced or long-tailed distribution~\cite{zhang2023deep}, a simple stage-wise training approach proves to be effective: first training a model using all the imbalanced data, and then retraining the classifier with balancing techniques~\cite{kang2019decoupling}.
Additionally,
retrieved data has a clear domain gap with the few-shot annotated examples (Fig.~\ref{fig:domain_gap}), which impacts model performance when applied across domains.
Transfer learning and finetuning on target-domain data~\cite{guo2019spottune, niu2020decade, neyshabur2020being} can effectively mitigate this gap. 
Our work shows that using VLM and RAL for FSR necessitates addressing these two issues jointly. Our proposed Stage-Wise retrieval-Augmented fineTuning (SWAT) effectively tackles both, significantly enhancing FSR performance.

Moreover, to tackle data scarcity and enhance model performance, 
data augmentation is widely adopted during model training~\cite{shorten2019survey, zhao2021camera, rebuffi2021data}. 
Some augmentation techniques are performed on individual data examples, e.g., random cropping and color jittering~\cite{chen2020simple, kim2020data, volpi2018generalizing},
while others mix multiple examples~\cite{cao2022survey}, e.g., MixUp~\cite{zhang2017mixup} and CutMix~\cite{yun2019cutmix}.
Among various data augmentation techniques,
CutMix proves to stabilize training, reduce overfitting, and improve model performance~\cite{yun2019cutmix}.
Later works extend CutMix by combining it with MixUp~\cite{park2022unified}, smoothing patch boundaries~\cite{park2022unified}, leveraging saliency masks~\cite{uddin2020saliencymix, kim2021co}, and considering long-tailed distribution of training data~\cite{park2022majority}. 
Our approach augments data by mixing retrieved images with few-shot annotated ones, 
yielding substantial improvements in FSR.

\section{Problem Formulation and Methods}
\label{sec:fsr-swat}

Our work focuses on solving few-shot recognition (FSR). We first outline the problem and evaluation setup, then introduce our methods, progressing from finetuning on few-shot data to our final approach, Stage-Wise retrieval-Augmented fineTuning (SWAT).

{\bf Problem Setup.}
Our FSR setup aims for high recognition accuracy on the concepts concerned by a downstream task.
Our setup follows recent studies~\cite{zhang2023prompt, lin2023multimodality, clap24} that leverage Vision-Language Models (VLMs) for FSR. 
This setup also permits the use of VLM pretraining data~\cite{laion5b, laion400m}, available as ``free lunch'', which has been exploited in zero-shot recognition~\cite{parashar2024neglected, liu2023learning, wallingford2023neural, iscen2023retrieval, li2023internet} and is thus applicable to FSR.
To support real-world applications like data annotation, this setup prioritizes recognition accuracy over limiting the number of parameters to learn, contrasting with many FSR works that treat FSR as a proxy for studying model robustness or generalization~\cite{maple, zhou2022learning, zhu2023prompt}, and PEFT methods~\cite{schmidt2009meaning, snell2017prototypical, liu2023learning, wang2018low}.

{\bf Evaluation Setup.}
Our work aims to develop practical FSR solutions for real-world applications, such as learning from few-shot examples in annotation guidelines for automated data annotation.
To support our study, we adopt a rigorous and realistic evaluation setup~\cite{clap24}, which prohibits using a large validation set for hyperparameter tuning.
This differs from most FSR works, which rely on artificially large validation sets for hyperparameter tuning, as noted in recent studies~\cite{clap24, lin2023multimodality}. 
Instead, we set hyperparameters to default values from the literature, including learning rate and weight decay, and follow ~\cite{clap24} to apply this same set of hyperparameters across all benchmark datasets.

{\bf Remarks.}
Previous FSR methods often evaluate the generalization of models in a base-to-novel setup, which artificially splits an existing dataset into 
separate base and novel class sets.
In this setup, models train on few-shot data from the ``base classes'' and are then evaluated on the ``novel classes''. 
However, this base-to-novel setup does not align with our motivation -- the practical application of data annotation,
where the annotation guidelines clearly define the target concepts of interest, without introducing extra novel classes.
More importantly, as noted by \cite{madan2024revisiting}, this base-to-novel setup appears to be artificial as VLMs have already seen examples of these ``novel'' classes in their Internet-scale pretraining dataset. 
Therefore, in this work, we follow \cite{madan2024revisiting} to evaluate FSR methods on the target classes concerned by a downstream task, without a base-novel split.

{
\setlength{\tabcolsep}{0.3em}
\begin{table}[t]
\small
\centering
\caption{\small
We demonstrate the domain gaps between two data sources: retrieved data and few-shot annotated data from
a downstream task.
Following \cite{torralba2011unbiased}, we train a binary classification model on the two data sources to identify 
the data source of each image from a held-out testing set.
We report the binary classification accuracy on the testing sets, which are created specifically for this study based on each downstream dataset.
Results validate the presence of domain gaps with $>$90\% mean accuracy over different downstream-task datasets.
Refer to Fig.~\ref{fig:domain_gap} for a visual demonstration.
}
% \vspace{-1mm}
\scalebox{0.9}{   
\begin{tabular}{ccccccc}
\toprule
dataset & Semi-Aves &Flowers & Aircraft &EuroSAT &DTD & \cellcolor{gray!15}\emph{mean acc.} \\ 
\midrule
accuracy & 86.8 & 90.0 & 94.4 & 98.5 & 81.4 & \cellcolor{gray!15}90.2 \\
\bottomrule
\end{tabular}
}
\label{tab:dataset_acc}
% \vspace{-1mm}
\end{table}
}

\subsection{Finetuning on Few-Shot Data}
\label{ssec:retrieval}

Recall that our FSR setup prioritizes high recognition accuracy for downstream tasks, different from prior works that focus on PEFT~\cite{lin2023multimodality, clap24}.
This motivated us to first test \emph{few-shot finetuning} (FSFT), which finetunes the VLM’s visual encoder only on the few-shot examples.
To the best of our knowledge, this method has not been explored in the literature.
Surprisingly, 
this embarrassingly simple method 
significantly outperforms previous FSR approaches by $>$3\% (Fig.~\ref{fig:ft_retrieve} and Table~\ref{tab:compare_sota})!
Although one may suspect overfitting when finetuning the whole visual encoder on the few-shot examples, our empirical results show that overfitting is not an issue (Table~\ref{tab:compare_sota} and Fig.~\ref{fig:no_overfit}).
This is likely because of the robust hyperparameters derived from other research lines~\cite{parashar2024neglected, lin2023multimodality, wortsman2022robust, kumar2022fine}, 
including classifier initialization using text embedding (Table~\ref{tab:ablate_cls_init}) and a smaller learning rate for updating the visual encoder (Appendix Section~\ref{sec:hyperparams}). 
Moreover, 
owning to the small scale of few-shot data, learning more parameters (i.e., finetuning the entire pretrained visual encoder) is still quite efficient (Table~\ref{tab:compute_cost}).

\subsection{Finetuning on Retrieved Data}
Next, we explore retrieval-augmented learning (RAL) with the VLM and its pretraining data. RAL retrieves pretraining examples relevant to the concepts concerned by the downstream task, which has been extensively studied in zero-shot recognition~\cite{parashar2024neglected, liu2023learning, wallingford2023neural}.
However, to the best of our knowledge, RAL has not been explored for FSR. Below, we detail the approach and the interesting challenges of using RAL for FSR.

{\bf Retrieval Augmentation.}
RAL has proven effective
in zero-shot recognition~\cite{liu2023learning, wallingford2023neural, parashar2024neglected}.
There are various strategies for retrieving open data, especially VLM's pretraining data, relevant to concepts of interest.
Some methods match feature similarities between textual embeddings of target concepts and features of pretraining images or captions~\cite{liu2023learning, wallingford2023neural}, while others use string matching to find relevant text and retrieve corresponding images~\cite{parashar2024neglected}. 
The latter approach is significantly more efficient, as it avoids the high storage costs of downloading all images and the computational expense of embedding calculations for each image.
Importantly, the latter helps achieve better zero-shot recognition and is more efficient than the former~\cite{parashar2024neglected}.
Therefore, we adopt the string matching approach~\cite{parashar2024neglected} to retrieve examples in this paper.

{\bf Novel Challenges.}
With the large amount of retrieved data,
we finetune the visual encoder for classification.
Surprisingly, 
finetuning on retrieved data underperforms the few-shot finetuning method, and performs even worse than the state-of-the-art zero-shot method (Fig.~\ref{fig:ft_retrieve})!
We identify two key culprits contributing to the inferior performance:
(1) domain gaps between the retrieved data and the downstream-task few-shot examples, 
and (2) imbalanced distributions of retrieved data. 
We elaborate on the two issues below and present our solution in the next subsection.

{\em Issue 1: Domain Gap.}
VLMs' pretraining dataset consists of a large number of web-scraped image-text pairs from diverse sources.
These images vary in styles, resolutions, and backgrounds, which are not aligned with the distribution of downstream-task images. 
We compare examples of retrieved images and few-shot annotated images for five image classification tasks 
in Fig.~\ref{fig:domain_gap}.
We further quantitatively verify the domain gaps by following \cite{torralba2011unbiased}. 
Concretely,
we train a binary classification model to distinguish whether a given image originates from the downstream-task data source or the retrieved data source. 
The resulting classifier achieves $>$90\% accuracy (Table~\ref{tab:dataset_acc}), confirming the significant domain gaps between the two data sources.

{\em Issue 2: Imbalanced Distribution.} 
The retrieved data follows an imbalanced distribution, as shown in Fig.~\ref{fig:domain_gap}.
Imbalanced distributions often negatively impact training.
It is worth noting that VLM's pretraining data naturally follows imbalanced or long-tailed distributions~\cite{parashar2024neglected}, because certain concepts are just rarer than others in the real world. 
Despite the extensive manual efforts to balance diverse concepts in recent work~\cite{xu2023demystifying}, the resulting pretraining dataset still exhibits long-tailed distributions.

{
\setlength{\tabcolsep}{0.9em}
\begin{table*}[t]
\small
\centering
\caption{\small
{\bf Comparison of SWAT with the state-of-the-art zero-shot and few-shot recognition methods.} 
Our first method that finetunes the visual encoder solely on few-shot examples already outperforms prior arts significantly ($\sim$3\%)! 
Following the rigorous setup without using a validation set for hyperparameter tuning~\cite{clap24}, this few-shot finetuning method has no overfitting issues, as further justified in Fig.~\ref{fig:no_overfit}.
Additionally, our final method, SWAT, achieves significant improvements by finetuning on a mix of retrieved and few-shot data.
It outperforms the previously best few-shot recognition method CLAP~\cite{clap24} by $>$6\% in accuracy.
For reference, we copy results from \cite{clap24} for the methods that use an artificially large validation set for hyperparameter tuning.
{\bf Bold} and \underline{underlined} numbers mark the best and second best numeric metrics; \textcolor{Green}{superscripts} denote improvements over CLAP~\cite{clap24}.
Detailed results on each dataset are provided in Appendix Table~\ref{tab:compare_sota_detail}.
}
\vspace{-2mm}
\label{tab:compare_sota}
\scalebox{1.0}{
\begin{tabular}{lllllll}
\toprule
&strategy & method & \textcolor{gray}{venue \& year} & \multicolumn{3}{c}{mean accuracy of nine datasets} \\
\midrule
\multirow{3}{*}{
{\bf zero-shot methods}
}   & \multirow{2}{*}{prompting-based} & OpenCLIP~\cite{cherti2023reproducible} & \textcolor{gray}{CVPR 2023} &\multicolumn{3}{c}{56.3} \\
& &REAL-Prompt~\cite{parashar2024neglected}  & \textcolor{gray}{CVPR 2024} & \multicolumn{3}{c}{62.6} \\

& retrieval-augmented & REAL-Linear~\cite{parashar2024neglected}  & \textcolor{gray}{CVPR 2024} & \multicolumn{3}{c}{64.8} \\ 

\midrule

& & & & \emph{4-shot} & \emph{8-shot} & \emph{16-shot} \\
\multirow{10}{*}{
{\bf few-shot methods}
} &\multirow{2}{*}{prompt-learning} &CoOp~\cite{zhou2022learning} &\textcolor{gray}{IJCV 2022} &61.0 &64.6 &68.4 \\
&&PLOT~\cite{chen2022plot} &\textcolor{gray}{ICLR 2023} &62.9  &65.7 &68.7 \\
\cmidrule{2-7}

& \multirow{6}{*}{adapter-based} &CLIP-Adapter~\cite{clipadapter} &\textcolor{gray}{IJCV 2023} &59.6 &64.5 &68.1 \\
&&TIP-Adapter~\cite{tipadapter}  & \textcolor{gray}{ECCV 2022} &56.6 &57.8 &59.5 \\
&&TIP-Adapter(f)~\cite{tipadapter}  & \textcolor{gray}{ECCV 2022} &60.8 &63.5 &67.1 \\
&&TaskRes(e)~\cite{taskres}  & \textcolor{gray}{ECCV 2022} &63.5 &67.1 &69.9 \\
&&CrossModal-LP~\cite{lin2023multimodality}  & \textcolor{gray}{CVPR 2023} &65.4 & 68.8 & 71.8 \\
& &CLAP~\cite{clap24} & \textcolor{gray}{CVPR 2024} & 66.9 & 70.0 & 72.9 \\
\cmidrule{2-7}

& \multirow{2}{*}{finetuning-based} 
&\cellcolor{gray!15}few-shot finetuning
& \cellcolor{gray!15}\textcolor{gray}{\bf ours}
&\cellcolor{gray!15}\underline{69.7}$^{\textcolor{Green}{+2.8}}$ 
&\cellcolor{gray!15}\underline{73.3}$^{\textcolor{Green}{+3.3}}$  
&\cellcolor{gray!15}\underline{76.3}$^{\textcolor{Green}{+3.4}}$  \\

&  & \cellcolor{gray!15}\textbf{SWAT} & \cellcolor{gray!15}\textcolor{gray}{\bf ours}
&\cellcolor{gray!15}\textbf{73.5$^{\textcolor{Green}{+6.6}}$} 
&\cellcolor{gray!15}\textbf{76.0$^{\textcolor{Green}{+6.0}}$}  
&\cellcolor{gray!15}\textbf{78.2$^{\textcolor{Green}{+5.3}}$}\\ 

\bottomrule
\end{tabular}
}
% \vspace{-2mm}
\end{table*}
}

\subsection{Stage-Wise Retrieval-Augmented Finetuning}
\label{ssec:SWAT}

To address the issues of domain gaps and imbalanced distributions, we propose Stage-Wise retrieval-Augmented fineTuning (SWAT), a two-stage solution (Fig.~\ref{fig:teaser}).
In the first stage, SWAT finetunes the VLM’s visual encoder end-to-end on a mixed set of retrieved and few-shot annotated data using cross-entropy loss. In the second stage, SWAT retrains the classifier exclusively on the few-shot data atop the finetuned encoder.
Below, we explain why SWAT can {\em jointly} address these two issues.

{\em Stage-wise learning addresses domain gaps.}
By training on a larger dataset (mixing the retrieved and few-shot annotated data),
SWAT produces a more generalizable feature representation, finetuning upon which serves as a practice of transfer learning~\cite{Li_2016_CVPR, Inoue_2018_CVPR, Hsu_2020_WACV}.
Through transfer learning, training on domain-shifted data (different from downstream-task data) can improve downstream task performance after finetuning~\cite{sharif2014cnn, xie2018pre, kornblith2019better}.
This explains why SWAT effectively addresses domain gaps and significantly enhances performance on the downstream task.

{\em Stage-wise learning addresses imbalanced distribution.}
Imbalanced distribution is a notorious challenge when training models in the real world, as extensively studied through the problem of long-tailed recognition~\cite{zhang2023deep, kang2019decoupling, cao2019learning, yang2020rethinking, alshammari2022long}.
In the literature, \cite{jiang2021decoupled} shows a simple yet effective strategy that trains a model over imbalanced or long-tailed data in the first stage to obtain generalizable feature representations, and then retrains the classifier with balancing techniques in the second stage.
Similarly, SWAT finetunes the VLM’s visual encoder on the imbalanced mix of retrieved and few-shot data to produce generalizable features. In the second stage, SWAT retrains the classifier on \emph{balanced} few-shot examples, 
effectively mitigating the imbalanced distribution issue.

{\bf Data Augmentation.}
In our work, we adopt data augmentation techniques to enrich our few-shot annotated examples.
Specifically, we apply the CutMix technique~\cite{yun2019cutmix} which creates new images by cutting a random patch from one image and pasting it to another.
CutMix has been shown to stabilize training and reduce overfitting~\cite{yun2019cutmix}.
In our experiments, CutMix significantly improves few-shot recognition accuracy with minimal computation overhead, while other more advanced augmentation techniques offer no additional gains but more computation overheads (Appendix Table~\ref{tab:ablate_msda}).

{
\setlength{\tabcolsep}{1.2em} 
\begin{table*}[t]
\small
\centering
\caption{\small
{\bf Comparison of the accuracy of common and rare classes with vs. without stage-2 classifier retraining.} 
We define the rare classes as the 10\% least frequent classes in retrieved data and the rest as the common classes.
Results show that stage-2 classifier retraining clearly improves recognition accuracy on both common and rare classes across all methods,
including finetuning on few-shot data only, on retrieved data only, and on mixed data with or without CutMix data augmentation.
Importantly, the improvement for rare classes is more significant than for common classes, confirming that classifier retraining mitigates the issue of imbalanced distribution in the retrieved data.
We report the mean accuracy over nine datasets using 16-shot examples.
Accuracy improvements compared to the model after stage-1 finetuning are marked in \textcolor{Green}{superscripts} 
and standard deviations across three runs with different random seeds are marked in \textcolor{Gray}{subscripts}.
See detailed improvements for each dataset in Appendix Table~\ref{tab:stage2_improvement_dataset}.
}
\vspace{-3mm}
\label{tab:stage2_improvement}
\begin{tabular}{l ll ll ll  cccccc}
\toprule
\multirow{4}{*}{data used in stage-1: finetuning} & \multicolumn{6}{c}{mean accuracy of nine datasets} \\ 
 & \multicolumn{3}{c}{\makecell{stage-1: finetuning}} & \multicolumn{3}{c}{\makecell{stage-2: classifier retraining on few-shot data}} \\
 \cmidrule(r){2-4} \cmidrule(r){5-7}
 & common & rare & average & common & rare & average \\ 
 \midrule
few-shot only (balanced) 
&76.8$_{\textcolor{Gray}{\pm0.2}}$  
&73.6$_{\textcolor{Gray}{\pm0.7}}$ 
&76.3$_{\textcolor{Gray}{\pm0.1}}$  
&77.0$_{\textcolor{Gray}{\pm0.1}}^{\textcolor{Green}{+0.2}}$ 
&75.1$_{\textcolor{Gray}{\pm0.9}}^{\textcolor{Green}{+1.5}}$ 
&76.8$_{\textcolor{Gray}{\pm0.2}}^{\textcolor{Green}{+0.5}}$ \vspace{+1mm}\\

retrieved only (imbalanced) 
& 64.8$_{\textcolor{Gray}{\pm0.0}}$  
& 44.2$_{\textcolor{Gray}{\pm0.0}}$  
& 62.8$_{\textcolor{Gray}{\pm0.0}}$
& 67.8$_{\textcolor{Gray}{\pm0.1}}^{\textcolor{Green}{+3.0}}$ 
& 55.3$_{\textcolor{Gray}{\pm0.9}}^{\textcolor{Green}{+11.1}}$ 
& 66.5$_{\textcolor{Gray}{\pm0.2}}^{\textcolor{Green}{+3.7}}$ \vspace{+1mm}\\

retrieved + few-shot 
& 76.1$_{\textcolor{Gray}{\pm0.1}}$  
& 68.2$_{\textcolor{Gray}{\pm0.8}}$  
& 75.3$_{\textcolor{Gray}{\pm0.1}}$  
& 77.0$_{\textcolor{Gray}{\pm0.1}}^{\textcolor{Green}{+0.9}}$ 
& 71.6$_{\textcolor{Gray}{\pm0.9}}^{\textcolor{Green}{+3.4}}$ 
& 76.4$_{\textcolor{Gray}{\pm0.1}}^{\textcolor{Green}{+1.1}}$ \vspace{+1mm}\\ 

\rowcolor{gray!15}retrieved + few-shot w/ CutMix 
& 78.0$_{\textcolor{Gray}{\pm0.1}}$  
& 71.9$_{\textcolor{Gray}{\pm0.4}}$  
& 77.3$_{\textcolor{Gray}{\pm0.1}}$  
& \bf{78.7}$_{\textcolor{Gray}{\pm0.1}}^{\textcolor{Green}{+0.7}}$ 
& \bf{74.1}$_{\textcolor{Gray}{\pm0.9}}^{\textcolor{Green}{+2.2}}$ 
& \bf{78.2}$_{\textcolor{Gray}{\pm0.2}}^{\textcolor{Green}{+0.9}}$
\\ 
\bottomrule
\end{tabular}
\vspace{-2mm}
\end{table*}
}

\section{Experiments}
\label{sec:experiments}
We conduct extensive experiments to demonstrate that SWAT significantly outperforms previous few-shot recognition methods. 
We begin with the experiment setup, followed by benchmarking results. 
We also ablate important design choices with extensive analyses.

\subsection{Experimental Setup}
{\bf Datasets and Metrics.}
Motivated from the data annotation perspective,
we study FSR through challenging real-world tasks where the data
annotation requires domain expert knowledge, 
such as
recognizing bird species, aircraft models, satellite images, etc. 
Specifically, we follow~\cite{fang2024does, saha2024improved} to use fine-grained recognition datasets where the VLM CLIP~\cite{clip} struggles to recognize the nuanced attributes~\cite{saha2024improved}.
We select nine datasets, namely Semi-Aves~\cite{semi-aves}, Flowers102~\cite{flowers}, FGVC-Aircraft~\cite{aircraft}, EuroSAT~\cite{aircraft}, DTD~\cite{dtd}, OxfordPets~\cite{pets}, Food101~\cite{food}, StanfordCars~\cite{cars}, and ImageNet~\cite{deng2009imagenet} (see detailed descriptions in Appendix Table~\ref{tab:datasets}). 
We evaluate SWAT using the OpenCLIP ViT-B/32 model~\cite{openclip} (unless otherwise specified) and retrieve images from its publicly available LAION-400M pretraining set~\cite{laion400m, laion5b}.
We report the accuracy averaged over nine datasets.
The appendix includes results using other VLM architectures, on which our conclusions still hold.

{
\setlength{\tabcolsep}{0.1em} 
\begin{table}[t]
\small
\centering
\caption{\small
{\bf Ablation study on important components in our SWAT.}
Compared to the state-of-the-art adapter-based few-shot method CLAP~\cite{clap24}, our few-shot finetuning (FSFT) obtains $>$2\% improvement! 
Naively adding retrieved data for finetuning brings small improvements $\sim$1\% due to the domain gaps and imbalanced distribution of the retrieved data.
Applying CutMix provides an additional $\sim$2\% improvement,
and retraining the classifier in the second stage brings a further $\sim$0.5\% increase.
Note the improvements of classifier retraining are more significant for the rare classes ($>$2\%), especially when the stage-1 finetuning data is imbalanced retrieved data ($>$11\%), as shown in Table~\ref{tab:stage2_improvement}. 
\textcolor{Green}{Superscripts} indicate the improvements of each component (relative to the corresponding row above).
We report mean accuracies averaged over nine benchmark datasets. 
Detailed results are listed in Appendix Table~\ref{tab:components_contribution_dataset}.
}
\vspace{-3mm}
\label{tab:components_contribution}
\scalebox{0.86}{
\begin{tabular}{lcccclll}
\toprule
\multirow{2}{*}{\makecell{method}} &\multirow{2}{*}{\makecell{finetune\\model}} &\multirow{2}{*}{\makecell{retrieve\\data}} & \multirow{2}{*}{\makecell{apply\\CutMix}} & \multirow{2}{*}{\makecell{retrain\\classifier}} & \multicolumn{3}{c}{mean acc. of nine datasets} \\ 
\cmidrule(r){6-8}
&& &  &  & 4-shot & 8-shot & 16-shot \\ 
\midrule
CLAP~\cite{clap24} & & & & & 66.9 & 70.0 & 72.9 \\

FSFT (ours) & \checkmark & &   & & 69.4$^{\textcolor{Green}{+2.5}}$ & 72.7$^{\textcolor{Green}{+2.7}}$ & 75.1$^{\textcolor{Green}{+2.2}}$ \\

&\checkmark &\checkmark &  &  & 70.8$^{\textcolor{Green}{+1.4}}$ & 73.0$^{\textcolor{Green}{+0.3}}$ &75.3$^{\textcolor{Green}{+0.2}}$ \\

&\checkmark &\checkmark & \checkmark &  & 73.0$^{\textcolor{Green}{+2.2}}$ & 75.2$^{\textcolor{Green}{+2.2}}$ & 77.3$^{\textcolor{Green}{+2.0}}$ \\

\rowcolor{gray!15}SWAT (ours)&\checkmark &\checkmark & \checkmark & \checkmark & \textbf{73.5$^{\textcolor{Green}{+0.5}}$} & \textbf{76.0$^{\textcolor{Green}{+0.8}}$} & \textbf{78.2$^{\textcolor{Green}{+0.5}}$}\\ 
\bottomrule
\end{tabular}}
\vspace{-3mm}
\end{table}
}

{\bf Compared Methods.}
We compare SWAT with state-of-the-art zero-shot and few-shot recognition methods.
For zero-shot methods, we study OpenCLIP~\cite{openclip} and
the recent methods: REAL-Prompt~\cite{parashar2024neglected} and REAL-Linear~\cite{parashar2024neglected}. 
For few-shot methods, we compare against prompt-learning-based methods CoOp~\cite{zhou2022learning} and PLOT~\cite{chen2022plot}, and adapter-based methods CLIP-Adapter~\cite{clipadapter}, TIP-Adapter~\cite{tipadapter}, TaskRes~\cite{taskres}, Cross-modal Linear Probing~\cite{lin2023multimodality} and CLAP~\cite{clap24}. 
Note that CLAP is the state-of-the-art FSR method that strictly adopts the realistic validation-free setup~\cite{clap24}.
We present the results of different finetuning methods in Table~\ref{tab:ablate_finetuning method} of the Appendix.

{\bf Implementations.}
For each dataset, we retrieve 500 images per class following~\cite{parashar2024neglected}.
We strictly follow the validation-free protocol as done in~\cite{clap24} and set the same hyperparameters in all datasets (see details in Appendix Section~\ref{sec:hyperparams}). 
We conduct experiments using 4, 8, and 16 shots of data, randomly sampled from downstream training sets with three different seeds. We report their average accuracy.
Note that in the motivational application, data annotation guidelines provide  \emph{multiple} visual examples.
Therefore, we do not conduct experiments with fewer shots (e.g., 
1 or 2-shot) in our experiments. 
We run all experiments on a Quadro RTX 6000 (24GB) GPU with 50GB storage for hosting retrieved data for all datasets.
SWAT takes 2.5 hours for larger datasets like Semi-Aves (200 classes) and less than 4 minutes for smaller ones like EuroSAT (10 classes).
Table~\ref{tab:compute_cost} compares its compute cost with previous methods.
We provide our code instructions in Appendix Section~\ref{sec:Demo-code}.

\subsection{Benchmarking Results and Ablation Studies}

{\em SWAT significantly outperforms previous methods.}
Table~\ref{tab:compare_sota} shows that simply finetuning on few-shot data readily outperforms the previous state-of-the-art methods without bells and whistles.
Moreover, our SWAT surpasses previous methods by over 6\% in accuracy.
In challenging datasets like Semi-Aves, EuroSAT, and Aircraft, our improvements are even more significant (10-25\%, cf. Appendix Table~\ref{tab:compare_sota_detail}).
These results highlight the efficacy of our SWAT for FSR.

{\em Stage-wise finetuning improves rare class accuracy.}
Table~\ref{tab:stage2_improvement} shows overall improvements, with particularly significant gains on rare classes, achieved through stage-wise finetuning and data augmentation. These results confirm the effectiveness of classifier retraining in handling imbalanced learning on retrieved data.

{\em Ablation study.} 
Table~\ref{tab:components_contribution} 
shows that full finetuning leads to substantial accuracy gains ($>$2\%).
Simply adding retrieved data for finetuning brings slight gains (0.2-1.4\%).
Applying the CutMix data augmentation and classifier retraining as our SWAT brings more significant accuracy improvement ($>$2\%).
The results confirm that our SWAT effectively mitigates the two issues of domain gaps and imbalanced distributions of retrieved data.

\begin{figure}[t]
  \centering
  \includegraphics[width=0.99\linewidth, clip=true,trim = 0mm 0mm 0mm 0mm]
  {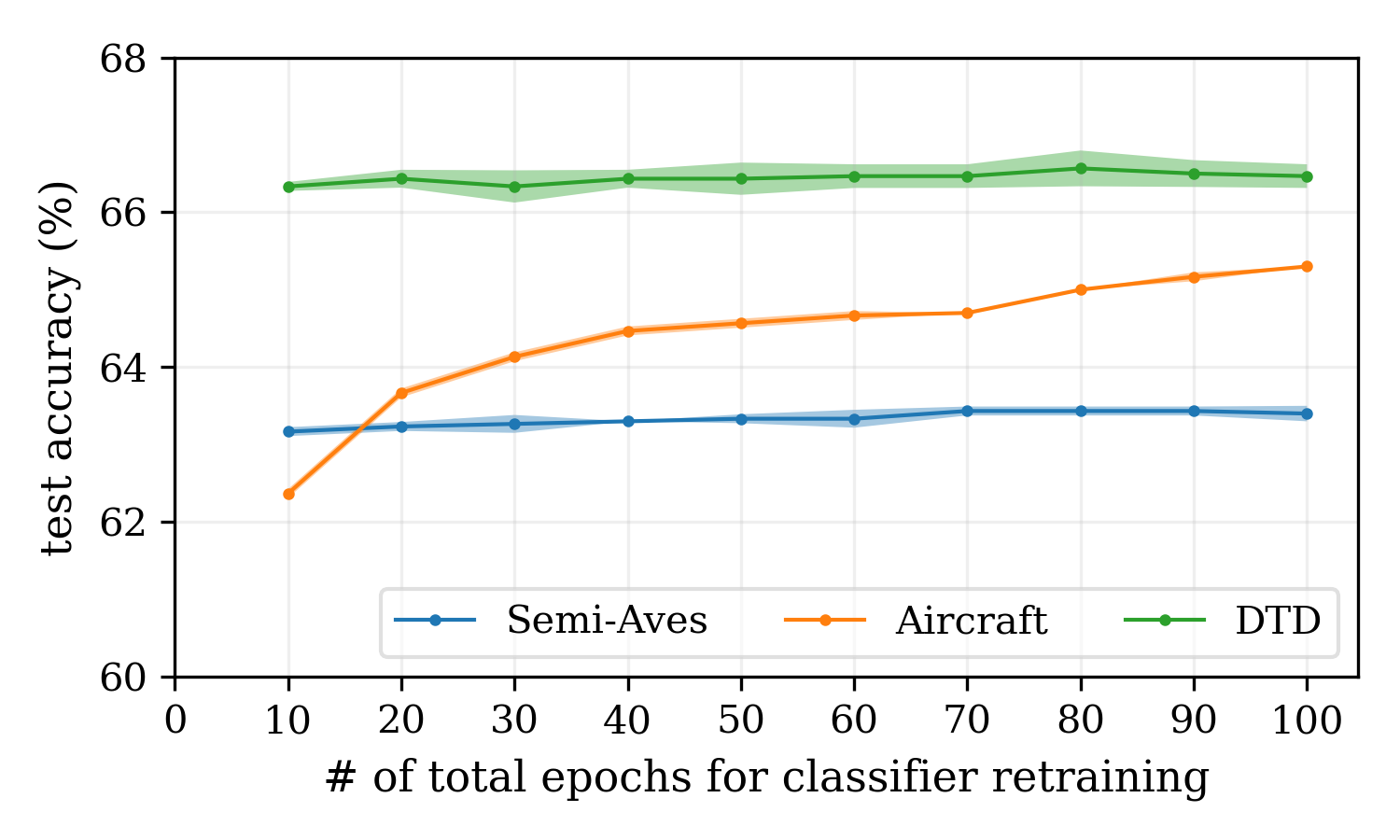}
  \vspace{-3mm}
  \caption{\small
  {\bf Retraining the classifier on the few-shot data does not suffer from overfitting.}
  We show the testing accuracies by retraining the classifier on 16 few-shot data at different epochs.
  We perform three runs of training with different random seeds.
  Results show that the testing accuracy does not decrease with more epochs and shows small standard deviations.  
  We show the accuracy plots for other datasets in Appendix Fig.~\ref{fig:no_overfit_more}.
  }
  \label{fig:no_overfit}
\vspace{-4mm}
\end{figure}

{\em Longer training does not suffer from overfitting.}
Fig.~\ref{fig:no_overfit} shows that testing accuracy does not decrease with more training epochs.
This validates that learning on few-shot data with a strong pretrained backbone does not suffer from overfitting.
This encourages future work in FSR to safely adopt the more rigorous validation-free setting, i.e., not using an artificially large validation set for hyperparameter tuning.

{\em Further analyses.}
We conduct more analyses and provide them in the Appendix.
We summarize the salient results here:
properly filtering retrieved data improves performance further on some datasets (Table~\ref{tab:dtd_cars});
the string-matching-based retrieval yields the best overall results, though different retrieval methods may excel on specific datasets (Table~\ref{tab:retrieve_methods});
CutMix~\cite{yun2019cutmix} slightly outperforms other data augmentation methods with minimal computation overhead (Table~\ref{tab:ablate_msda});
SWAT outperforms other ensembling-based\cite{wortsman2022robust} or contrastive finetuning methods\cite{goyal2023finetune} (Table~\ref{tab:ablate_finetuning method});
initializing the classifier with text embedding before finetuning surpasses random initialization (Table~\ref{tab:ablate_cls_init}), partly explaining why finetuning on few-shot data does not overfit;
learning on more retrieved images shows diminishing performance gains (Table~\ref{tab:ablation-retr}).

\section{Discussions}
\label{sec:discussion}

{\bf Broader Impacts.} 
Our work explores practical few-shot recognition methods with a real-world motivation of data annotation,
where annotation guidelines provide a few visual examples for each concept of interest \cite{madan2024revisiting}.
Yet, similar to prior works,
our methods exploit a pretrained VLM,
using which may have negative impacts as VLMs could have learned biases from the Internet data. 
Moreover, our work exploits retrieval augmented learning to retrieve VLM's pretraining data, which may also deliver biases.
Lastly,
our work has not evaluated the proposed method in the real-world data annotation application; we expect more challenges when applying our methods in the real world.

{
\setlength{\tabcolsep}{0.4em} 
\begin{table}[t]
\centering
\caption{\small
{\bf Comparison of the compute cost of our SWAT with state-of-the-art few-shot recognition methods.}
We measure the GPU memory used and total training time using the Semi-Aves dataset (200 classes with 16-shot), and report mean accuracies averaged over nine datasets.
Results show that simply finetuning on the few-shot data improves accuracy by 3\% with slightly more training time and GPU memory cost than CLAP.
In addition, SWAT improves accuracy significantly by >5\% with very affordable retrieval and training costs.
We highlight the accuracy of our methods in \textbf{bold}.
\textcolor{Green}{Superscripts} mark improvements compared to previous state-of-the-art CLAP~\cite{clap24}.
}
\vspace{-2mm}
\label{tab:compute_cost}
\scalebox{0.9}{
\begin{tabular}{l l ccl}
\toprule
\multicolumn{1}{l}{method} & \textcolor{gray}{venue \& yr} & \makecell{mem.} & time & \makecell{mean acc.} \\
\midrule
\multicolumn{1}{l}{CrossModal LP~\cite{lin2023multimodality}} & \textcolor{gray}{CVPR'23} & 2 GB & 2 mins & 71.8 \\
\multicolumn{1}{l}{CLAP~\cite{clap24}} & \textcolor{gray}{CVPR'24} & 2 GB & 2 mins & 72.9 \\
\midrule
\multicolumn{1}{l}{few-shot finetuning} & \textcolor{gray}{ours} & 5 GB & 20 mins & \textbf{76.3$^{\textcolor{Green}{+3.4}}$} \\ 
% \midrule
\multirow{1}{*}{SWAT retrieval} & \textcolor{gray}{ours} & 2 GB & 1 hr 
& \multirow{2}{*}{\textbf{78.2$^{\textcolor{Green}{+5.3}}$}} \\
\multirow{1}{*}{SWAT training} & \textcolor{gray}{ours} & 5 GB & 2.5 hrs &  \\
\bottomrule
\end{tabular}}
\vspace{-3mm}
\end{table}
}

{\bf Limitations and Future Work.}
We note several limitations and future directions.
First, although our method 
achieves
state-of-the-art performance by simply 
using hyperparameters reported in the literature,
we believe that they are not necessarily optimal for each FSR task.
Future work can explore human-in-the-loop intervention for hyperparameter tuning, e.g., manually constructing a validation set using retrieved data to tune hyperparameters.
Second, as our method exploits concept names to retrieve pretrained data, future work can employ language models to generate richer descriptions or use the definitions provided in annotation guidelines for better retrieval.
Third, despite our work showing that CutMix is effective in mitigating domain gaps and imbalanced distribution, it is worth exploring novel data augmentation methods to further address these issues.
Lastly, although our method jointly solves the imbalance issue and domain gap of the retrieved data, future work can design experiments to decouple these two factors for further analysis and develop novel solutions.

\section{Conclusions}

We explore few-shot recognition (FSR) motivated from a data annotation perspective.
This encourages one to solve FSR towards higher accuracy with a rigorous setup, unlike contemporary FSR works that focus on parameter-efficient finetuning a pretrained Vision-Language Model (VLM) in a flawed setup, which uses an unrealistically large validation set for hyperparameter tuning.
We first examine an embarrassingly simple method that finetunes a VLM's visual encoder on few-shot data.
Surprisingly, it readily outperforms previous methods.
Then, we adopt the retrieval-augmented learning strategy, which retrieves VLM's pretraining examples relevant to the concepts of interest to facilitate learning. 
We identify two interesting issues in the retrieved data: its imbalanced distributions, and its domain gaps with the few-shot examples.
We propose a simple yet effective method, Stage-Wise retrieval-Augmented fineTuning (SWAT), to address these two issues.
Across nine benchmark datasets,
our SWAT outperforms previous methods by 6\% accuracy on average,
achieving the state-of-the-art FSR performance.

\section*{Acknowledgements}

This work was supported by FDCT (0067/2024/ITP2), University of Macau (SRG2023-00044-FST), and the Institute of Collaborative Innovation.
Tian Liu acknowledges the support from Prof. James Caverlee and the CSE Department at Texas A\&M University.
Portions of this research were conducted with the advanced computing resources and consultation provided by Texas A\&M High Performance Research Computing.

{
    \small
    \bibliographystyle{ieeenat_fullname}
    \bibliography{main}
}

\clearpage
\maketitlesupplementary

\renewcommand{\thesection}{\Alph{section}}
\renewcommand{\theHsection}{\Alph{section}}
\setcounter{section}{0} 

\section*{}
\begin{center}
    \emph{\bf \em \large Outline}
\end{center}
This document supports our main paper with detailed results and comprehensive analyses. The document is organized as below: 

\begin{itemize}
\item {\bf Section \ref{sec:datasets}.} 
We provide a detailed summary of benchmarking datasets used in our experiments. 

\item {\bf Section \ref{sec:hyperparams}.} 
We provide details of hyperparameters used in our work.

\item {\bf Section \ref{sec:swat-compare}.} 
We report detailed results of comparing SWAT with previous FSR methods for each benchmark dataset.

\item {\bf Section \ref{sec:retrieval_methods}.} 
We provide details on how we retrieve pretraining data and compare different retrieval and filtering methods.

\item {\bf Section \ref{sec:msda_example}.} 
We compare different mixed sample data augmentation methods and analyze the impact of the mixing ratio within a batch.

\item {\bf Section \ref{sec:swat+}.} 
We validate the design of our SWAT by ablating different stage-2 training strategies and comparing SWAT with recent state-of-the-art finetuning methods.

\item {\bf Section \ref{sec:ablation-studies}.} 
We provide further analyses on SWAT, including the impact of training epochs, different classifier initialization methods, and more detailed experimental results.

\item {\bf Section \ref{sec:retrieved-stats}.} 
We provide analysis of the imbalance of retrieved data and the impact of retrieval size.

\item {\bf Section \ref{sec:Demo-code}.} 
We provide code and instructions for replicating our experiments.

\end{itemize}

{
\setlength{\tabcolsep}{0.1em} 
\section{Summary of Datasets}
\label{sec:datasets}
We summarize the nine fine-grained datasets used in our experiments in Table~\ref{tab:datasets}. 
Following \cite{lin2023multimodality, clap24}, we sample few-shot data from the official training set and evaluate model performance on the official test set, except for ImageNet where we report performance on its validation set. We repeat each experiment three times with three random seeds.
Note that we strictly follow the validation-free protocol \cite{clap24} that we do not use any validation data for hyperparameter tuning or model selection.

{
\setlength{\tabcolsep}{0.2em}
\begin{table}[ht]
\centering
\small
\caption{\small \textbf{Statistics of nine fine-grained datasets repurposed in our work.}
We list the number of images in the official training, validation, and test sets for each dataset. 
The protocol of few-shot recognition samples few-shot data from the official training set; we use them as \emph{our train set}. We repeat the sampling and training three times for each method with three random seeds.
To evaluate methods, we repurpose their official test set as \emph{our test set} (except on ImageNet where we use its official validation set as our test set).
We benchmark methods on \emph{our test sets}.
Note again that we \emph{do not} use any validation examples for model selection or hyperparameter tuning; instead, we strictly adhere to the realistic validation-free protocol for few-shot research \cite{clap24}.
}
\label{tab:datasets}
\vspace{-2mm}
\scalebox{0.73}{
\begin{tabular}{lrrrrc}
\toprule 
dataset & \# cls &official-train & official-val & official-test & task \\
\midrule
Semi-Aves~\cite{semi-aves} & 200 & 3,959 & 2,000 & 4,000 & recognize birds \\
Flowers~\cite{flowers} & 102 & 4,093 & 1,633 & 2,463 & recognize flowers  \\
Aircraft~\cite{aircraft} & 100 & 3,334 & 3,333 & 3,333 & recognize aircrafts \\
EuroSAT~\cite{eurosat} & 10 & 13,500 & 5,400 & 8,100 & classify satellite images  \\
DTD~\cite{dtd} & 47 & 2,820 & 1,128 & 1,692 & recognize textures \\
OxfordPets~\cite{pets} &37 &2,944 & 736 & 3,669 & recognize pets \\
Food101~\cite{food} &101 &50,500 & 20,200 & 30,300 & recognize food \\
StanfordCars~\cite{cars} &196 &6,509 & 1,635 & 8,041 & recognize cars \\
ImageNet~\cite{deng2009imagenet} &1,000 &1.28M & 50,000 & N/A &large scale recognition  \\
\bottomrule
\end{tabular}}
% \vspace{-3mm}
\end{table}
}
}

\begin{figure*}[t]
    \centering
    \small
    % \vspace{-3mm}
    \begin{tabular}{p{5.5cm}p{5.5cm}p{5.5cm}}
    \includegraphics[width=1.0\linewidth, clip=true,trim = 0mm 0mm 0mm 0mm]{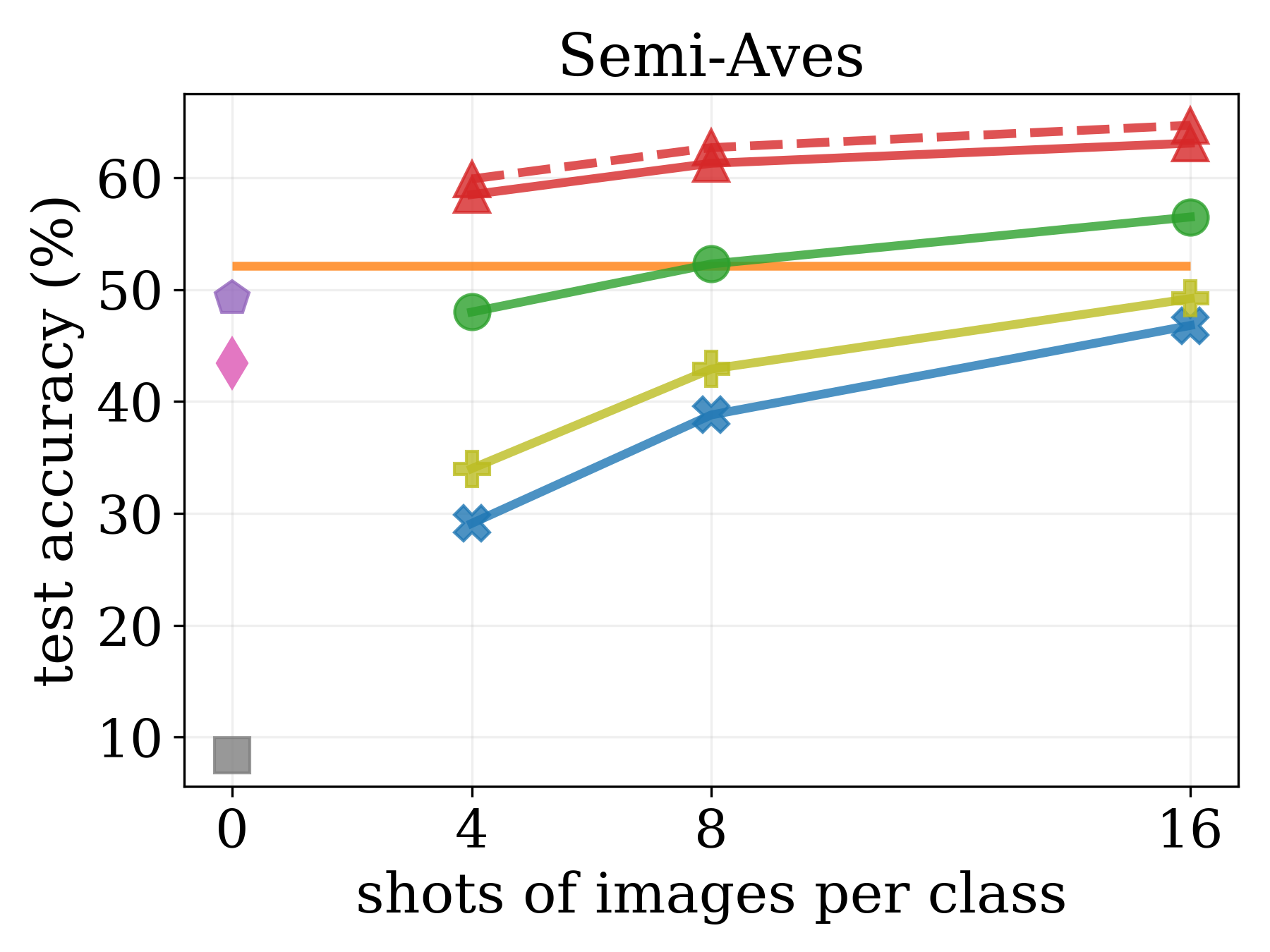} 
    &
    \includegraphics[width=1.0\linewidth, clip=true,trim = 0mm 0mm 0mm 0mm]{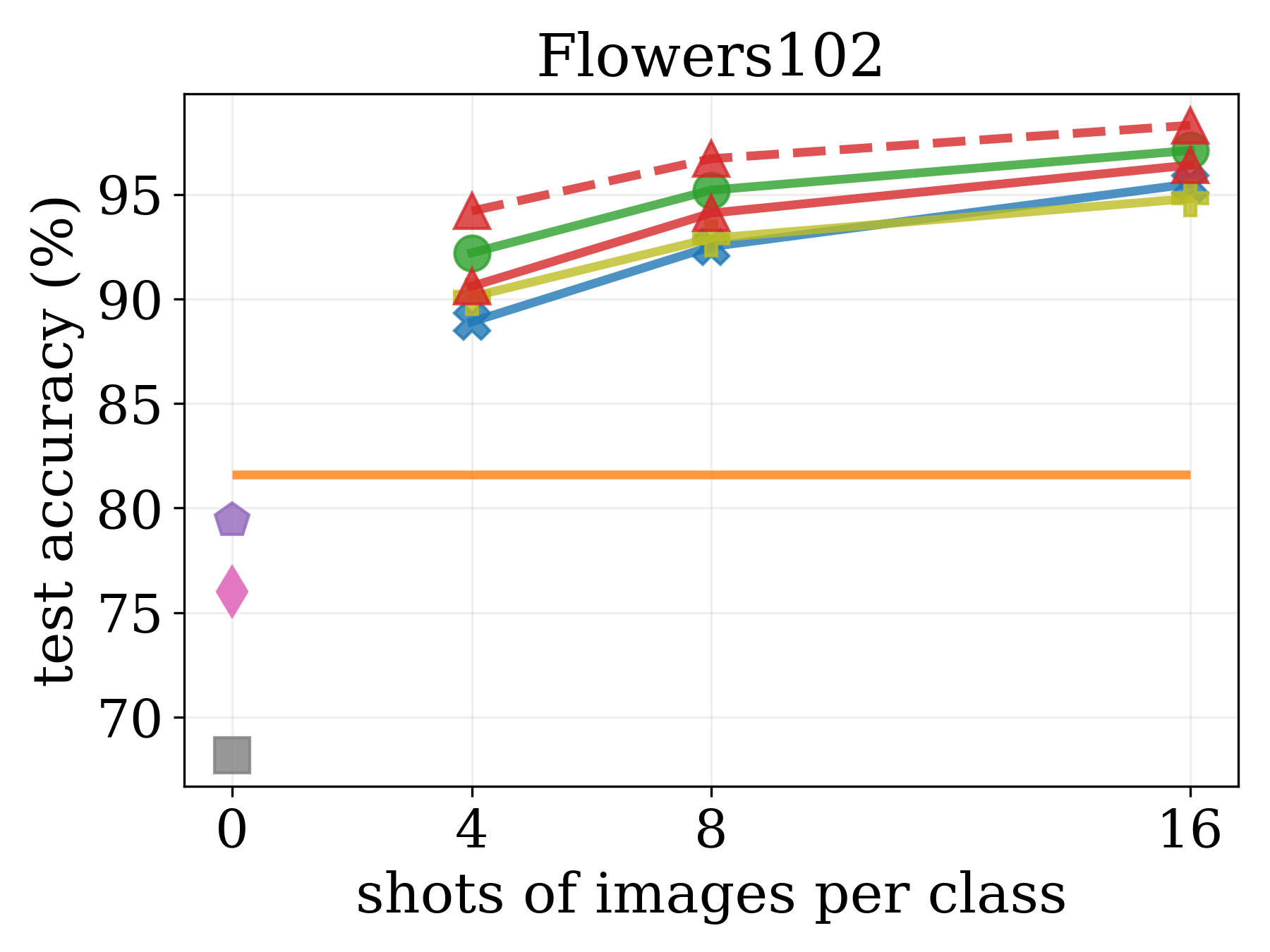} 
    &
    \includegraphics[width=1.0\linewidth, clip=true,trim = 0mm 0mm 0mm 0mm]{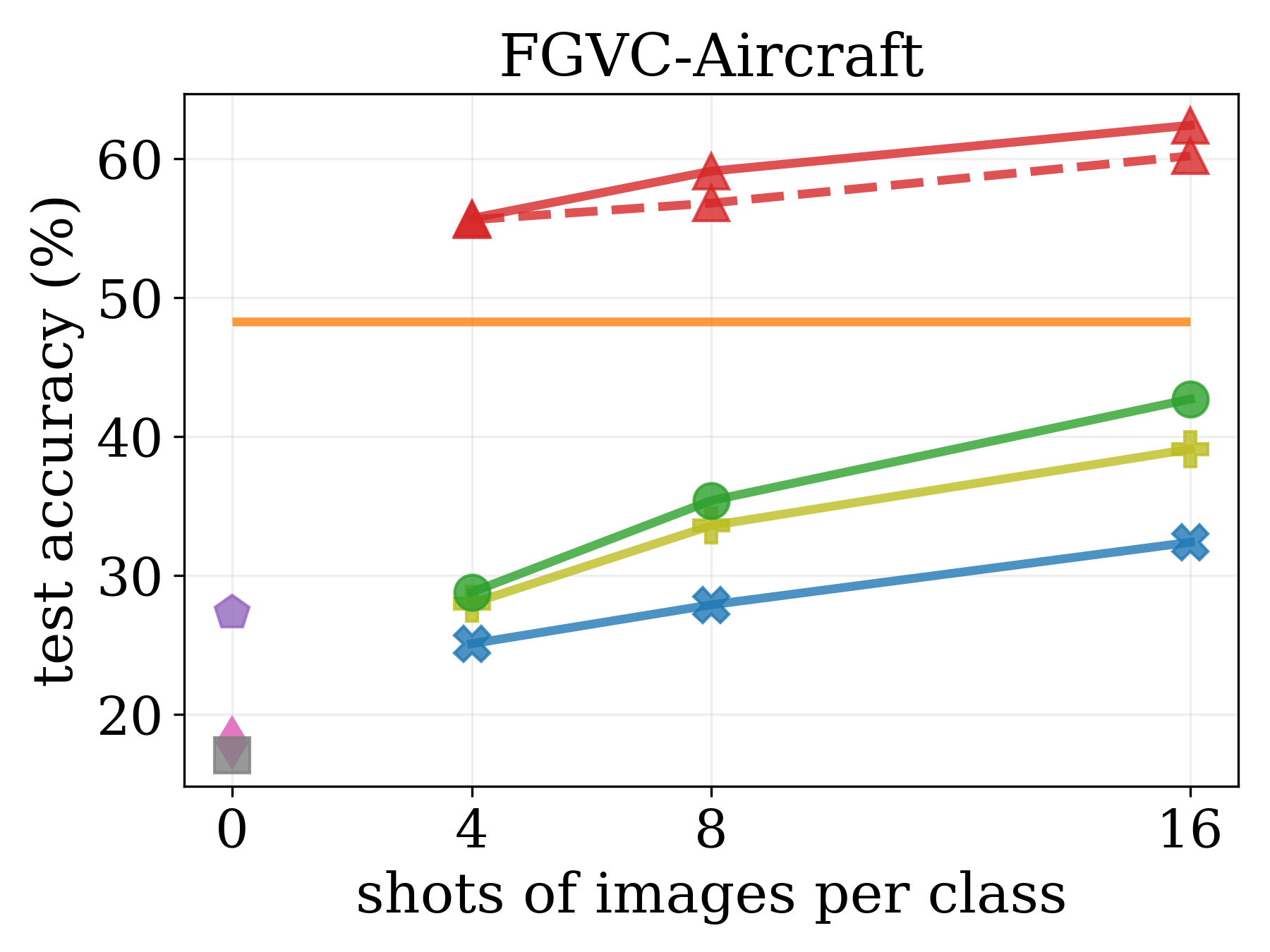}
    
    \\
    \includegraphics[width=1.0\linewidth, clip=true,trim = 0mm 0mm 0mm 0mm]{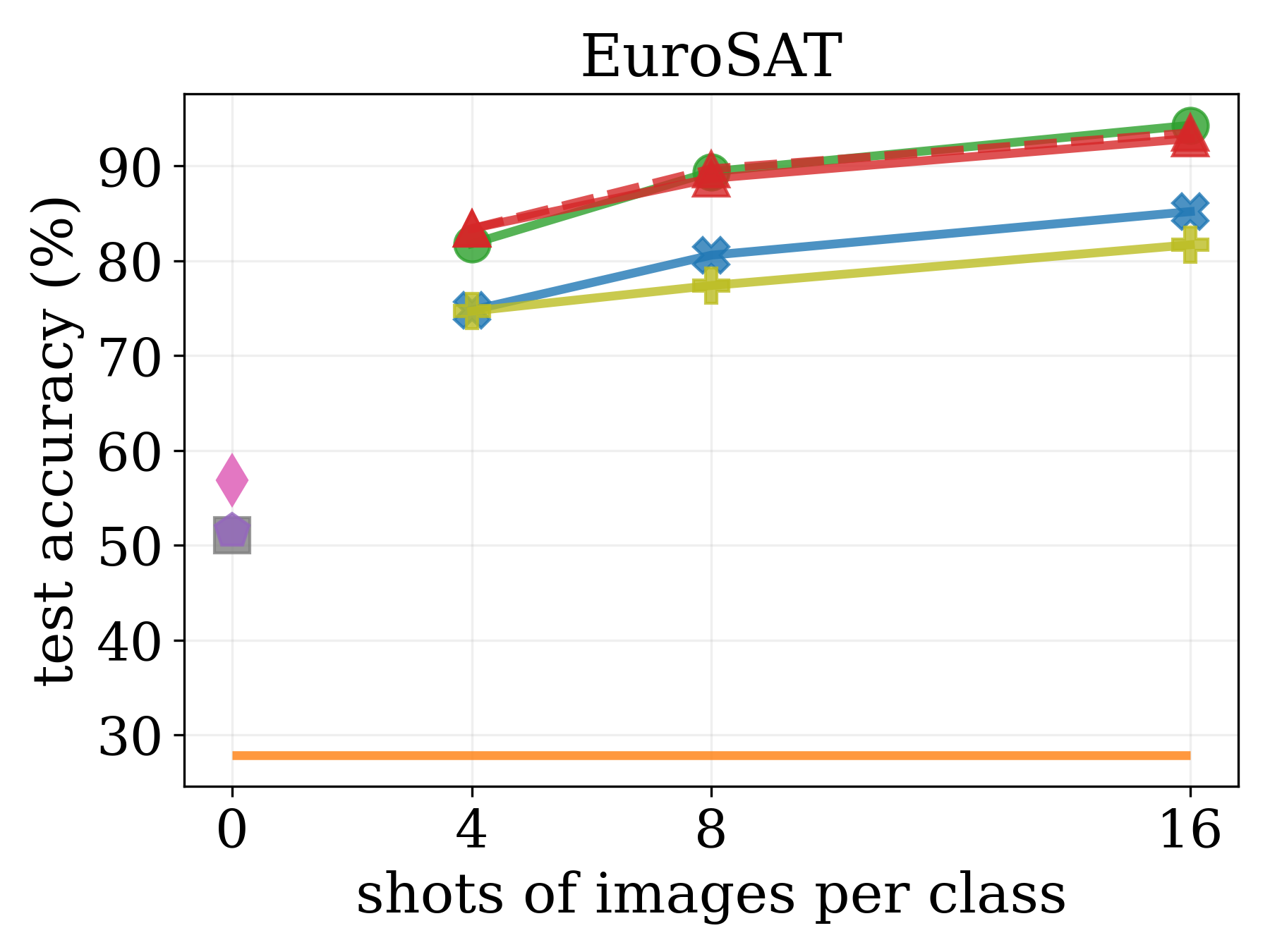}   
    &
    \includegraphics[width=1.0\linewidth, clip=true,trim = 0mm 0mm 0mm 0mm]{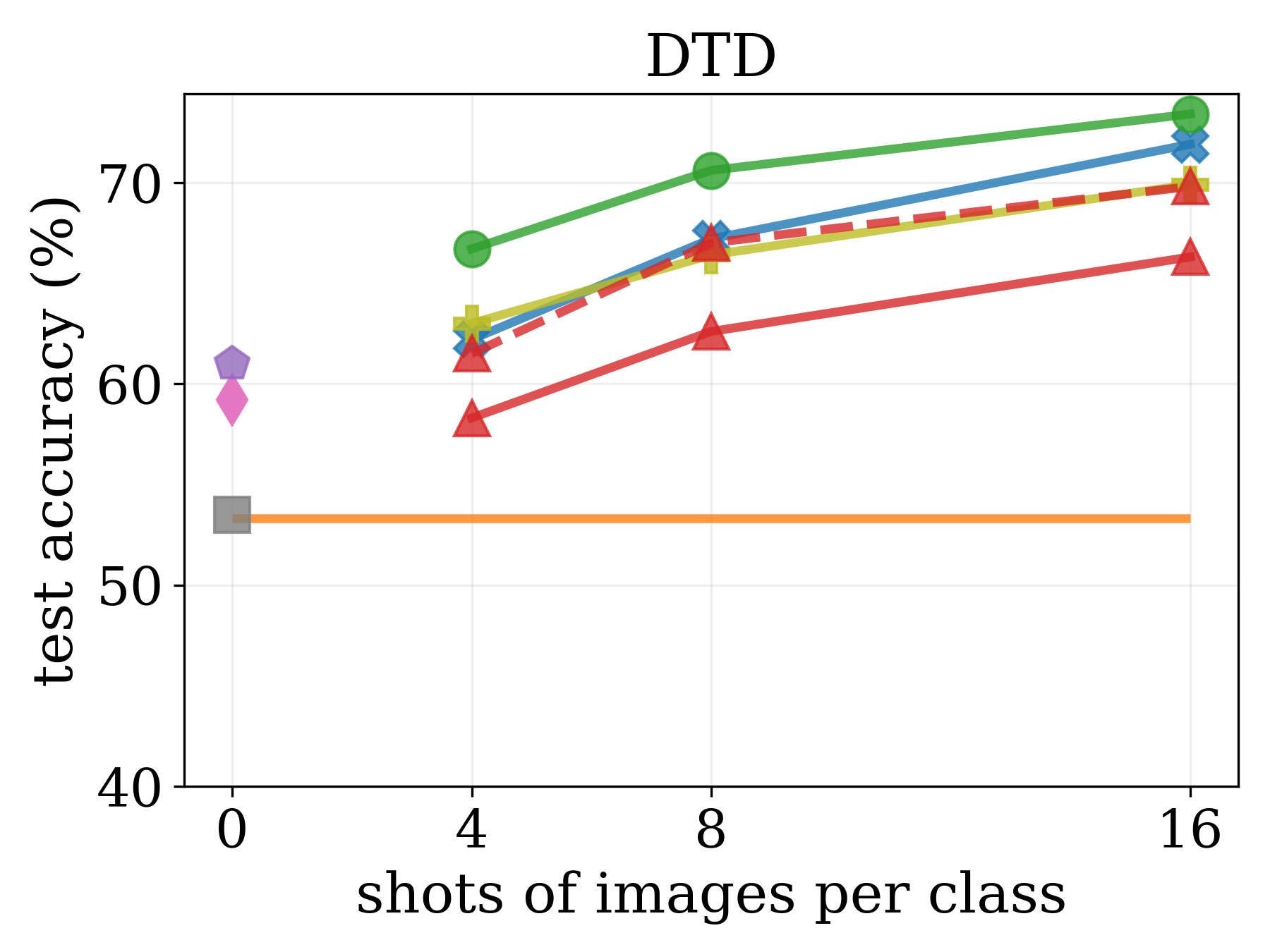}
    &
    \includegraphics[width=1.0\linewidth, clip=true,trim = 0mm 0mm 0mm 0mm]{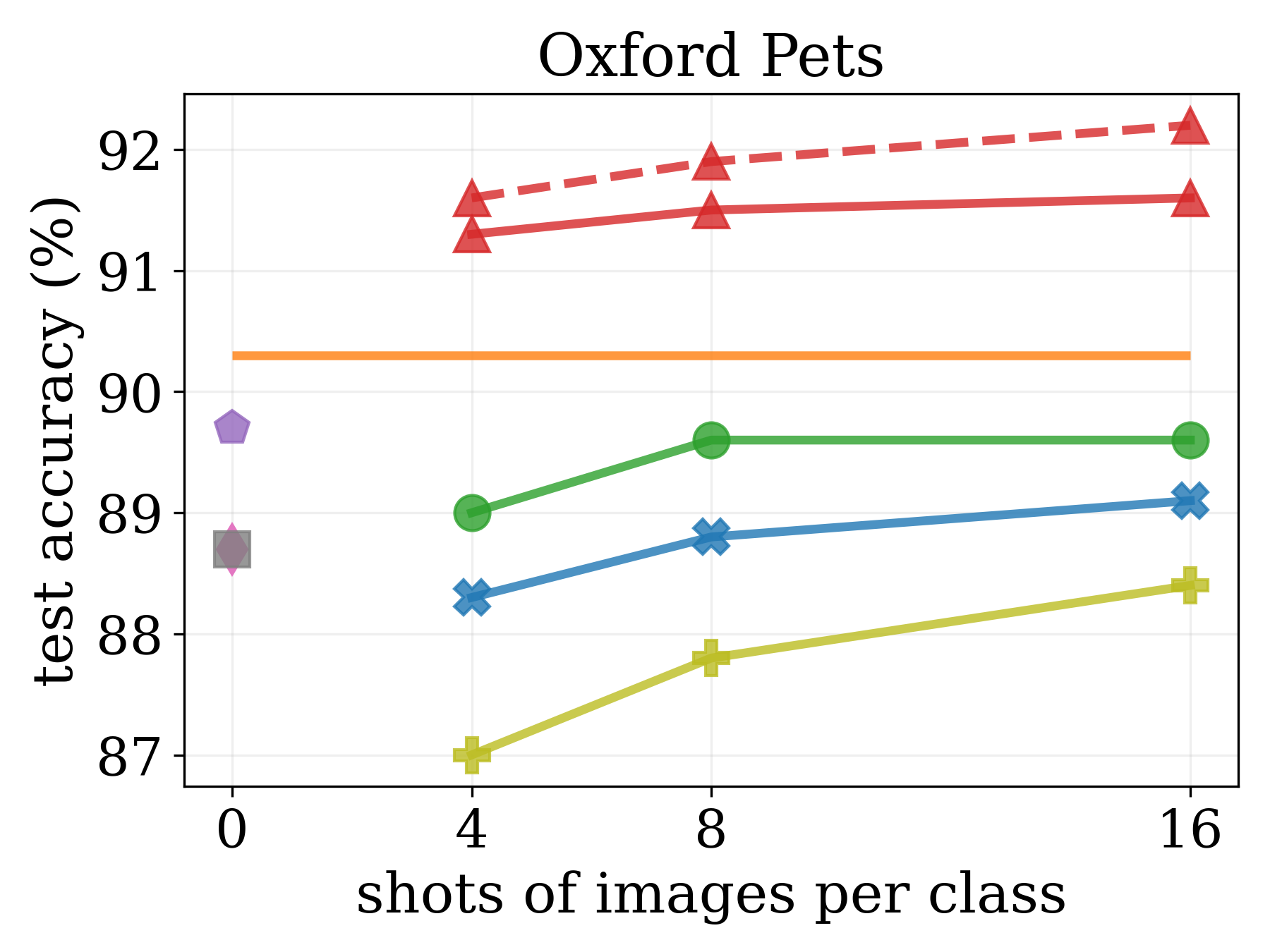}      
    
    \\
    \includegraphics[width=1.0\linewidth, clip=true,trim = 0mm 0mm 0mm 0mm]{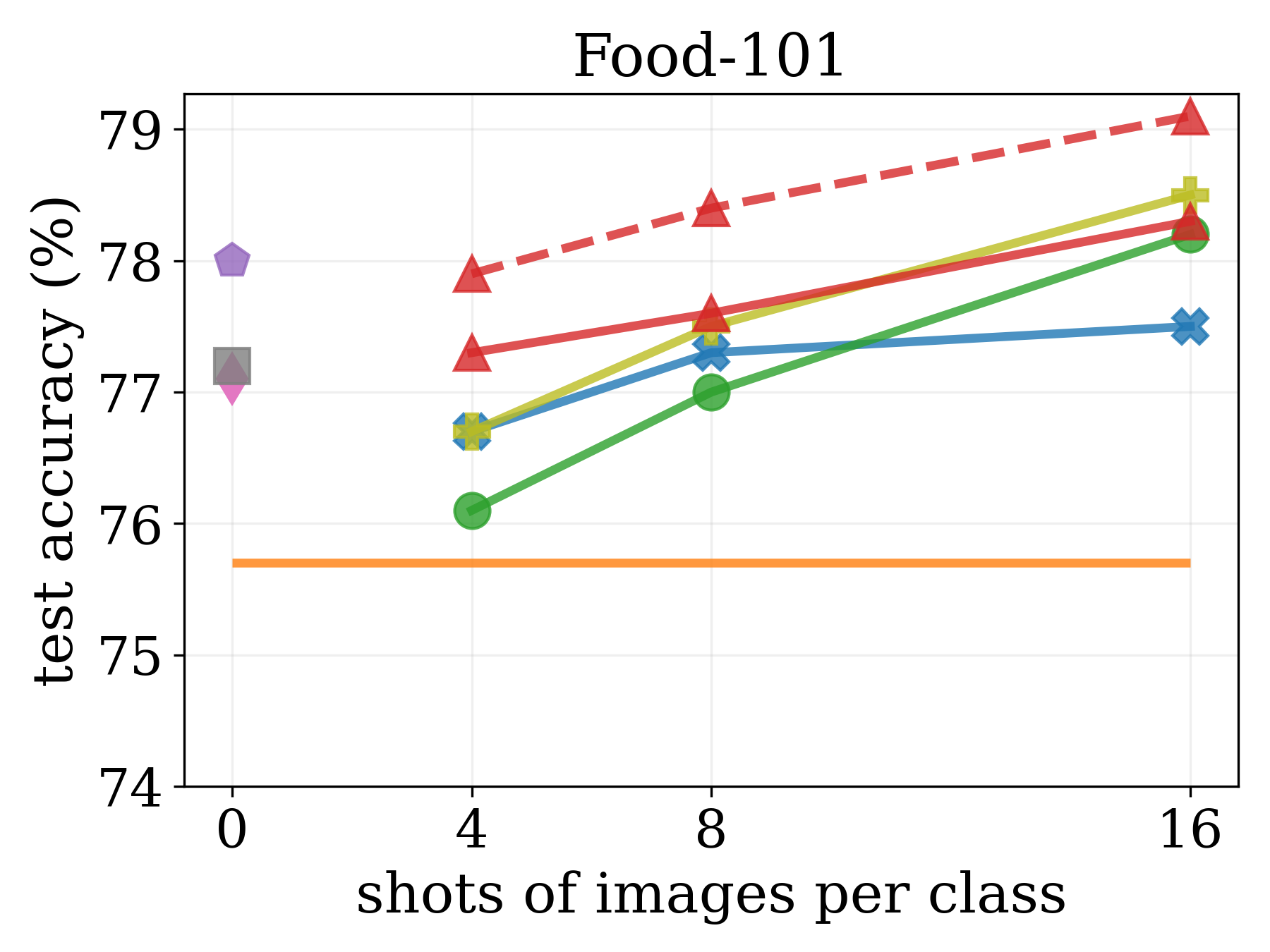}    
    &    
    \includegraphics[width=1.0\linewidth, clip=true,trim = 0mm 0mm 0mm 0mm]{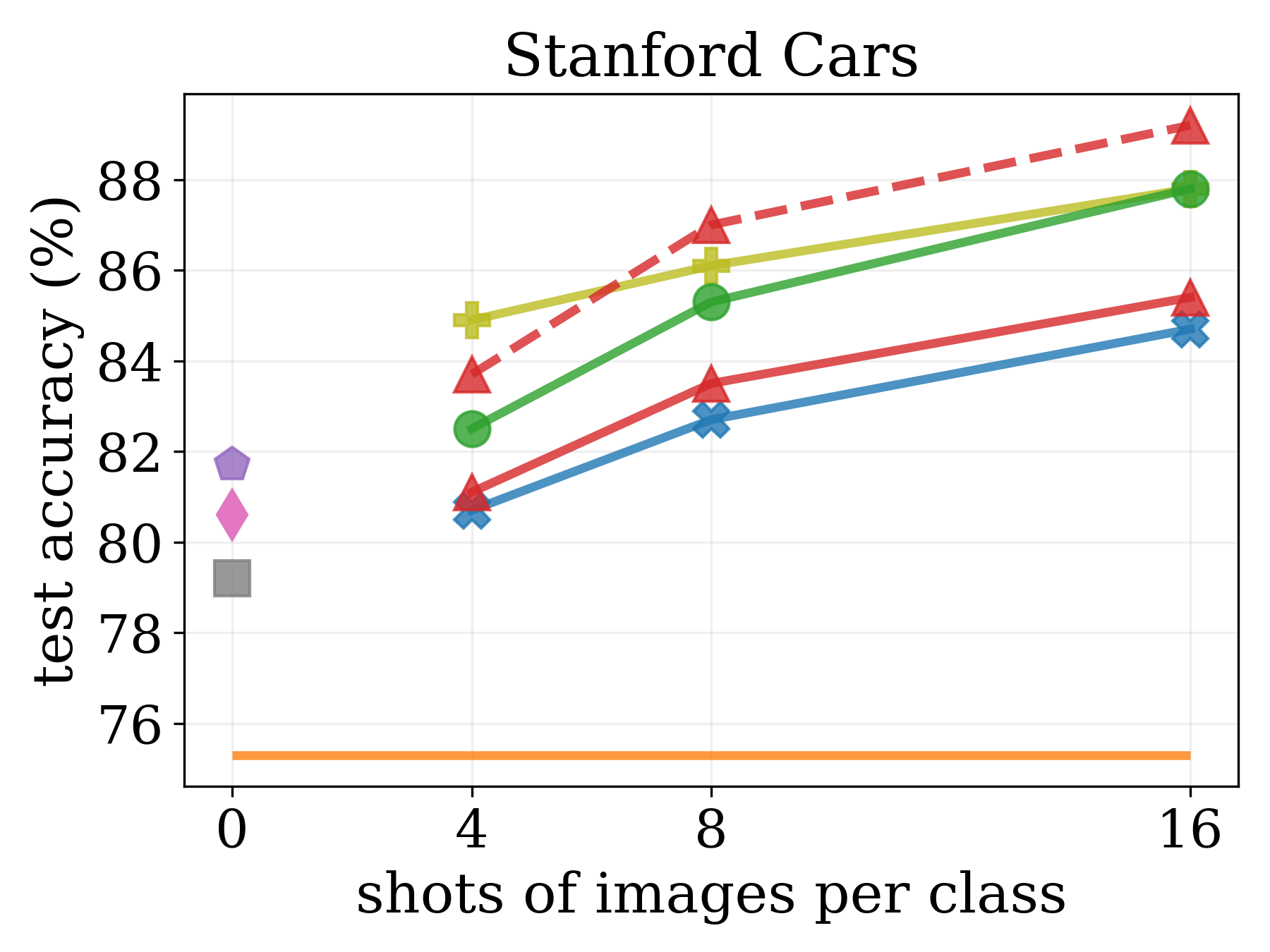}  
    &
    \includegraphics[width=1.0\linewidth, clip=true,trim = 0mm 0mm 0mm 0mm]{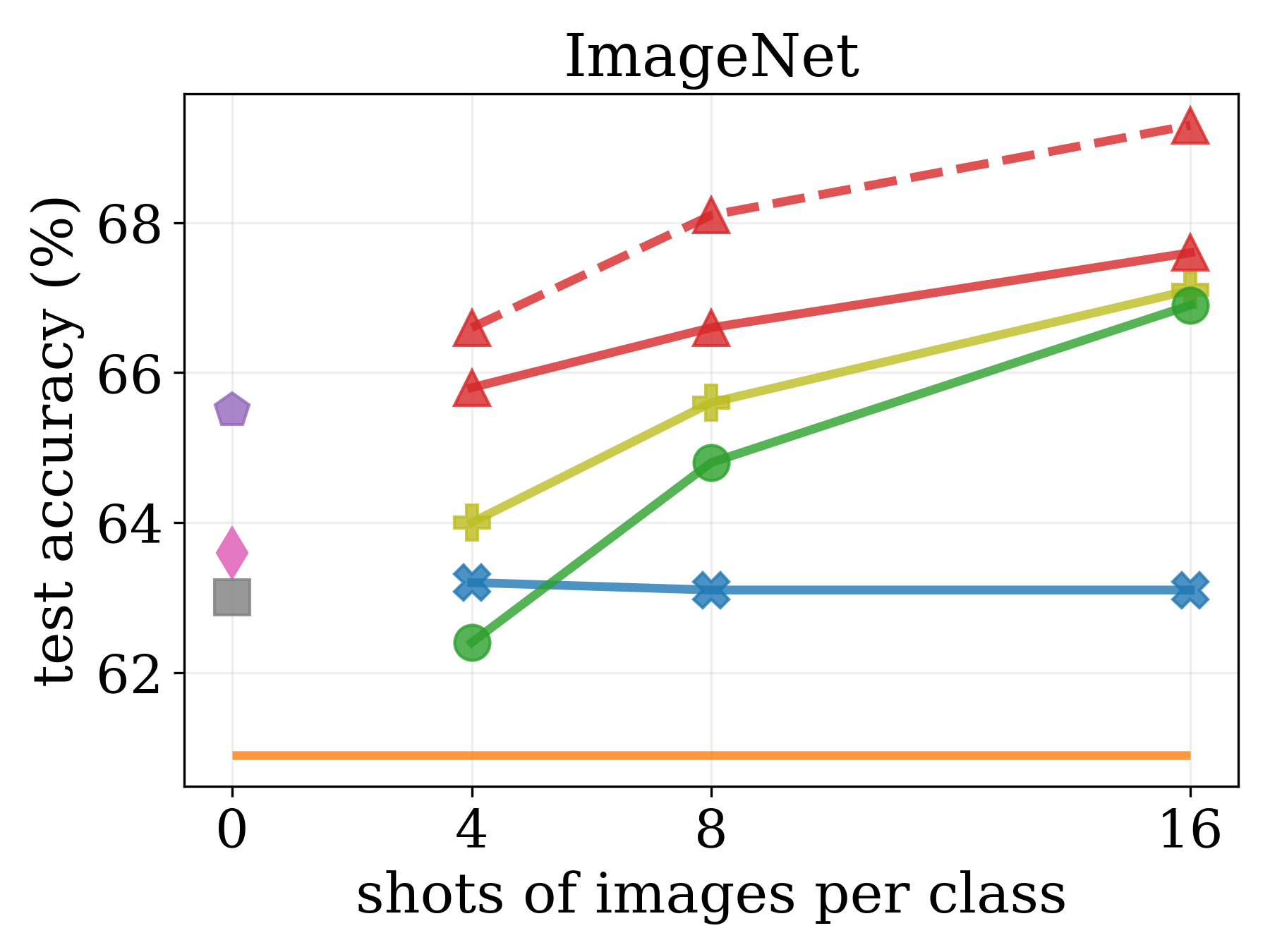} 
    
    \\    
    \includegraphics[width=1.0\linewidth, clip=true,trim = 0mm 0mm 0mm 0mm]{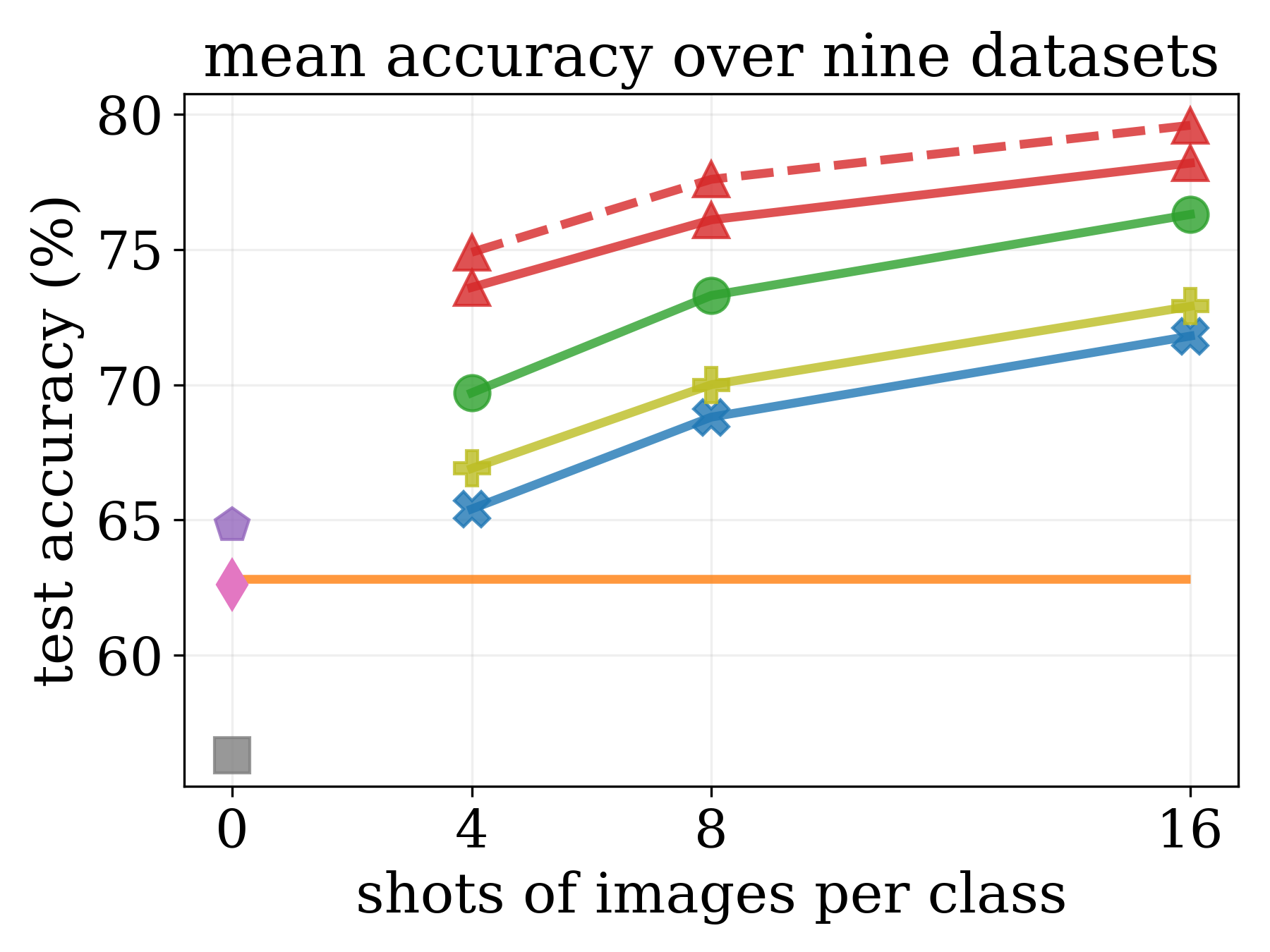}
    &
    \includegraphics[width=1.0\linewidth, clip=true,trim = 0mm 0mm 0mm 0mm]{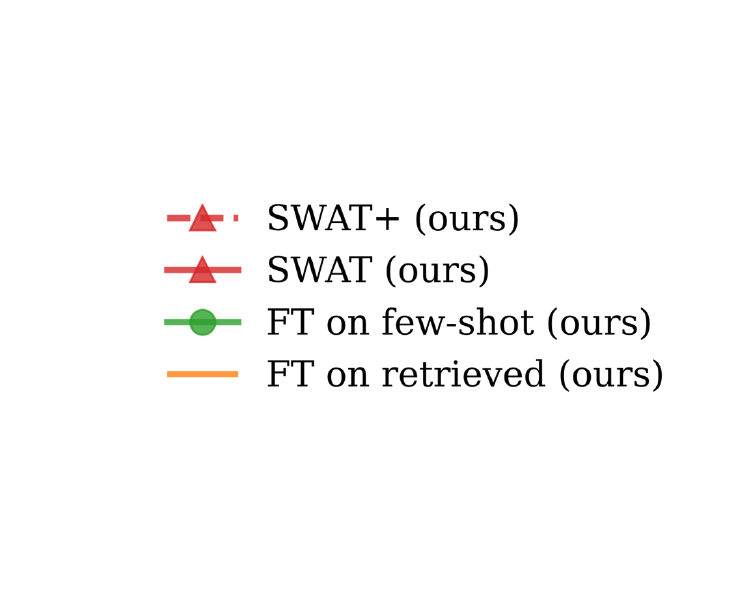}    
    &
    \includegraphics[width=1.0\linewidth, clip=true,trim = 0mm 0mm 0mm 0mm]{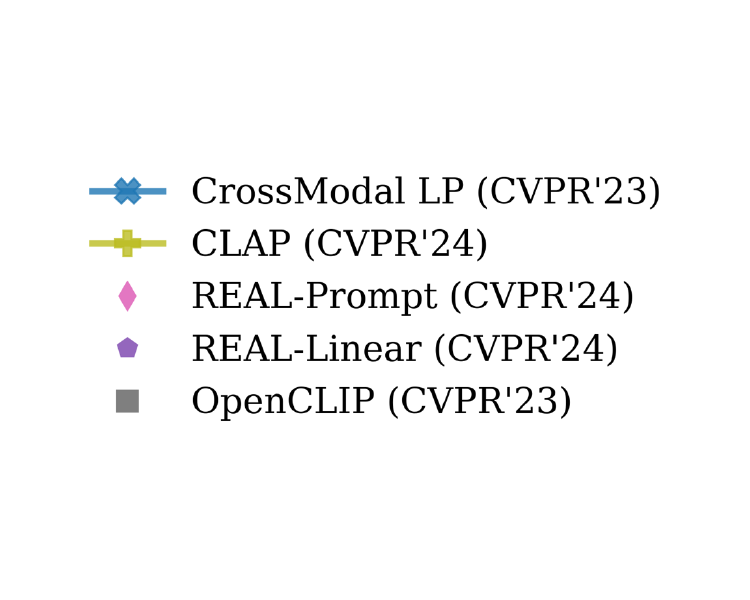}    
        
    \end{tabular}
    \vspace{-6mm}
    \caption{\small {\bf Comparison of SWAT with state-of-the-art zero-shot and few-shot methods.} 
    We show that simply finetuning the whole visual encoder on few-shot data (our few-shot finetuning, \textcolor{darkgreen}{green line}) outperforms previous FSR methods while finetuning on retrieved data (\textcolor{orange}{orange line}) underperforms zero-shot methods (e.g., ImageNet, EuroSAT, Food, DTD, and Stanford Cars) due to the large domain gap and imbalanced distributions of retrieved data.
    Our SWAT (\textcolor{red}{red line}) outperforms previous methods by $>$6\% w.r.t accuracy over nine datasets, with significant improvements (20-30\%) on challenging datasets like Semi-Aves and Aircraft.
    The results validate the effectiveness of our SWAT in mitigating the domain gap and imbalanced distribution issues. 
    We also show that our SWAT+ (\textcolor{red}{red dashed line}) which finetunes both visual encoder and classifier on few-shot data in stage 2 improves further over SWAT (cf. Section~\ref{sec:swat+}).
    Detailed performance on each dataset is provided in Table~\ref{tab:compare_sota_detail}.
    For Flowers, EuroSAT, DTD, and Stanford Cars datasets, we show that SWAT can be further improved by 1-6\% of accuracy with proper filtering on the retrieved data (cf. Table~\ref{tab:dtd_cars}).
    }
% \vspace{-3mm}
\label{fig:compare_sota}
\end{figure*}

\section{Hyperparameter Setting}
\label{sec:hyperparams}

{\bf Stage-1 End-to-End Finetuning.}
We follow previous work in other lines~\cite{parashar2024neglected, lin2023multimodality, kumar2022fine} to set hyperparameters in our work.
Specifically, 
for stage-1 end-to-end finetuning of SWAT, we follow
suggestions from~\cite{kumar2022fine, wortsman2022robust} to use a smaller learning rate (1e-6) for updating the visual encoder and a larger learning rate (1e-4) for the linear classifier. 
We initialize the classifier weights using the text embedding following \cite{parashar2024neglected} (cf. Table~\ref{tab:ablate_cls_init}). 
For other hyperparameters, 
we adopt the values reported in \cite{parashar2024neglected, lin2023multimodality} which include the AdamW optimizer, a batch size of 32, weight decay of 1e-2, and a cosine-annealing learning rate schedule with 50 warm-up iterations. 
We do not do early stopping as we strictly follow the validation-free protocol \cite{clap24}. Instead, we train for 50 epochs. The only exception is for ImageNet, we train for 10 epochs due to the large amount of retrieved data for its 1,000 classes.
The temperature factor is learned during the finetuning process with an initial learning rate of 1e-4 and the same cosine-annealing learning rate schedule. 
For data augmentation, we mix retrieved data with few-shot data using CutMix~\cite{yun2019cutmix}, following~\cite{park2022majority} to sample the mixing ratio from a uniform distribution ($\alpha =1.0$ for beta distribution) and apply CutMix with a probability of 0.5. Our few-shot finetuning (FSFT) adopts the same set of training recipe.

{\bf Stage-2 Classifier Retraining.}
We use the same set of hyperparameters and follow the practice in~\cite{parashar2024neglected} to train for 10 epochs with a fixed temperature of 0.01. 
We initialize the classifier in stage 2 using the learned classifier weights from stage-1 end-to-end finetuning, following \cite{kumar2022fine}.

{
\setlength{\tabcolsep}{0.15em} 
\begin{table*}[t]
\small
\centering
\caption{\small
\textbf{Detailed comparison of our SWAT and few-shot finetuning (FSFT) with state-of-the-art zero-shot and few-shot recognition methods using OpenCLIP ViT/B-32 model.}
SWAT significantly outperforms previous few-shot recognition methods by 6\% across nine datasets.
We also include the results of FSFT with and without CutMix, as well as SWAT+ where we finetuning the whole model rather than only the classifier on few-shot data in the second stage (cf. Section~\ref{sec:swat+}).
We highlight the best number in \textbf{bold} and underline \underline{the second best}. 
\textcolor{Green}{Superscripts} mark improvements compared to previous state-of-the-art FSR method CLAP~\cite{clap24}.
}
\vspace{-3mm}
\label{tab:compare_sota_detail}
\scalebox{0.85}{
\begin{tabular}{llllllllllll!{\vrule}l}
\toprule
shots & strategy & methods & Semi-Aves & Flowers & Aircraft & EuroSAT & DTD & Pets & Food & Cars & ImageNet & average \\ 
\midrule
\multirow{3}{*}{0} & \multirow{2}{*}{prompting} & OpenCLIP~\cite{cherti2023reproducible} & 8.4 & 68.2 & 17.1 & 51.1 & 53.5 & 88.7 & 77.2 & 79.2 & 63.0 & 56.3 \\
 &  & REAL-Prompt~\cite{parashar2024neglected} & 43.4 & 76.0 & 18.0 & 56.9 & 59.2 & 88.7 & 77.1 & 80.6 & 63.6 & 62.6 \\
 &retrieval-augmented  & REAL-Linear~\cite{parashar2024neglected} & 49.2 & 79.4 & 27.3 & 51.5 & 61.0 & 89.7 & 78.0 & 81.7 & 65.5 & 64.8 \\ 
 
 \midrule
 
\multirow{12}{*}{4} & \multirow{2}{*}{prompt-learning} & CoOp~\cite{zhou2022learning} & 38.1 & 86.1 & 20.6 & 68.6 & 53.9 & 86.7 & 73.5 & 62.7 & 58.5 & 61.0 \\
 &  & PLOT~\cite{chen2022plot} & 37.2 & 87.8 & 22.4 & 72.4 & 56.0 & 88.6 & 77.2 & 63.4 & 61.5 & 62.9 \\ 
 \cmidrule{2-13} 
 
 & \multirow{6}{*}{adapter-based} & CLIP-Adapter~\cite{clipadapter} & 39.2 & 85.3 & 23.0 & 72.5 & 47.2 & 80.0 & 72.1 & 61.0 & 55.7 & 59.6 \\
 &  & TIP-Adapter~\cite{tipadapter} & 37.4 & 69.8 & 19.6 & 54.3 & 53.5 & 82.3 & 74.7 & 57.7 & 60.2 & 56.6 \\
 &  & TIP-Adapter (f)~\cite{tipadapter} & 42.4 & 74.4 & 21.9 & 66.8 & 58.0 & 85.5 & 75.3 & 61.1 & 61.5 & 60.8 \\
 &  & TaskRes(e)~\cite{taskres} & 43.2 & 89.4 & 25.9 & 73.0 & 58.4 & 84.6 & 74.5 & 64.7 & 58.0 & 63.5 \\
 &  & CrossModal-LP~\cite{lin2023multimodality} & 29.1 & 88.9 & 25.1 & 74.8 & 62.2 & 88.3 & {76.7} & 80.7 & 63.2 & 65.4 \\
 &  & CLAP~\cite{clap24} & 34.0 & 90.1 & 28.0 & 74.7 & 63.0 & 87.0 & {76.7} & \textbf{84.9} & {64.0} & 66.9 \\
 \cmidrule{2-13} 
 
 & \multirow{4}{*}{finetuning-based} 
 &\cellcolor{gray!15}  FSFT (ours) 
 &\cellcolor{gray!15} 47.5$^{\textcolor{Green}{+13.5}}$ 
 &\cellcolor{gray!15} \underline{92.5}$^{\textcolor{Green}{+1.4}}$
 &\cellcolor{gray!15} 27.9$^{\textcolor{Red}{-0.1}}$ 
 &\cellcolor{gray!15} 81.6$^{\textcolor{Green}{+6.9}}$ 
 &\cellcolor{gray!15} \underline{66.6}$^{\textcolor{Green}{+3.6}}$
 &\cellcolor{gray!15} 88.7$^{\textcolor{Green}{+1.7}}$ 
 &\cellcolor{gray!15} 75.8$^{\textcolor{Red}{-0.9}}$ 
 &\cellcolor{gray!15} 81.5$^{\textcolor{Red}{-3.4}}$ 
 &\cellcolor{gray!15} 62.3$^{\textcolor{Red}{-1.7}}$ 
 &\cellcolor{gray!15} 69.4$^{\textcolor{Green}{+2.5}}$ \\
 
 &  &\cellcolor{gray!15} FSFT w/ CutMix (ours) 
 &\cellcolor{gray!15} {48.0}$^{\textcolor{Green}{+14.0}}$ 
 &\cellcolor{gray!15} {92.2}$^{\textcolor{Green}{+1.1}}$
 &\cellcolor{gray!15} {28.8}$^{\textcolor{Green}{+0.8}}$ 
 &\cellcolor{gray!15} {81.8}$^{\textcolor{Green}{+7.1}}$
 &\cellcolor{gray!15} \textbf{66.7$^{\textcolor{Green}{+3.7}}$}
 &\cellcolor{gray!15} {89.0}$^{\textcolor{Green}{+2.0}}$ 
 &\cellcolor{gray!15} 76.1$^{\textcolor{Red}{-0.6}}$ 
 &\cellcolor{gray!15} {82.5}$^{\textcolor{Red}{-2.4}}$
 &\cellcolor{gray!15} 62.4$^{\textcolor{Red}{-1.6}}$ 
 &\cellcolor{gray!15} {69.7}$^{\textcolor{Green}{+2.8}}$ \\
 
 &  &\cellcolor{gray!15}  SWAT (ours) 
 &\cellcolor{gray!15} \underline{58.5}$^{\textcolor{Green}{+24.5}}$ 
 &\cellcolor{gray!15} 90.6$^{\textcolor{Green}{+0.5}}$ 
 &\cellcolor{gray!15} \textbf{55.7$^{\textcolor{Green}{+27.7}}$} 
 &\cellcolor{gray!15} \underline{83.2}$^{\textcolor{Green}{+8.5}}$
 &\cellcolor{gray!15} 58.3$^{\textcolor{Red}{-4.7}}$ 
 &\cellcolor{gray!15} \underline{91.3}$^{\textcolor{Green}{+4.3}}$
 &\cellcolor{gray!15} \underline{77.3}$^{\textcolor{Green}{+0.6}}$ 
 &\cellcolor{gray!15} 81.1$^{\textcolor{Red}{-3.8}}$ 
 &\cellcolor{gray!15} \underline{65.8}$^{\textcolor{Green}{+1.8}}$
 &\cellcolor{gray!15} \underline{73.5}$^{\textcolor{Green}{+6.6}}$ \\

 &  &\cellcolor{gray!15} SWAT+ (ours) 
 &\cellcolor{gray!15} \textbf{59.9$^{\textcolor{Green}{+25.9}}$} 
 &\cellcolor{gray!15} \textbf{94.2$^{\textcolor{Green}{+4.1}}$} 
 &\cellcolor{gray!15} \underline{55.6}$^{\textcolor{Green}{+27.6}}$ 
 &\cellcolor{gray!15} \textbf{83.4$^{\textcolor{Green}{+8.7}}$} 
 &\cellcolor{gray!15} 61.5$^{\textcolor{Red}{-1.5}}$ 
 &\cellcolor{gray!15} \textbf{91.6$^{\textcolor{Green}{+4.6}}$} 
 &\cellcolor{gray!15} \textbf{77.9$^{\textcolor{Green}{+1.2}}$} 
 &\cellcolor{gray!15} \underline{83.7}$^{\textcolor{Red}{-1.2}}$ 
 &\cellcolor{gray!15} \textbf{66.6$^{\textcolor{Green}{+2.6}}$} 
 &\cellcolor{gray!15} \textbf{74.9$^{\textcolor{Green}{+8.0}}$}
 \\

 \midrule
 
\multirow{12}{*}{8} & \multirow{2}{*}{prompt-learning} & CoOp~\cite{zhou2022learning} & 42.0 & 91.3 & 26.6 & 77.1 & 59.7 & 85.4 & 71.6 & 67.6 & 60.4 & 64.6 \\
 &  & PLOT~\cite{chen2022plot} & 41.4 & 92.4 & 26.2 & 78.2 & 61.7 & 87.4 & 75.3 & 67.0 & 61.9 & 65.7 \\
 \cmidrule{2-13} 
 & \multirow{6}{*}{adapter-based} & CLIP-Adapter~\cite{clipadapter} & 41.2 & 91.9 & 27.9 & 78.5 & 61.4 & 83.4 & 72.1 & 66.8 & 57.0 & 64.5 \\
 &  & TIP-Adapter~\cite{tipadapter} & 39.8 & 73.8 & 19.4 & 62.3 & 51.5 & 82.3 & 73.9 & 57.6 & 59.4 & 57.8 \\
 &  & TIP-Adapter (f)~\cite{tipadapter} & 46.2 & 84.3 & 23.8 & 70.3 & 59.8 & 85.6 & 75.0 & 64.4 & 61.8 & 63.5 \\
 &  & TaskRes(e)~\cite{taskres} & 47.1 & 94.3 & 30.9 & 78.8 & 63.5 & 85.7 & 74.4 & 69.7 & 59.1 & 67.1 \\
 &  & CrossModal-LP~\cite{lin2023multimodality} & 38.8 & 92.5 & 27.9 & 80.6 & 67.2 & 88.8 & {77.3} & 82.7 & 63.1 & 68.8 \\
 &  & CLAP~\cite{clap24} & 42.9 & 92.9 & 33.6 & 77.4 & 66.4 & 87.8 & \underline{77.5} & \underline{86.1} &{65.6} & 70.0 \\
 \cmidrule{2-13} 
 
 & \multirow{4}{*}{finetuning-based} 
 &\cellcolor{gray!15}  FSFT (ours) 
 &\cellcolor{gray!15} 51.2$^{\textcolor{Green}{+8.3}}$ 
 &\cellcolor{gray!15} \underline{95.4}$^{\textcolor{Green}{+2.5}}$
 &\cellcolor{gray!15} 33.1$^{\textcolor{Red}{-0.5}}$ 
 &\cellcolor{gray!15} \textbf{90.3$^{\textcolor{Green}{+12.9}}$} 
 &\cellcolor{gray!15} \textbf{71.0$^{\textcolor{Green}{+4.6}}$} 
 &\cellcolor{gray!15} 89.3$^{\textcolor{Green}{+1.5}}$ 
 &\cellcolor{gray!15} 76.0$^{\textcolor{Red}{-1.5}}$ 
 &\cellcolor{gray!15} 83.5$^{\textcolor{Red}{-2.6}}$ 
 &\cellcolor{gray!15} 64.4$^{\textcolor{Red}{-1.2}}$ 
 &\cellcolor{gray!15} 72.7$^{\textcolor{Green}{+2.7}}$ \\
 
 &  &\cellcolor{gray!15}  FSFT w/ CutMix (ours) 
 &\cellcolor{gray!15} {52.3}$^{\textcolor{Green}{+9.4}}$ 
 &\cellcolor{gray!15} {95.2}$^{\textcolor{Green}{+2.3}}$ 
 &\cellcolor{gray!15} {35.4}$^{\textcolor{Green}{+1.8}}$ 
 &\cellcolor{gray!15} {89.4}$^{\textcolor{Green}{+12.0}}$ 
 &\cellcolor{gray!15} \underline{70.6}$^{\textcolor{Green}{+4.2}}$ 
 &\cellcolor{gray!15} {89.6}$^{\textcolor{Green}{+1.8}}$ 
 &\cellcolor{gray!15} 77.0$^{\textcolor{Red}{-0.5}}$ 
 &\cellcolor{gray!15} {85.3}$^{\textcolor{Red}{-0.8}}$ 
 &\cellcolor{gray!15} 64.8$^{\textcolor{Red}{-0.8}}$ 
 &\cellcolor{gray!15} {73.3}$^{\textcolor{Green}{+3.3}}$ \\
 
 &  &\cellcolor{gray!15}  SWAT (ours) 
 &\cellcolor{gray!15} \underline{60.8}$^{\textcolor{Green}{+17.9}}$ 
 &\cellcolor{gray!15} 94.1$^{\textcolor{Green}{+1.2}}$ 
 &\cellcolor{gray!15} \textbf{59.1$^{\textcolor{Green}{+25.5}}$}
 &\cellcolor{gray!15} 89.2$^{\textcolor{Green}{+11.8}}$ 
 &\cellcolor{gray!15} 62.6$^{\textcolor{Red}{-3.8}}$ 
 &\cellcolor{gray!15} \underline{90.8}$^{\textcolor{Green}{+3.0}}$ 
 &\cellcolor{gray!15} \underline{77.5}$^{\textcolor{Gray}{+0.0}}$ 
 &\cellcolor{gray!15} 83.5$^{\textcolor{Red}{-2.6}}$ 
 &\cellcolor{gray!15} \underline{66.6}$^{\textcolor{Green}{+1.0}}$ 
 &\cellcolor{gray!15} \underline{76.0}$^{\textcolor{Green}{+6.0}}$ \\

 &  &\cellcolor{gray!15} SWAT+ (ours) 
 &\cellcolor{gray!15} \textbf{62.7$^{\textcolor{Green}{+19.8}}$} 
 &\cellcolor{gray!15} \textbf{96.7$^{\textcolor{Green}{+3.8}}$} 
 &\cellcolor{gray!15} \underline{56.8}$^{\textcolor{Green}{+23.2}}$ 
 &\cellcolor{gray!15} \underline{89.7}$^{\textcolor{Green}{+12.3}}$ 
 &\cellcolor{gray!15} 67.0$^{\textcolor{Green}{+0.6}}$ 
 &\cellcolor{gray!15} \textbf{91.9$^{\textcolor{Green}{+4.1}}$} 
 &\cellcolor{gray!15} \textbf{78.4$^{\textcolor{Green}{+0.9}}$} 
 &\cellcolor{gray!15} \textbf{87.0$^{\textcolor{Green}{+0.9}}$} 
 &\cellcolor{gray!15} \textbf{68.1$^{\textcolor{Green}{+2.5}}$} 
 &\cellcolor{gray!15} \textbf{77.6$^{\textcolor{Green}{+7.6}}$} \\

\midrule
 
\multirow{12}{*}{16} & \multirow{2}{*}{prompt-learning} & CoOp~\cite{zhou2022learning} & 46.1 & 94.5 & 31.4 & 83.7 & 62.5 & 87.0 & 74.5 & 73.6 & 61.9 & 68.4 \\
 &  & PLOT~\cite{chen2022plot} & 44.4 & 94.8 & 31.5 & 82.2 & 65.6 & 87.2 & 77.1 & 72.8 & 63.0 & 68.7 \\
 \cmidrule{2-13} 
 & \multirow{6}{*}{adapter-based} & CLIP-Adapter~\cite{clipadapter} & 43.6 & 94.6 & 34.2 & 83.2 & 65.7 & 84.9 & 74.0 & 73.5 & 59.0 & 68.1 \\
 &  & TIP-Adapter~\cite{tipadapter} & 42.0 & 78.4 & 22.0 & 67.9 & 54.8 & 81.1 & 73.0 & 58.8 & 57.8 & 59.5 \\
 &  & TIP-Adapter (f)~\cite{tipadapter} & 50.1 & 91.2 & 29.3 & 76.6 & 64.6 & 85.4 & 74.7 & 69.6 & 62.3 & 67.1 \\
 &  & TaskRes(e)~\cite{taskres} & 48.5 & 96.1 & 36.5 & 83.7 & 65.9 & 86.3 & 75.4 & 75.4 & 60.9 & 69.9 \\
 &  & CrossModal-LP~\cite{lin2023multimodality} & 46.8 & 95.5 & 32.4 & 85.2 & 71.9 & 89.1 & 77.5 & 84.7 & 63.1 & 71.8 \\
 &  & CLAP~\cite{clap24} & 49.2 & 94.8 & 39.1 & 81.7 & 69.9 & 88.4 & \underline{78.5} & \underline{87.8} & {67.1} & 72.9 \\
 \cmidrule{2-13} 
 
 & \multirow{4}{*}{finetuning-based} 
 &\cellcolor{gray!15} FSFT (ours) 
 &\cellcolor{gray!15} 55.3$^{\textcolor{Green}{+6.1}}$ 
 &\cellcolor{gray!15} {97.0}$^{\textcolor{Green}{+2.2}}$ 
 &\cellcolor{gray!15} 37.0$^{\textcolor{Red}{-2.1}}$
 &\cellcolor{gray!15} \underline{94.0}$^{\textcolor{Green}{+12.3}}$ 
 &\cellcolor{gray!15} \underline{73.3}$^{\textcolor{Green}{+3.4}}$ 
 &\cellcolor{gray!15} 89.5$^{\textcolor{Green}{+1.1}}$ 
 &\cellcolor{gray!15} 77.1$^{\textcolor{Red}{-1.4}}$ 
 &\cellcolor{gray!15} {85.7}$^{\textcolor{Red}{-2.1}}$ 
 &\cellcolor{gray!15} 66.7$^{\textcolor{Red}{-0.4}}$ 
 &\cellcolor{gray!15} 75.1$^{\textcolor{Green}{2.2}}$ \\
 
 &  &\cellcolor{gray!15}  FSFT w/ CutMix (ours) 
 &\cellcolor{gray!15} {56.5}$^{\textcolor{Green}{+7.3}}$ 
 &\cellcolor{gray!15} \underline{97.1}$^{\textcolor{Green}{+2.3}}$ 
 &\cellcolor{gray!15} {42.7}$^{\textcolor{Green}{+3.6}}$ 
 &\cellcolor{gray!15} \textbf{94.3$^{\textcolor{Green}{+12.6}}$} 
 &\cellcolor{gray!15} \textbf{73.4$^{\textcolor{Green}{+3.5}}$} 
 &\cellcolor{gray!15} {89.6}$^{\textcolor{Green}{+1.2}}$ 
 &\cellcolor{gray!15} 78.2$^{\textcolor{Red}{-0.3}}$ 
 &\cellcolor{gray!15} \underline{87.8}$^{\textcolor{Gray}{+0.0}}$ 
 &\cellcolor{gray!15} 66.9$^{\textcolor{Red}{-0.2}}$ 
 &\cellcolor{gray!15} {76.3}$^{\textcolor{Green}{+3.4}}$ \\
 
 &  &\cellcolor{gray!15}  SWAT (ours) 
 &\cellcolor{gray!15} \underline{63.1}$^{\textcolor{Green}{+13.9}}$ 
 &\cellcolor{gray!15} 96.4$^{\textcolor{Green}{+1.6}}$ 
 &\cellcolor{gray!15} \textbf{62.4$^{\textcolor{Green}{+23.3}}$} 
 &\cellcolor{gray!15} 92.6$^{\textcolor{Green}{+10.9}}$ 
 &\cellcolor{gray!15} 66.3$^{\textcolor{Red}{-3.6}}$ 
 &\cellcolor{gray!15} \underline{91.6}$^{\textcolor{Green}{+3.2}}$ 
 &\cellcolor{gray!15} {78.3}$^{\textcolor{Red}{-0.2}}$ 
 &\cellcolor{gray!15} 85.4$^{\textcolor{Red}{-2.4}}$ 
 &\cellcolor{gray!15} \underline{67.6}$^{\textcolor{Green}{+0.5}}$
 &\cellcolor{gray!15} \underline{78.2}$^{\textcolor{Green}{+5.3}}$ \\

 &  &\cellcolor{gray!15} SWAT+ (ours) 
 &\cellcolor{gray!15} \textbf{64.7$^{\textcolor{Green}{+5.5}}$} 
 &\cellcolor{gray!15} \textbf{98.3$^{\textcolor{Green}{+3.5}}$} 
 &\cellcolor{gray!15} \underline{60.2}$^{\textcolor{Green}{+21.1}}$ 
 &\cellcolor{gray!15} 93.5$^{\textcolor{Green}{+11.8}}$ 
 &\cellcolor{gray!15} 69.8$^{\textcolor{Red}{-0.1}}$ 
 &\cellcolor{gray!15} \textbf{92.2$^{\textcolor{Green}{+3.8}}$} 
 &\cellcolor{gray!15} \textbf{79.1$^{\textcolor{Green}{+0.6}}$} 
 &\cellcolor{gray!15} \textbf{89.2$^{\textcolor{Green}{+1.4}}$} 
 &\cellcolor{gray!15} \textbf{69.3$^{\textcolor{Green}{+2.2}}$} 
 &\cellcolor{gray!15} \textbf{79.6$^{\textcolor{Green}{+6.7}}$}
 \\

 \bottomrule
\end{tabular}
}
% \vspace{-1mm}
\end{table*}
}

{\bf Baselines.}
For baseline methods, we reimplement CrossModal Linear Probing~\cite{lin2023multimodality} using the same hyperparameters as in stage-2 classifier retraining and training for 50 epochs. We obtain the results of CLAP~\cite{clap24} using OpenCLIP models with its default hyperparameters. For other baseline methods that originally used an unrealistically large validation set for hyperparameter tuning, we copy their results from \cite{clap24}.

{
\setlength{\tabcolsep}{0.15em} 
\begin{table*}[t]
\small
\centering
\caption{\small \textbf{Detailed comparison of SWAT with state-of-the-art zero-shot and few-shot recognition methods using OpenCLIP ViT-B/16 model.}
Results show that SWAT achieves larger performance gains ($\sim$8\%) over CLAP~\cite{clap24} with a larger backbone of ViT-B/16.
We also include the results of FSFT with and without CutMix, as well as SWAT+ where we finetuning the whole model rather than only the classifier on few-shot data in the second stage (cf. Section~\ref{sec:swat+}).
We highlight the best number in \textbf{bold} and \underline{underline} the second best.
\textcolor{Green}{Superscripts} mark improvements compared to previous state-of-the-art CLAP~\cite{clap24}.
}
\vspace{-2mm}
\scalebox{0.85}{
\begin{tabular}{llllllllllll!{\vrule}l}

\toprule
shots & strategy & methods & Semi-Aves & Flowers & Aircraft & EuroSAT & DTD & Pets & Food & Cars & ImageNet & average \\
\midrule
\multirow{3}{*}{0} & \multirow{2}{*}{prompting} & OpenCLIP~\cite{cherti2023reproducible} & 8.5 & 68.3 & 17.9 & 50.1 & 49.2 & 91.0 & 82.7 & 83.6 & 67.2 & 57.6 \\
 &  & REAL-Prompt~\cite{parashar2024neglected} & 51.2 & 76.0 & 19.4 & 51.2 & 56.7 & 91.0 & 82.8 & 84.4 & 67.6 & 64.5 \\
 & retrieval-augmented & REAL-Linear~\cite{parashar2024neglected} & 57.1 & 80.3 & 29.2 & 46.8 & 60.3 & 91.4 & 83.3 & 85.5 & 69.8 & 67.1 \\

\midrule

\multirow{6}{*}{4} & \multirow{2}{*}{adapter-based} & CrossModal-LP~\cite{lin2023multimodality} & 37.7 & 90.1 & 27.9 & 74.8 & 62.4 & 90.6 & 82.2 & 85.6 & {67.8} & 68.8 \\
 &  & CLAP~\cite{clap24} & 40.0 & 91.0 & 29.9 & 76.7 & 64.6 & 88.9 & 80.4 & {86.8} & 66.9 & 69.5 \\
 \cmidrule{2-13} 
 
 & \multirow{4}{*}{finetuning-based} &\cellcolor{gray!15} FSFT (ours) 
 &\cellcolor{gray!15} 57.7$^{\textcolor{Green}{+17.7}}$ 
 &\cellcolor{gray!15} {93.6}$^{\textcolor{Green}{+2.6}}$ 
 &\cellcolor{gray!15}\cellcolor{gray!15} 33.0$^{\textcolor{Green}{+3.1}}$ 
 &\cellcolor{gray!15} \textbf{85.5$^{\textcolor{Green}{+8.8}}$} 
 &\cellcolor{gray!15} \textbf{69.1$^{\textcolor{Green}{+4.5}}$} 
 &\cellcolor{gray!15} 91.4$^{\textcolor{Green}{+2.5}}$ 
 &\cellcolor{gray!15} 81.9$^{\textcolor{Green}{+1.5}}$ 
 &\cellcolor{gray!15} 86.1$^{\textcolor{Red}{-0.7}}$ 
 &\cellcolor{gray!15} 67.4$^{\textcolor{Green}{+0.5}}$ 
 &\cellcolor{gray!15} 74.0$^{\textcolor{Green}{+4.5}}$ \\
 
 &  &\cellcolor{gray!15} FSFT w/ CutMix (ours) 
 &\cellcolor{gray!15} {58.8}$^{\textcolor{Green}{+18.8}}$ 
 &\cellcolor{gray!15} 93.4$^{\textcolor{Green}{+2.4}}$ 
 &\cellcolor{gray!15} {33.4}$^{\textcolor{Green}{+3.5}}$ 
 &\cellcolor{gray!15} 83.4$^{\textcolor{Green}{+7.7}}$ 
 &\cellcolor{gray!15} \underline{68.6}$^{\textcolor{Green}{+4.2}}$ 
 &\cellcolor{gray!15} 91.8$^{\textcolor{Green}{+2.9}}$ 
 &\cellcolor{gray!15} {82.7}$^{\textcolor{Green}{+2.3}}$ 
 &\cellcolor{gray!15} \underline{87.0}$^{\textcolor{Green}{+0.2}}$ 
 &\cellcolor{gray!15} {67.8}$^{\textcolor{Green}{+0.9}}$ 
 &\cellcolor{gray!15} {74.1}$^{\textcolor{Green}{+4.6}}$ \\
 
 &  &\cellcolor{gray!15} SWAT (ours) 
 &\cellcolor{gray!15} \underline{69.2}$^{\textcolor{Green}{+29.2}}$ 
 &\cellcolor{gray!15} \underline{93.8}$^{\textcolor{Green}{+2.8}}$ 
 &\cellcolor{gray!15} \textbf{66.5$^{\textcolor{Green}{+36.6}}$} 
 &\cellcolor{gray!15} {84.2}$^{\textcolor{Green}{+8.5}}$ 
 &\cellcolor{gray!15} 62.6$^{\textcolor{Red}{-2.0}}$ 
 &\cellcolor{gray!15} \underline{92.9}$^{\textcolor{Green}{+4.0}}$ 
 &\cellcolor{gray!15} \underline{83.3}$^{\textcolor{Green}{+2.9}}$ 
 &\cellcolor{gray!15} 85.2$^{\textcolor{Red}{-1.6}}$ 
 &\cellcolor{gray!15} \underline{70.6}$^{\textcolor{Green}{+3.7}}$ 
 &\cellcolor{gray!15} \underline{78.7}$^{\textcolor{Green}{+9.2}}$ \\

 &  &\cellcolor{gray!15} SWAT+ (ours) 
 &\cellcolor{gray!15} \textbf{70.5$^{\textcolor{Green}{+30.5}}$} 
 &\cellcolor{gray!15} \textbf{96.0$^{\textcolor{Green}{+5.0}}$} 
 &\cellcolor{gray!15} \underline{64.5}$^{\textcolor{Green}{+34.6}}$ 
 &\cellcolor{gray!15} \underline{84.4}$^{\textcolor{Green}{+7.7}}$ 
 &\cellcolor{gray!15} 64.7$^{\textcolor{Green}{+0.1}}$ 
 &\cellcolor{gray!15} \textbf{93.4$^{\textcolor{Green}{+4.5}}$} 
 &\cellcolor{gray!15} \textbf{83.9$^{\textcolor{Green}{+3.5}}$} 
 &\cellcolor{gray!15} \textbf{88.5$^{\textcolor{Green}{+1.7}}$} 
 &\cellcolor{gray!15} \textbf{71.8$^{\textcolor{Green}{+4.9}}$} 
 &\cellcolor{gray!15} \textbf{79.7$^{\textcolor{Green}{+10.2}}$} \\

\midrule

\multirow{6}{*}{8} & \multirow{2}{*}{adapter-based} & CrossModal-LP~\cite{lin2023multimodality} & 49.4 & 93.6 & 32.5 & 81.8 & 67.8 & 90.9 & 82.9 & 87.4 & 68.0 & 72.7 \\
 &  & CLAP~\cite{clap24} & 49.1 & 93.4 & 36.1 & 79.0 & 67.7 & 89.6 & 81.5 & {88.4} & 68.5 & 72.6 \\
  \cmidrule{2-13} 
 & \multirow{3}{*}{finetuning-based} &\cellcolor{gray!15} FSFT (ours) 
 &\cellcolor{gray!15} 61.9$^{\textcolor{Green}{+12.8}}$ 
 &\cellcolor{gray!15} \underline{96.6}$^{\textcolor{Green}{+3.2}}$ 
 &\cellcolor{gray!15} 39.6$^{\textcolor{Green}{+3.5}}$ 
 &\cellcolor{gray!15} \textbf{90.9$^{\textcolor{Green}{+11.9}}$} 
 &\cellcolor{gray!15} \underline{73.3}$^{\textcolor{Green}{+6.6}}$
 &\cellcolor{gray!15} 91.4$^{\textcolor{Green}{+1.8}}$ 
 &\cellcolor{gray!15} 82.0$^{\textcolor{Green}{+0.5}}$ 
 &\cellcolor{gray!15} 87.8$^{\textcolor{Red}{-0.6}}$ 
 &\cellcolor{gray!15} 69.4$^{\textcolor{Green}{+0.9}}$ 
 &\cellcolor{gray!15} 77.0$^{\textcolor{Green}{+4.4}}$ \\
 
 &  &\cellcolor{gray!15} FSFT w/ CutMix (ours) 
 &\cellcolor{gray!15} {63.0}$^{\textcolor{Green}{+13.9}}$ 
 &\cellcolor{gray!15} 96.4$^{\textcolor{Green}{+3.0}}$ 
 &\cellcolor{gray!15} {42.9}$^{\textcolor{Green}{+6.8}}$ 
 &\cellcolor{gray!15} \underline{90.3}$^{\textcolor{Green}{+11.3}}$ 
 &\cellcolor{gray!15} \textbf{73.5$^{\textcolor{Green}{+6.8}}$} 
 &\cellcolor{gray!15} {92.1}$^{\textcolor{Green}{+2.5}}$ 
 &\cellcolor{gray!15} {83.2}$^{\textcolor{Green}{+1.7}}$ 
 &\cellcolor{gray!15} \underline{89.6}$^{\textcolor{Green}{+1.2}}$ 
 &\cellcolor{gray!15} {69.8}$^{\textcolor{Green}{+1.3}}$ 
 &\cellcolor{gray!15} {77.9}$^{\textcolor{Green}{+5.3}}$ \\
 
 &  &\cellcolor{gray!15} SWAT (ours) 
 &\cellcolor{gray!15} \underline{71.4}$^{\textcolor{Green}{+22.3}}$ 
 &\cellcolor{gray!15} {96.5}$^{\textcolor{Green}{+3.1}}$ 
 &\cellcolor{gray!15} \textbf{69.1$^{\textcolor{Green}{+33.0}}$} 
 &\cellcolor{gray!15} 88.8$^{\textcolor{Green}{+4.6}}$ 
 &\cellcolor{gray!15} 66.3$^{\textcolor{Red}{-1.4}}$ 
 &\cellcolor{gray!15} \underline{93.2}$^{\textcolor{Green}{+3.6}}$ 
 &\cellcolor{gray!15} \underline{83.8}$^{\textcolor{Green}{+0.5}}$ 
 &\cellcolor{gray!15} 87.2$^{\textcolor{Red}{-1.2}}$ 
 &\cellcolor{gray!15} \underline{71.5}$^{\textcolor{Green}{+3.0}}$ 
 &\cellcolor{gray!15} \underline{80.9}$^{\textcolor{Green}{+8.3}}$ \\

 &  &\cellcolor{gray!15} SWAT+ (ours) 
 &\cellcolor{gray!15} \textbf{73.2$^{\textcolor{Green}{+24.1}}$} 
 &\cellcolor{gray!15} \textbf{98.2$^{\textcolor{Green}{+4.8}}$} 
 &\cellcolor{gray!15} \underline{67.3}$^{\textcolor{Green}{+31.2}}$ 
 &\cellcolor{gray!15} 88.9$^{\textcolor{Green}{+9.9}}$ 
 &\cellcolor{gray!15} 68.5$^{\textcolor{Green}{+0.8}}$ 
 &\cellcolor{gray!15} \textbf{93.9$^{\textcolor{Green}{+4.3}}$} 
 &\cellcolor{gray!15} \textbf{84.3$^{\textcolor{Green}{+2.8}}$} 
 &\cellcolor{gray!15} \textbf{90.7$^{\textcolor{Green}{+2.3}}$} 
 &\cellcolor{gray!15} \textbf{73.2$^{\textcolor{Green}{+4.7}}$} 
 &\cellcolor{gray!15} \textbf{82.0$^{\textcolor{Green}{+9.4}}$} \\

\midrule

\multirow{6}{*}{16} & \multirow{2}{*}{adapter-based} & CrossModal-LP~\cite{lin2023multimodality} & 57.7 & 96.5 & 38.9 & 84.5 & 73.3 & 90.7 & 83.3 & 88.8 & 68.0 & 75.7 \\
 &  & CLAP~\cite{clap24} & 56.9 & 95.2 & 42.4 & 82.2 & 71.4 & 90.3 & 82.3 & {89.8} & 70.0 & 75.6 \\
  \cmidrule{2-13} 
 & \multirow{3}{*}{finetuning-based} &\cellcolor{gray!15} FSFT (ours)
 &\cellcolor{gray!15} 66.3$^{\textcolor{Green}{+9.4}}$ 
 &\cellcolor{gray!15} {98.0}$^{\textcolor{Green}{+2.8}}$ 
 &\cellcolor{gray!15} 45.6$^{\textcolor{Green}{+3.2}}$ 
 &\cellcolor{gray!15} \underline{94.1}$^{\textcolor{Green}{+11.9}}$ 
 &\cellcolor{gray!15} \underline{75.8}$^{\textcolor{Green}{+4.4}}$ 
 &\cellcolor{gray!15} 91.5$^{\textcolor{Green}{+1.2}}$ 
 &\cellcolor{gray!15} 82.5$^{\textcolor{Green}{+0.2}}$ 
 &\cellcolor{gray!15} 89.7$^{\textcolor{Red}{-0.1}}$ 
 &\cellcolor{gray!15} 70.2$^{\textcolor{Green}{+0.2}}$ 
 &\cellcolor{gray!15} 79.3$^{\textcolor{Green}{+3.7}}$ \\
 
 &  &\cellcolor{gray!15} FSFT w/ CutMix (ours) 
 &\cellcolor{gray!15} {67.3}$^{\textcolor{Green}{+10.4}}$ 
 &\cellcolor{gray!15} \underline{98.2}$^{\textcolor{Green}{+3.0}}$ 
 &\cellcolor{gray!15} {51.2}$^{\textcolor{Green}{+8.8}}$ 
 &\cellcolor{gray!15} \textbf{94.2$^{\textcolor{Green}{+12.0}}$} 
 &\cellcolor{gray!15} \textbf{76.1$^{\textcolor{Green}{+4.7}}$} 
 &\cellcolor{gray!15} {92.3}$^{\textcolor{Green}{+2.0}}$ 
 &\cellcolor{gray!15} {84.0}$^{\textcolor{Green}{+1.7}}$ 
 &\cellcolor{gray!15} \underline{91.3}$^{\textcolor{Green}{+1.5}}$ 
 &\cellcolor{gray!15} {72.1}$^{\textcolor{Green}{+2.1}}$ 
 &\cellcolor{gray!15} {80.7}$^{\textcolor{Green}{+5.1}}$ \\
 
 &  &\cellcolor{gray!15} SWAT (ours) 
 &\cellcolor{gray!15} \underline{73.9}$^{\textcolor{Green}{+17.0}}$ 
 &\cellcolor{gray!15} \underline{98.2}$^{\textcolor{Green}{+3.0}}$ 
 &\cellcolor{gray!15} \textbf{72.6$^{\textcolor{Green}{+30.2}}$} 
 &\cellcolor{gray!15} 93.0$^{\textcolor{Green}{+10.8}}$ 
 &\cellcolor{gray!15} 69.0$^{\textcolor{Red}{-2.4}}$ 
 &\cellcolor{gray!15} \underline{93.3}$^{\textcolor{Green}{+3.0}}$ 
 &\cellcolor{gray!15} \underline{84.4}$^{\textcolor{Green}{+2.1}}$ 
 &\cellcolor{gray!15} 89.0$^{\textcolor{Red}{-0.8}}$ 
 &\cellcolor{gray!15} \underline{72.3}$^{\textcolor{Green}{+2.3}}$ 
 &\cellcolor{gray!15} \underline{82.9}$^{\textcolor{Green}{+7.3}}$ \\

  &  &\cellcolor{gray!15} SWAT+ (ours) 
  &\cellcolor{gray!15} \textbf{75.0$^{\textcolor{Green}{+18.1}}$} 
  &\cellcolor{gray!15} \textbf{99.0$^{\textcolor{Green}{+3.8}}$} 
  &\cellcolor{gray!15} \underline{69.8}$^{\textcolor{Green}{+27.4}}$ 
  &\cellcolor{gray!15} 93.0$^{\textcolor{Green}{+10.8}}$ 
  &\cellcolor{gray!15} 72.5$^{\textcolor{Green}{+1.1}}$ 
  &\cellcolor{gray!15} \textbf{94.1$^{\textcolor{Green}{+3.8}}$} 
  &\cellcolor{gray!15} \textbf{85.0$^{\textcolor{Green}{+2.7}}$} 
  &\cellcolor{gray!15} \textbf{92.3$^{\textcolor{Green}{+2.5}}$} 
  &\cellcolor{gray!15} \textbf{74.2$^{\textcolor{Green}{+4.2}}$} 
  &\cellcolor{gray!15} \textbf{83.9$^{\textcolor{Green}{+8.3}}$} \\
  
 \bottomrule
 
\end{tabular}}
\label{tab:compare_sota_detail_vitb16}
\end{table*}
}

\section{Detailed Benchmarking Results}
\label{sec:swat-compare}

We compare our SWAT and few-shot finetuning (FSFT) with prior state-of-the-art zero-shot~\cite{openclip, parashar2024neglected} and few-shot recognition methods~\cite{lin2023multimodality,clap24} using the OpenCLIP ViT-B/32 model in Fig.~\ref{fig:compare_sota} and list the detailed performance in Table~\ref{tab:compare_sota_detail}. We also include the performance of our few-shot finetuning without CutMix. Results show that SWAT outperforms previous FSR methods by $>$6\% accuracy over nine datasets, with substantial gains (20-30\%) on challenging datasets where prior FSR accuracy~\cite{lin2023multimodality, clap24} was below 50\% (e.g., Semi-Aves and Aircraft). Additionally, SWAT with OpenCLIP ViT-B/16 model (Table~\ref{tab:compare_sota_detail_vitb16}) yields even higher gains of 8\% over \cite{clap24} across nine datasets.

{\bf Further Analysis.}
Our experiments show that SWAT underperforms prior state-of-the-art FSR method \cite{clap24} on DTD and Stanford Cars. We conjecture that this is due to the significant domain gaps in the retrieved data, finetuning on which could hurt the model's performance. 
This motivates us to apply a filtering technique to remove excessively out-of-domain retrieved data.
Indeed, as shown in Table~\ref{tab:dtd_cars}, applying proper filtering on the retrieved data significantly boosts the performance of SWAT, allowing it to outperform CLAP~\cite{clap24}. We also find filtering improves SWAT on other datasets, including Semi-Aves, Flowers, and EuroSAT (cf. Table~\ref{tab:retrieve_methods}). 

\begin{figure*}[t]
    \centering
    \small
    % \vspace{-3mm}
    \includegraphics[width=1.0\linewidth, clip=true,trim = 0mm 0mm 0mm 0mm]{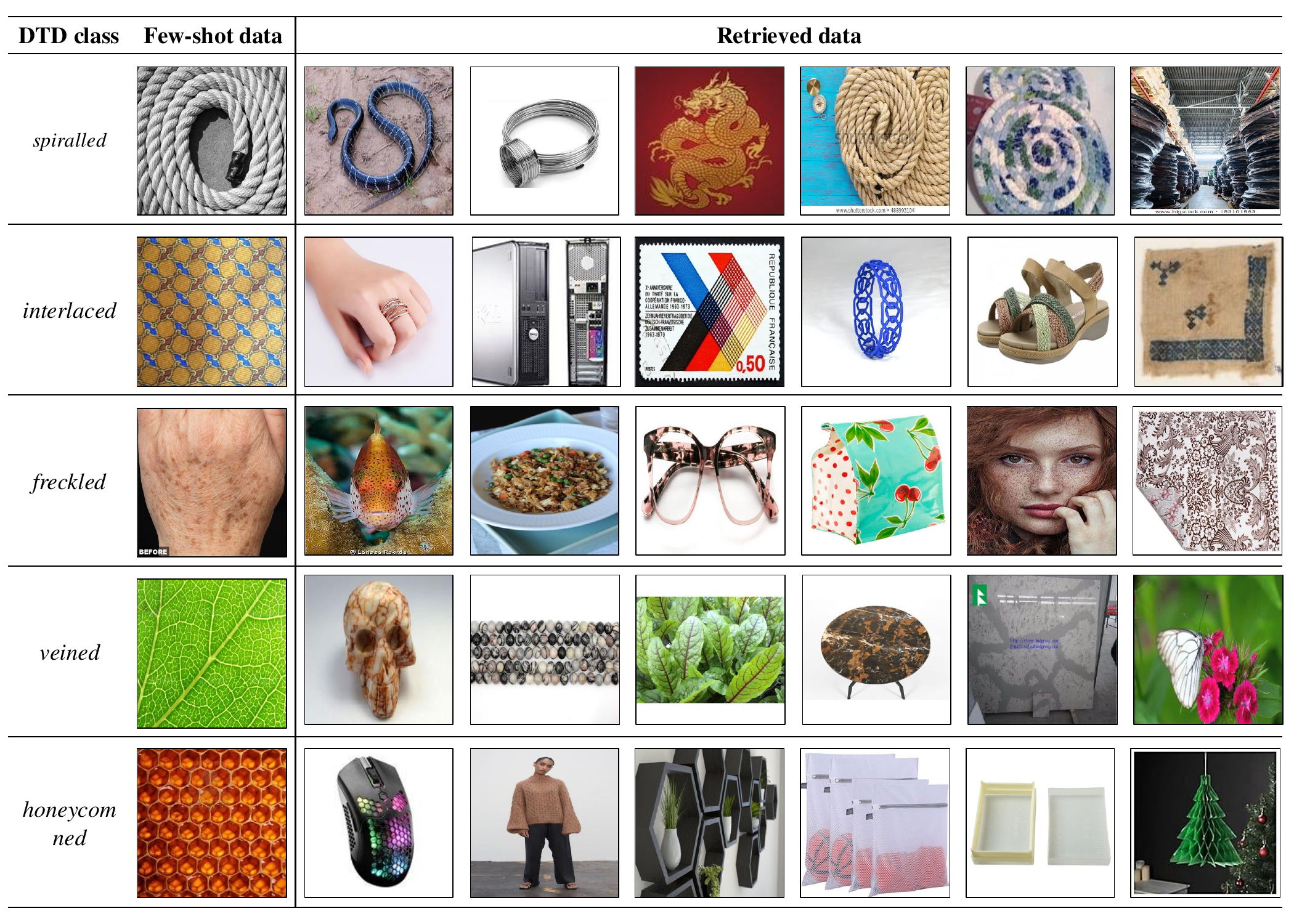}
    \vspace{-3mm}
    \caption{
    {\bf Visual comparison between downstream DTD images and the retrieved images (from LAION-400M~\cite{laion400m}) for various DTD concepts}. 
    Clearly, a large domain gap exists between the two data resources regarding styles, backgrounds, semantics, etc. 
    In addition, the retrieved images only have a partial region depicting the texture, contrasting to the few-shot images which are ``almost entirely filled with a texture'' according to DTD's strict data collection rules~\cite{dtd}.  
    We suggest future work to explore better retrieval methods that are closely aligned with downstream data distribution, e.g., by referring to the data collection/annotation rules provided in the data annotation guidelines of a downstream task.
    }
% \vspace{-3mm}
\label{fig:dtd_retrieved_compare}
\end{figure*}

Moreover, the improved SWAT still underperforms our few-shot finetuning (FSFT) on DTD datasets. We hypothesize that the discrepancy is because of DTD's strict data collection rules, which include only images that are almost entirely filled with a texture~\cite{dtd}.
In contrast, the retrieved images often have only part of the region depicting the texture (Fig.~\ref{fig:retrived_imgs_more} and \ref{fig:dtd_retrieved_compare}). 
This suggests future work to explore better retrieval or filtering methods to find images that are better aligned with downstream distribution, e.g., by referring to the data collection rules provided in the data annotation guidelines. We explore different retrieval methods in Section~\ref{sec:retrieval_methods} below.

\begin{table}[]
\caption{\small
\textbf{Comparison of SWAT's performance with prior state-of-the-art FSR method CLAP~\cite{clap24}.} SWAT underperforms CLAP on DTD and Cars datasets due to the significant domain gaps. However, with proper filtering on retrieved images (by keeping the top-10 retrieved images for each class that are ranked by prompt-to-caption or T2T similarity and discarding others), SWAT outperforms CLAP. 
We show results of different retrieval sizes in Table~\ref{tab:ablate_retr_size_dataset}.
Subscripts mark the performance difference compared with CLAP.
}
\vspace{-2mm}
\label{tab:dtd_cars}
\scalebox{0.9}{
\begin{tabular}{lllll}
\toprule
dataset & methods & 4-shot & 8-shot & 16-shot \\
\midrule
\multirow{3}{*}{DTD} & CLAP~\cite{clap24} & 63.0 & 66.4 & 69.9 \\
 & SWAT & 58.3$^{\textcolor{Red}{-2.0}}$ & 62.6$^{\textcolor{Red}{-3.8}}$ & 66.3$^{\textcolor{Red}{-3.6}}$ \\
 & SWAT+filtering & 63.5$^{\textcolor{Green}{+0.5}}$ & 69.1$^{\textcolor{Green}{+2.7}}$ & 72.9$^{\textcolor{Green}{+3.0}}$ \\
 \midrule
 
\multirow{3}{*}{Cars} & CLAP~\cite{clap24} & 84.9 & 86.1 & 87.8 \\
 & SWAT& 81.1$^{\textcolor{Red}{-3.8}}$ & 83.5$^{\textcolor{Red}{-2.6}}$ & 85.4$^{\textcolor{Red}{-2.4}}$ \\
 & SWAT+filtering & 83.5$^{\textcolor{Red}{-1.4}}$ & 86.8$^{\textcolor{Green}{+0.7}}$ & 88.6$^{\textcolor{Green}{+0.8}}$ \\
 \bottomrule
 
\end{tabular}}
\end{table}

{
\setlength{\tabcolsep}{0.9em} 
\begin{table*}[]
\small
\centering
\caption{\small \textbf{Comparison of SWAT using different retrieval methods.}
We conduct experiments on six datasets by first conducting string matching following~\cite{parashar2024neglected} to download images whose captions contain any of the concepts' synonyms, then ranking the images using different text (few-shot concepts or database captions) and image (database images or few-shot images) features for selecting the images most relevant to downstream concepts. The top-ranking 500 images for each class are selected for running SWAT with 16 few-shot data.
Results show that despite all methods outperforming random sampling by $<$1\% in average accuracy, no single method is the best for all datasets.
We highlight the best number in \textbf{bold} and \underline{underline} the second best. 
We further explore adding text-to-image filtering before text-to-text ranking to remove noisy images with image-to-FS-concept similarity of less than 0.25. Results show that T2I filtering improves SWAT's performance significantly, especially for the DTD dataset (6\% improvement). 
We show examples of T2I filtered images in Fig.~\ref{fig:filtered_imgs}.
}
\vspace{-2mm}
\label{tab:retrieve_methods}
\scalebox{1.0}{
\begin{tabular}{lcccccc!{\vrule}c}
\toprule
retrieval/ranking method & Semi-Aves & Flowers & Aircraft & EuroSAT & DTD & Cars & average \\
\midrule
random sampling & 62.8 & 96.0 & 62.2 & 92.6 & 64.9 & 84.7 & 77.2 \\
text-to-text: FS-concept \& DB-caption & \textbf{63.4} & 96.4 & 62.7 & \underline{93.0} & \underline{65.8} & \underline{85.4} & \underline{77.8} \\
image-to-image: FS-image \& DB-image & 63.0 & \textbf{97.1} & \underline{62.8} & 92.7 & 64.9 & 84.9 & 77.6 \\
image-to-text: FS-image \& DB-caption & \underline{63.2} & \underline{96.8} & \underline{62.8} & \textbf{93.4} & \textbf{66.7} & \textbf{86.9} & \textbf{78.3} \\
image-to-text: FS-concept \& DB-image & 62.9 & \underline{96.8} & \textbf{63.3} & \textbf{93.4} & 65.7 & 83.7 & 77.6 \\
\midrule
text-to-image filtering (0.25) + text-to-text & 63.8 &97.5 & 62.6 & 93.7 & 71.6 &85.8 & 79.2 \\
\bottomrule
\end{tabular}}
\end{table*}
}

\section{Analysis of Retrieval and Filtering Methods}
\label{sec:retrieval_methods}

Retrieval-augmented learning has been extensively studied for zero-shot recognition~\cite{liu2023learning, wallingford2023neural, parashar2024neglected}. Previous work~\cite{liu2023learning} utilizes text-to-text (T2T) or text-to-image (T2I) similarity to retrieve images relevant to each downstream concept. However, as noted by \cite{parashar2024neglected}, such similarity-based retrieval requires significant storage for downloading all the source images (e.g., $>$10TB for LAION-400M) and high compute costs for computing image and text features (>250 T4 GPU hours). In addition, \cite{wallingford2023neural} points out the challenge of threshold selection in similarity-based retrieval: setting it too low includes irrelevant images, which can negatively impact training. Moreover, the proper threshold varies for different concepts, making it infeasible to search at scale.
Given the above limitations, in this study, we adopt the {\em string-matching-based retrieval} by \cite{parashar2024neglected}, detailed in the following two steps. 

{\bf Step 1: String Matching with Synonyms.}
We use string matching to retrieve images whose captions contain any of the downstream concepts' synonyms. This circumvents the large storage cost, as now we only need to download the text (60GB for LAION texts) for string matching and then the images with matching captions (50GB for all nine datasets). Additionally, \cite{parashar2024neglected} shows that using concept synonyms helps retrieve diverse images which benefits retrieval-augmented learning.

{\bf Step 2: Selection by Ranking.}
To select images that are most relevant to downstream concepts, we rank the retrieved images based on prompt-to-caption (T2T) similarities and select the top-ranking 500 images for each downstream concept. We compare different ranking methods using text (image captions or downstream concepts) and image (pretraining images or few-shot images) features in Table~\ref{tab:retrieve_methods}. The results show that, despite all ranking methods outperforming the random sampling, no single ranking method is the best across all datasets. This suggests future work to design retrieval methods customized to each downstream task.

{\bf T2I Filtering Improves SWAT Performance.}
To explore better retrieval methods, we follow the practice in the curation of the LAION dataset to apply text-to-image (T2I) filtering, excluding noisy retrieved images with T2I (few-shot concepts and retrieved images) similarities below 0.25. 
Despite that adding T2I filtering increases the imbalance of retrieved data, it notably improves SWAT's performance (cf. Table~\ref{tab:retrieve_methods}), especially on the DTD dataset ($>$6\%). We show examples of T2I-filtered noisy images in Fig.~\ref{fig:filtered_imgs}. 
This suggests future retrieval methods to explore better filtering techniques.
By default, our SWAT does not apply T2I filtering as post-processing, because determining a proper threshold for each class requires a large validation set which is not allowed in our realistic FSR setup.

\section{Analysis of Data Augmentation Methods}
\label{sec:msda_example}

\begin{figure*}[t]
    \centering
    \includegraphics[width=1.0\linewidth, clip=true,trim = 5mm 0mm 5mm 0mm]{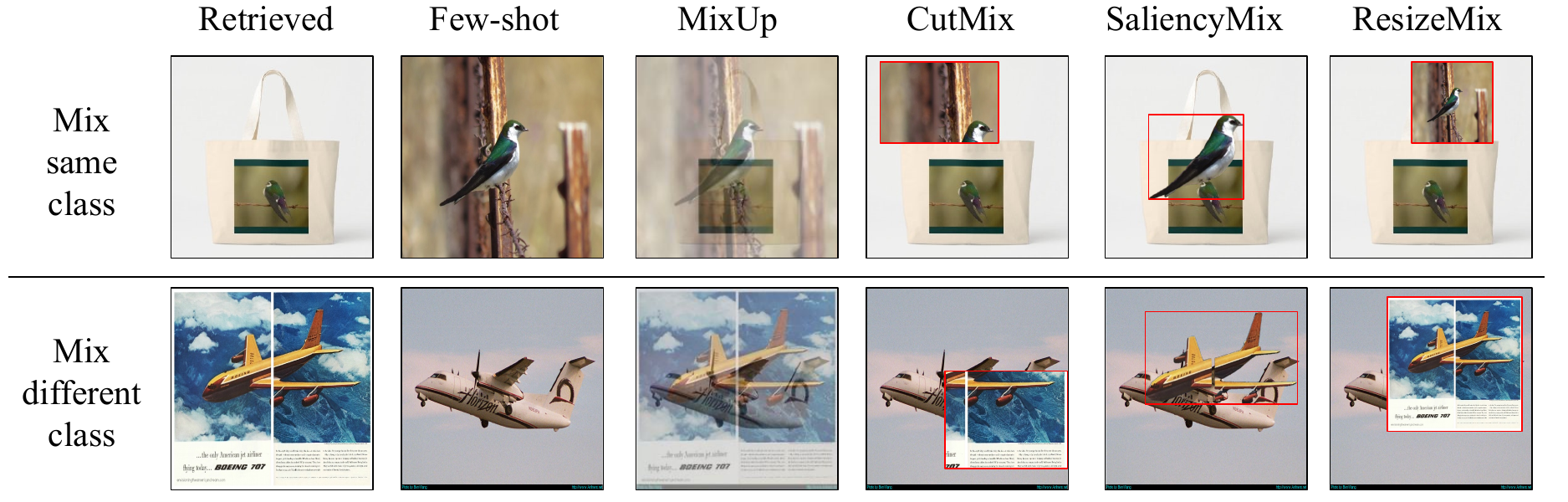}
    \vspace{-4mm}
    \caption{\small
    {\bf Examples of different mixed sample data augmentation (MSDA) methods.} We show two examples where the first row shows mixing the retrieved and few-shot images from the same class in Semi-Aves dataset~\cite{semi-aves}, and the second row shows mixing images from different classes in the FGVC-Aircraft dataset~\cite{aircraft}. 
    These MSDA methods encourage the model to learn from small discriminative parts of the object or details in the background (e.g. part of a bird or airplane), thereby improving the performance.
    Compared to CutMix~\cite{yun2019cutmix} and its variants (SaliencyMix~\cite{uddin2020saliencymix}, ResizeMix~\cite{qin2020resizemix}), MixUp~\cite{zhang2017mixup} augments data by simply interpolating two images, which
    may create unnatural artifacts that could confuse the model~\cite{yun2019cutmix}. 
    }
    \label{fig:example_msda}
    % \vspace{-5mm}
\end{figure*}

We show examples of various mixed sample data augmentation (MSDA) methods in Fig.~\ref{fig:example_msda} and compare their performance using SWAT across five datasets (Semi-Aves, Flowers, Aircraft, EuroSAT and DTD) in Table~\ref{tab:ablate_msda}. Results show that CutMix performs the best with minimal computation overhead, while SaliencyMix~\cite{uddin2020saliencymix} performs similarly but incurs significant overhead due to the extraction of saliency maps.

{
\setlength{\tabcolsep}{0.4em} 
\begin{table}[t]
\centering
\caption{\small {\bf Comparison of using different Mixed Sample Data Augmentation (MSDA) methods in SWAT.} 
Compared with no mixing,
all mixing methods
increase accuracy by 1-2\%.
MixUp~\cite{zhang2017mixup} slightly underperforms other CutMix variants, likely because it creates unnatural artifacts that could confuse the model~\cite{yun2019cutmix}.
We also find that randomly applying CutMix regardless of few-shot and retrieved images performs better than strictly cutting few-shot patches and pasting them into retrieved images (CutMix-strict), likely because doing so limits the diversity of data augmentation.
By default, SWAT uses CutMix~\cite{yun2019cutmix}, which achieves the best performance and low computation overhead among all the compared MSDA methods.
{\bf Bold} and \underline{underlined} numbers mark the best and second best numeric metrics; \textcolor{Green}{superscripts} denote improvements over no mixing.
See visual examples of different MSDA methods in Fig.~\ref{fig:example_msda}.
}
\vspace{-2mm}
\label{tab:ablate_msda}
\scalebox{0.9}{
\begin{tabular}{lllll}
\toprule
\multirow{2}{*}{\makecell{MSDA\\method}} &\multirow{2}{*}{\makecell{compute\\overhead}} & \multicolumn{3}{c}{mean accuracy of five datasets} \\ 
\cmidrule{3-5}
 & & 4-shot & 8-shot & 16-shot \\
\midrule
No mixing &None &68.3 &71.9 &75.6  \\
MixUp~\cite{zhang2017mixup} &Low &69.1$^{\textcolor{Green}{+0.8}}$ &73.0$^{\textcolor{Green}{+1.1}}$ &76.6$^{\textcolor{Green}{+1.0}}$ \\
SaliencyMix~\cite{uddin2020saliencymix} &High &\underline{70.1}$^{\textcolor{Green}{+1.8}}$ &\textbf{74.4}$^{\textcolor{Green}{+2.5}}$ &\underline{77.7}$^{\textcolor{Green}{+2.1}}$ \\
CMO~\cite{park2022majority} & Med &69.9$^{\textcolor{Green}{+1.6}}$ &74.1$^{\textcolor{Green}{+2.2}}$ &77.1$^{\textcolor{Green}{+1.5}}$ \\
ResizeMix~\cite{qin2020resizemix} &Med &69.6$^{\textcolor{Green}{+1.3}}$ &74.1$^{\textcolor{Green}{+2.2}}$ &77.2$^{\textcolor{Green}{+1.6}}$ \\
CutMix-strict &Med &\underline{70.1}$^{\textcolor{Green}{+1.8}}$ &73.8$^{\textcolor{Green}{+1.9}}$ &77.6$^{\textcolor{Green}{+2.0}}$ \\
\rowcolor{gray!15}CutMix~\cite{yun2019cutmix} &Low &\textbf{70.5}$^{\textcolor{Green}{+2.2}}$ &\underline{74.2}$^{\textcolor{Green}{+2.3}}$ & \textbf{77.8}$^{\textcolor{Green}{+2.2}}$ \\
\bottomrule
\end{tabular}}
\end{table}
}

\begin{figure}[t]
  \centering
  \includegraphics[width=0.99\linewidth, clip=true,trim = 0mm 0mm 0mm 0mm]
  {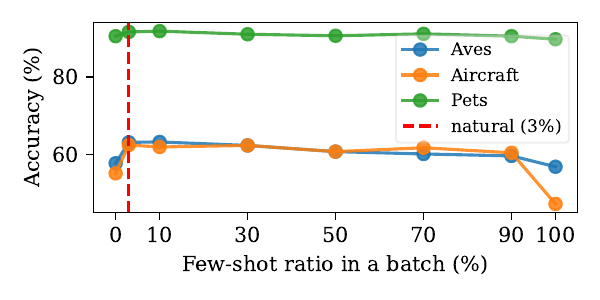}
  \vspace{-3mm}
  \caption{\small
  {\bf Comparison of final accuracy with varying few-shot ratio in a batch.}
  Our SWAT adopts a ``natural ratio'' by combining retrieved data and few-shot annotated data without sophisticated resampling methods. The natural ratio is 3\%, meaning 3\% data in each batch is from the few-shot data. Results show that the ``natural ratio'' (red dashed line) performs better than either increasing the ratio (which reduces data diversity) or decreasing it (which increases domain gap).
  }
  \label{fig:ablate_fs-ratio}
% \vspace{-4mm}
\end{figure}

{\bf Impact of Mixing Ratio.}
We further explore the impact of the mixing ratio between retrieved and few-shot data within a batch when applying CutMix augmentation.
Results in Fig.~\ref{fig:ablate_fs-ratio} shows that
SWAT achieves the best performance when applying a ``natural ratio'' by combining retrieved data and few-shot annotated data without sophisticated resampling methods. 
This encourages future work to explore better mixed sample data augmentation methods.

\section{Validating the Design of SWAT}
\label{sec:swat+}

{
\setlength{\tabcolsep}{0.3em}
\begin{table}[t]
\vspace{-3.5mm}
\centering
\caption{
{\bf Comparison of ImageNet accuracy and training time cost of different stage-2 training strategies.}
We experiment by finetuning the stage-1 trained model on 16-shot data from ImageNet following different training strategies.
Results show that finetuning only the classifier (as done in SWAT) improves the rare class accuracy significantly more than finetuning the visual encoder only. In addition, the training time cost of retraining the classifier is much less than finetuning the visual encoder. 
Moreover, finetuning both the visual encoder and classifier achieves further improvement over SWAT, likely due to the insufficient representation learning in stage 1 with only 50 training epochs. 
We denote this scenario as \textbf{SWAT+} and report its performance across all datasets in Fig.~\ref{fig:compare_sota}, Table~\ref{tab:compare_sota_detail} and Table~\ref{tab:compare_sota_detail_vitb16}. 
}
\vspace{-2mm}
% \vspace{-3mm}
\label{tab:compare_stage2_finetuning}
\scalebox{0.85}{
\begin{tabular}{cclllc}
\toprule
FT encoder   & FT classifier   & Avg    & common      & rare        & time \\
\midrule
\multicolumn{2}{l}{acc after stage-1} & 67.1        & 68.3        & 56.1        &               \\
\midrule
& \checkmark & 67.6$^{\textcolor{Green}{+0.5}}$ & 68.3$^{\textcolor{gray}{+0.0}}$ & \underline{61.2}$^{\textcolor{Green}{+5.1}}$ & \textbf{\textcolor{Green}{0.5}} mins \\
\checkmark & & 67.4$^{\textcolor{Green}{+0.3}}$ & 68.5$^{\textcolor{Green}{+0.2}}$ & 57.3$^{\textcolor{Green}{+1.2}}$ & 15 mins  \\
\checkmark & \checkmark &69.3$^{\textcolor{Green}{+2.2}}$ &70.1$^{\textcolor{Green}{+1.8}}$ &\textbf{62.0$^{\textcolor{Green}{+5.9}}$} & 15 mins\\
\bottomrule
\end{tabular}}
% \caption{}
\label{tab:my-table}
% \vspace{-4mm}
\end{table}
}

{
\setlength{\tabcolsep}{0.52em} 
\begin{table*}[t]
\small
\centering
\caption{{\bf Comparison of different finetuning methods.} 
We compare SWAT with state-of-the-art probing-based and finetuning-based methods using the 
same training data (a mix of retrieved and few-shot data).
We experiment with the T2I-filtered retrieved data for each dataset (cf. Table~\ref{tab:retrieve_methods}).
We use the same set of hyperparameters in Section~\ref{sec:hyperparams} for all methods 
except using a larger batch size of 256 for FLYP following~\cite{goyal2023finetune}.
Results show that finetuning-based methods largely outperform probing-based methods, indicating the necessity of finetuning the visual encoder to learn better representation. 
In addition, ensembling the finetuned model with the zero-shot model (WiSE-FT with $\alpha=0.5$~\cite{wortsman2022robust}) leads to much worse accuracy than standard finetuning, 
likely because the zero-shot OpenCLIP model 
struggles to recognize these fine-grained concepts~\cite{saha2024improved}. 
Finally, SWAT outperforms other finetuning methods, validating its effectiveness in mitigating domain gaps and imbalanced distribution issues in retrieved data.
We highlight the best number in \textbf{bold} and \underline{underline} the second best.
}
% \vspace{-2mm}
\label{tab:ablate_finetuning method}
\scalebox{0.85}{
\begin{tabular}{lccccccccccccccc!{\vrule}ccc}
\toprule
method & \multicolumn{3}{c}{Semi-Aves} & \multicolumn{3}{c}{Flowers} & \multicolumn{3}{c}{Aircraft} & \multicolumn{3}{c}{EuroSAT} & \multicolumn{3}{c}{DTD} & \multicolumn{3}{c}{mean accuracy} \\
\cmidrule(lr){2-4} \cmidrule(lr){5-7} \cmidrule(lr){8-10} \cmidrule(lr){11-13} \cmidrule(lr){14-16} \cmidrule(lr){17-19}
(shots) & 4 & 8 & 16 & 4 & 8 & 16 & 4 & 8 & 16 & 4 & 8 & 16 & 4 & 8 & 16 & 4 & 8 & 16 \\ 
 \midrule
linear probing~\cite{radford2021learning} \textsubscript{ICML'24} & 49.8 & 52.4 & 54.4 & 86.9 & 89.4 & 92.8 & 34.6 & 35.8 & 38.2 & 68.0 & 78.2 & 82.4 & 61.7 & 65.5 & 68.9 & 60.2 & 64.3 & 67.3 \\
CMLP~\cite{lin2023multimodality} \textsubscript{CVPR'23} & 49.2 & 51.9 & 53.6 & 87.0 & 89.3 & 92.9 & 34.1 & 35.4 & 37.8 & 70.1 & 79.4 & 83.5 & 61.3 & 64.8 & 68.6 & 60.3 & 64.2 & 67.3 \\
REAL-Linear~\cite{parashar2024neglected} \textsubscript{CVPR'24} & 51.0 & 52.5 & 54.3 & 85.0 & 86.4 & 88.7 & 31.2 & 31.8 & 33.8 & 66.5 & 73.4 & 76.2 & 62.2 & 64.7 & 67.4 & 59.2 & 61.8 & 64.1 \\
\midrule

standard FT~\cite{radford2021learning} \textsubscript{ICML'24} & 55.2 & 57.6 & \underline{60.4} & \underline{89.4} & \underline{92.8} & \underline{95.5} & \underline{48.9} & \underline{51.2} & \underline{53.0} & \underline{83.3} & \underline{88.3} & \underline{92.8} & 61.5 & 65.6 & \underline{70.3} & \underline{67.7} & \underline{71.1} & \underline{74.4} \\

WiSE-FT~\cite{wortsman2022robust} \textsubscript{CVPR'22} & 51.7 & 53.2 & 56.1 & 82.1 & 84.6 & 87.0 & 32.2 & 33.2 & 34.0 & 77.4 & 85.2 & 87.4 & \textbf{64.1} & 66.7 & 69.4 & 61.5 & 64.6 & 66.8 \\
FLYP~\cite{goyal2023finetune} \textsubscript{CVPR'23} & \underline{56.0} & \underline{57.7} & 59.6 & 88.1 & 91.1 & 94.4 & 47.9 & 49.7 & 51.2 & 75.4 & 83.3 & 90.6 & \underline{63.1} & \underline{67.4} & \underline{70.3} & 66.1 & 69.2 & 72.6 \\

\rowcolor{gray!15}SWAT (ours) & \textbf{58.6} & \textbf{61.3} & \textbf{63.8} & \textbf{91.0} & \textbf{94.7} & \textbf{97.5} & \textbf{55.5} & \textbf{58.1} & \textbf{62.6} & \textbf{84.6} & \textbf{89.2} & \textbf{93.7} & 63.0 & \textbf{67.6} & \textbf{71.6} & \textbf{70.5} & \textbf{74.2} & \textbf{77.8} \\
\bottomrule
\end{tabular}}
\end{table*}
}

\begin{figure}[t]
  \centering
  \includegraphics[width=0.99\linewidth, clip=true,trim = 0mm 0mm 0mm 0mm]
  {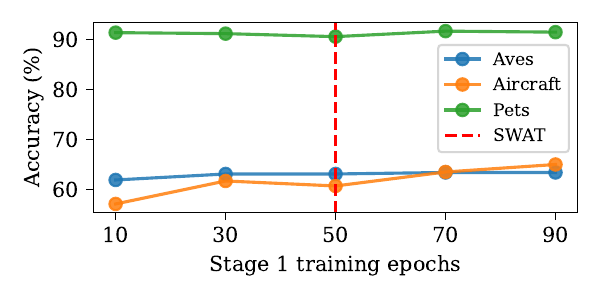}
  \vspace{-3mm}
  \caption{\small
  {\bf Comparison of final accuracy with increasing stage-1 training epochs.}
  Results show that increasing stage-1 training epochs generally increases final accuracy slightly, without overfitting issues.
  This is likely due to the improved representation learning. We set stage-1 training epochs to 50 for all datasets, following the realistic FSR setup that does not tune hyperparameters using a large validation set.
  }
  \label{fig:ablate_stage1_epochs}
% \vspace{-4mm}
\end{figure}

{\bf Ablation of Stage-2 Training Strategy.}
To validate the design of stage-2 classifier retraining in SWAT, we compare the performance of different stage-2 training strategies in Table~\ref{tab:compare_stage2_finetuning}.
Results show that retraining only the classifier achieves significantly larger accuracy improvement on rare classes than retraining only the visual encoder, validating its effectiveness in mitigating imbalanced distribution.
Furthermore, we find that retraining both the visual encoder and classifier improves further over SWAT by 1$\sim$2\%. 
We hypothesize that this is due to the insufficient representation learning in stage 1 with only 50 training epochs (recall that we follow realistic evaluation protocol that do not use validation set to tune hyperparameters).
A supporting evidence is found in Fig.~\ref{fig:ablate_stage1_epochs} where we show that longer training in stage 1 generally yields better final accuracy. We denote this strategy as \textbf{SWAT+} and report its performance across all datasets in Fig.~\ref{fig:compare_sota}, Table~\ref{tab:compare_sota_detail} and Table~\ref{tab:compare_sota_detail_vitb16}.
Considering the comparable performance and much less training time cost, we adopt classifier retraining for stage 2 in our SWAT.

{\bf Comparison with SOTA Finetuning Methods.}
Table~\ref{tab:ablate_finetuning method} shows that our SWAT outperforms recent probing-based or finetuning-based methods using the same retrieved and few-shot data. 
SWAT also outperforms recent ensembling-based~\cite{wortsman2022robust} and contrastive finetuning~\cite{goyal2023finetune} methods, highlighting the effectiveness of our proposed stage-wise training in mitigating domain gap and imbalanced distribution issues.

\section{Further Analyses on SWAT}
\label{sec:ablation-studies}

{\bf Impact of Stage-1 Training Epochs.}
We compare the final accuracy with a varying number of epochs for stage-1 end-to-end finetuning in Fig.~\ref{fig:ablate_stage1_epochs}. 
Results show that longer training generally yields better performance due to improved representation learning. Please note that our realistic FSR setup does not allow using a validation set to tune training epochs. Our paper sets the number of training epochs to 50 for all datasets (cf. Section~\ref{sec:hyperparams}).

{
\setlength{\tabcolsep}{0.9em} 
\begin{table}[t]
\caption{\small 
{\bf Comparison of classifier initialization methods in SWAT.}
We compare the final test accuracy by initializing the classifier before stage-1 end-to-end finetuning in different ways.
Initializing classifier weights with text embedding features leads to better performance than random initialization. 
\cite{kumar2022fine} explains that using randomly initialized classifier weights to finetune the model can distort the features of pretrained model, leading to worse finetuning performance.
Throughout this work, we use prompts in~\cite{parashar2024neglected} to initialize classifier weights in SWAT.
\textcolor{Green}{Subscripts} mark the performance improvement compared with random initialization.
}
% \vspace{-2mm}
\label{tab:ablate_cls_init}
\scalebox{0.9}{
\begin{tabular}{llll}
\toprule
\multirow{2}{*}{\makecell{classifier\\initialization}} & \multicolumn{3}{c}{mean accuracy of nine datasets} \\ 
\cmidrule(r){2-4}
& 4-shot & 8-shot & 16-shot \\
\midrule
random & 72.7 & 75.1 & 77.5 \\
\rowcolor{gray!15}text embedding~\cite{parashar2024neglected} 
& 73.6$^{\textcolor{Green}{+0.9}}$ 
& 76.1$^{\textcolor{Green}{+1.0}}$ 
& 78.2$^{\textcolor{Green}{+0.7}}$
\\ 
\bottomrule
\end{tabular}}
\end{table}
}

{\bf Classifier Initialization.}
We compare different classifier initialization methods for SWAT (Table~\ref{tab:ablate_cls_init}). Results show that initializing with text embedding yields better performance than random initialization.

\begin{figure*}[t]
    \centering
    \small
    % \vspace{-3mm}
    \begin{tabular}{p{8.3cm}p{8.3cm}}
    \includegraphics[width=1.0\linewidth, clip=true,trim = 0mm 0mm 0mm 0mm]{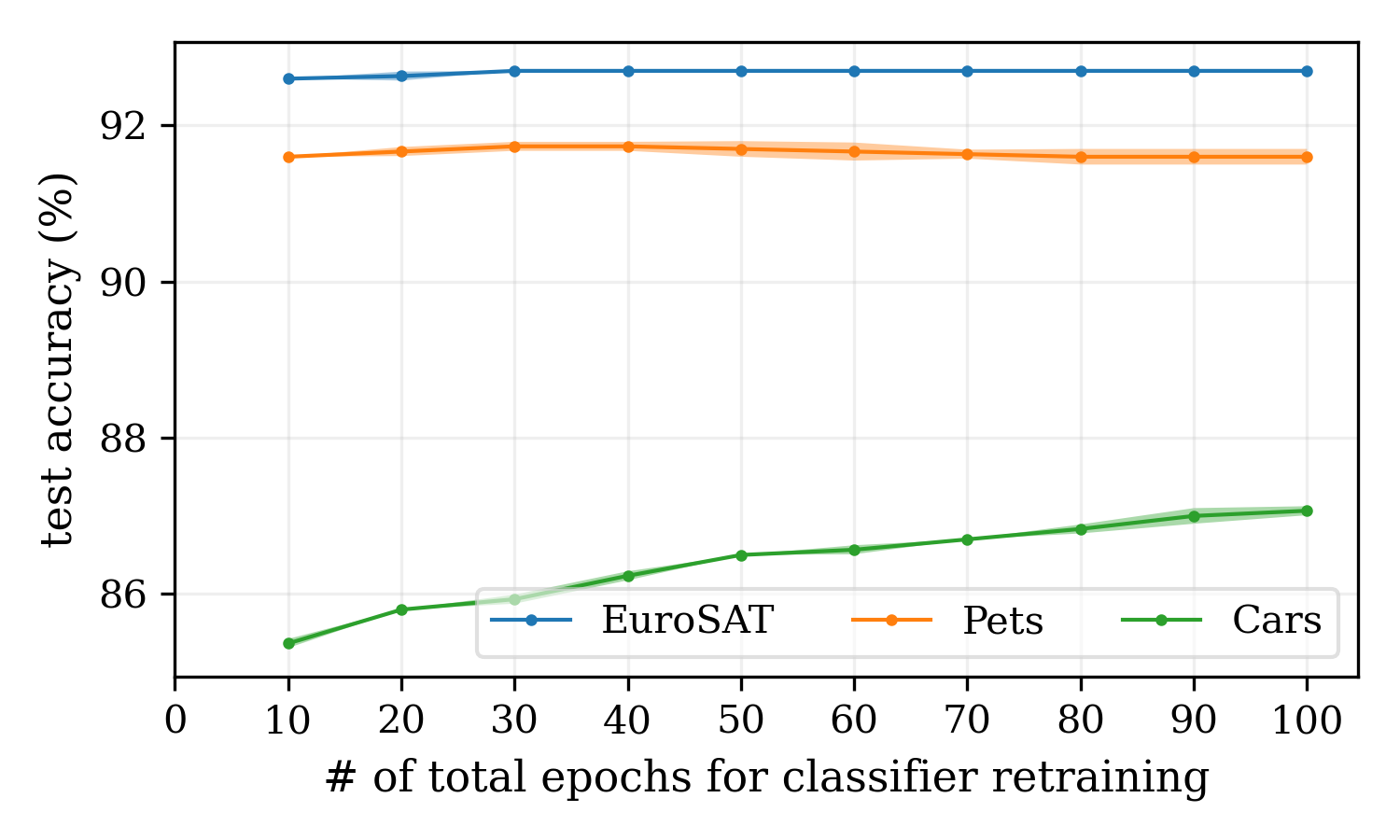}
    &
    \includegraphics[width=1.0\linewidth, clip=true,trim = 0mm 0mm 0mm 0mm]{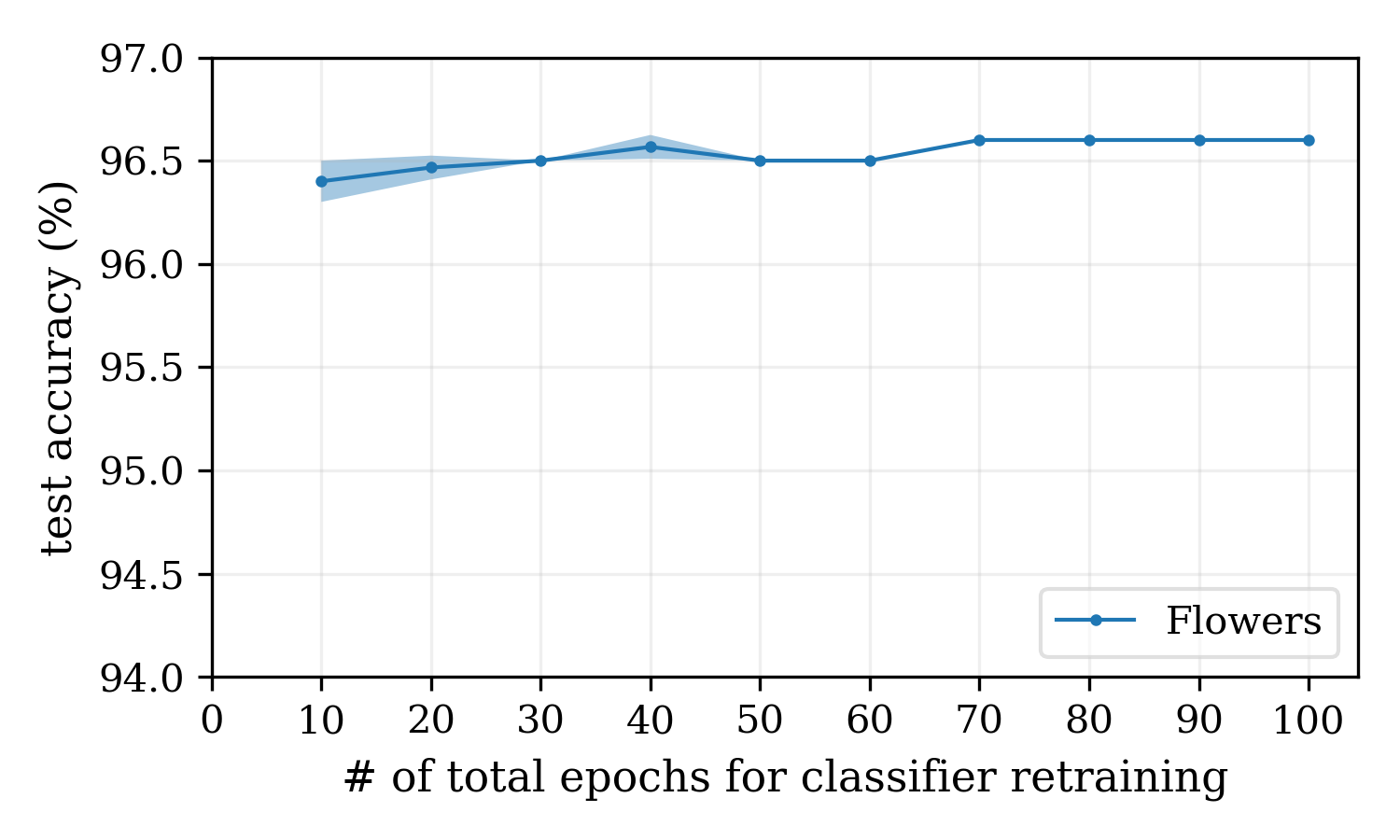}
    \\
    \includegraphics[width=1.0\linewidth, clip=true,trim = 0mm 0mm 0mm 0mm]{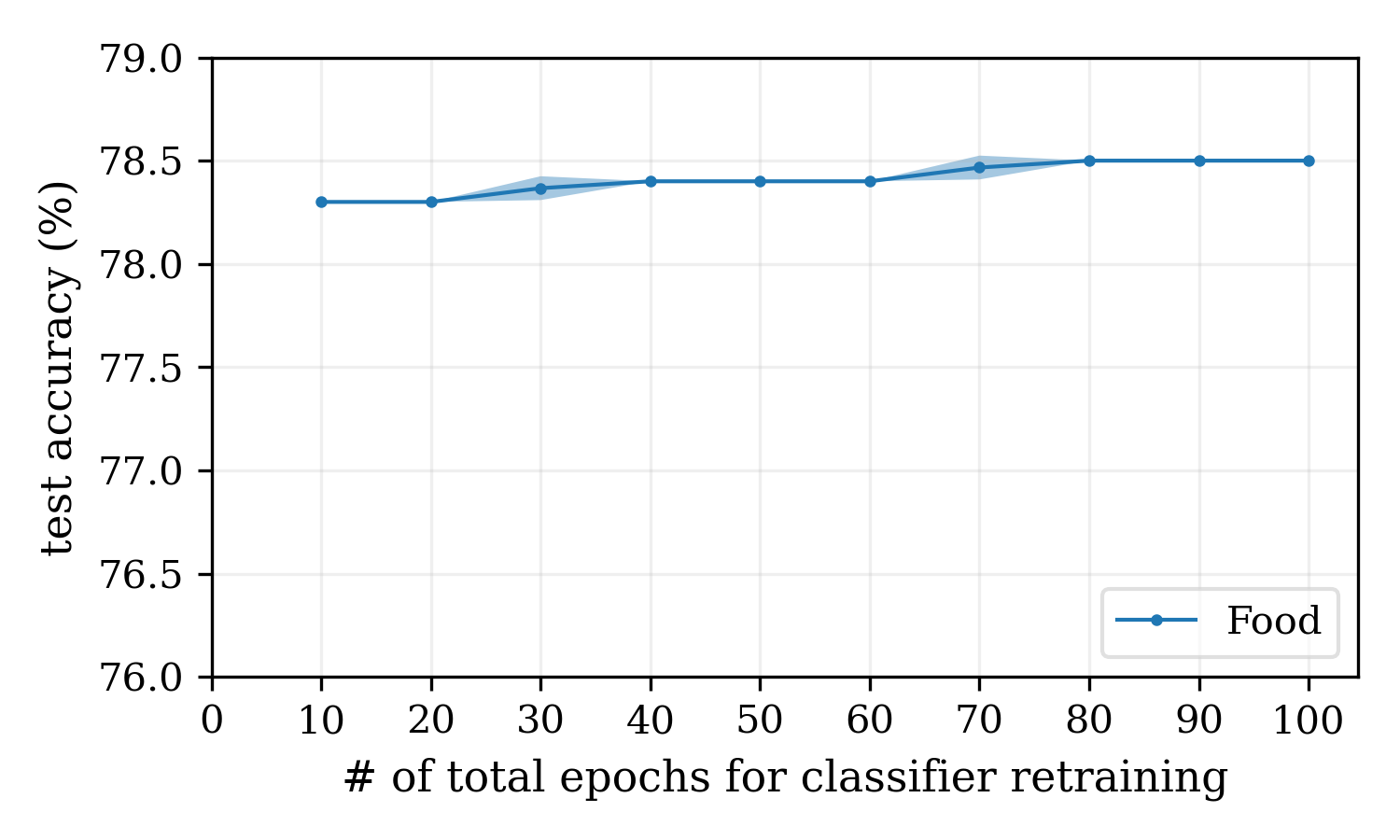} 
    &
    \includegraphics[width=1.0\linewidth, clip=true,trim = 0mm 0mm 0mm 0mm]{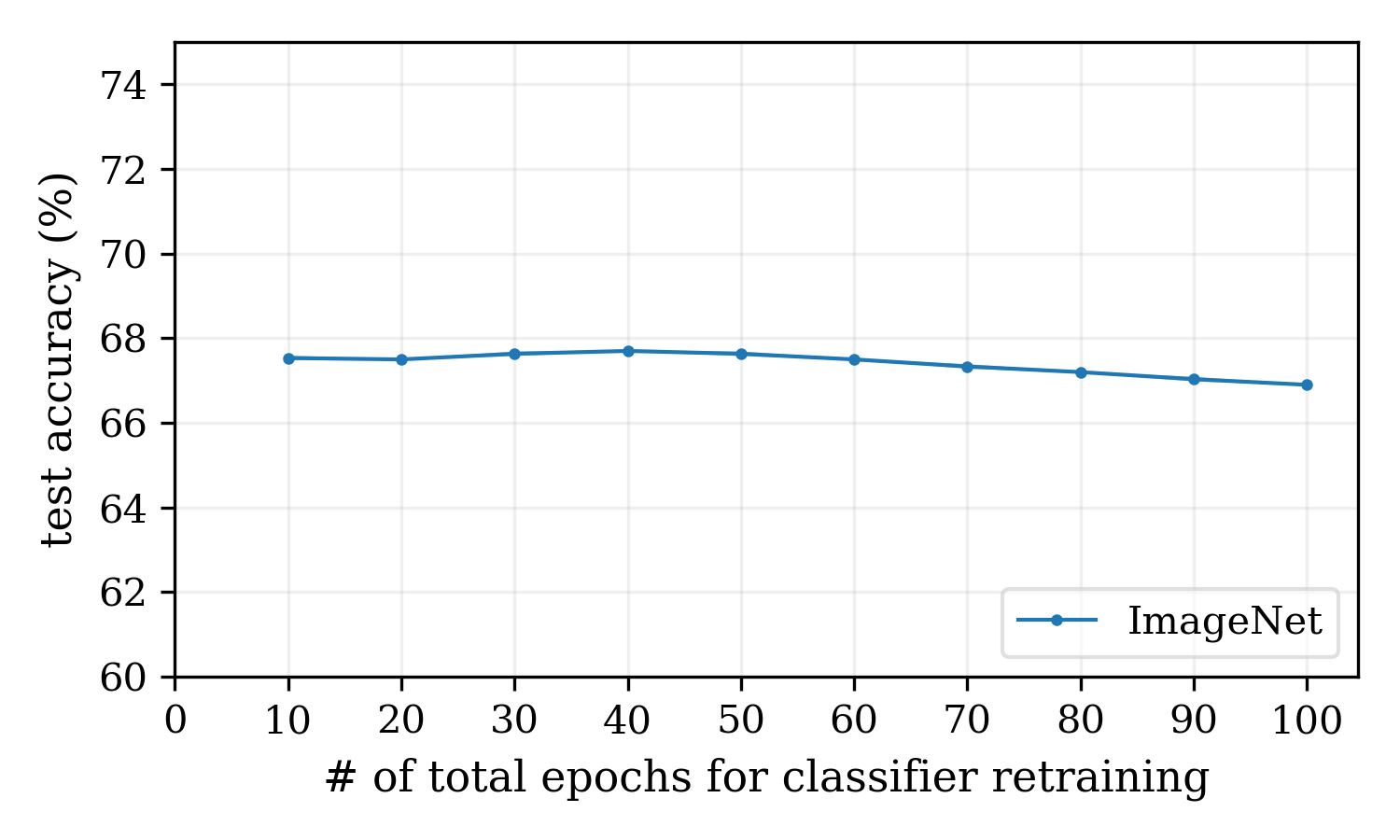}
    \\
    \end{tabular}
    \caption{\small {\bf Retraining the classifier on the few-shot data does not suffer from overfitting.} 
    We show the final test accuracies by retraining the classifier on the few-shot data for different epoch numbers.
    For each dataset, we perform three runs of training with different random seeds.
    Results show that testing accuracy remains stable with more epochs and shows small standard deviations, indicating classifier retraining does not suffer from overfitting. 
    }
% \vspace{-3mm}
\label{fig:no_overfit_more}
\end{figure*}

{\bf Retraining Classifier does not Overfit.}
In Fig.~\ref{fig:no_overfit_more}, we show the final test accuracy after retraining the classifier across varying epoch numbers. 
For all datasets, accuracy remains stable with increasing epochs. The small standard deviations across three runs with different random seeds confirm that stage-2 classifier retraining with few-shot data does not suffer from overfitting.

{\bf More Detailed Experimental Results.}
In addition. we show the detailed performance of classifier retraining for each dataset in Table~\ref{tab:stage2_improvement_dataset}. The rare classes of the Aircraft dataset show significant performance gains (>10\%) after classifier retraining, demonstrating the efficacy of classifier retraining with few-shot data in mitigating domain gaps and imbalanced distribution.
In addition, we include the detailed ablation of SWAT components on each dataset in Table~\ref{tab:components_contribution_dataset}. 
Results show that applying CutMix~\cite{yun2019cutmix} and classifier retraining effectively mitigate the domain gap and imbalanced distribution problem, verifying the design of SWAT.

{
\setlength{\tabcolsep}{0.4em} 
\begin{table*}[t]
\small
\centering
\caption{\small 
{\bf Detailed comparison of the accuracy of common and rare classes after stage-1 and stage-2 training.}
We define the rare classes as the 10\% least frequent classes in retrieved data and the rest as the common classes.
Results show that stage-2 classifier retraining clearly improves recognition accuracy on both common and rare classes in all methods,
including finetuning on few-shot data only, on retrieved data only, and on mixed data with or without CutMix data augmentation.
Importantly, the improvement on rare classes is more significant than that on common classes, confirming that classifier retraining mitigates the issue of imbalanced distribution in the retrieved data.
We report the accuracy for each dataset using 16-shot examples. 
}
% \vspace{-3mm}
\label{tab:stage2_improvement_dataset}
\scalebox{0.9}{
\begin{tabular}{cccccccccccc!{\vrule}c}
\toprule
\makecell{data used in\\stage-1: finetuning} & stage & classes & Semi-Aves & Flowers & Aircraft & EuroSAT & DTD & Pets & Food & Cars & ImageNet & average \\

\midrule

\multirow{6}{*}{\makecell{few-shot only\\(balanced)}} & \multirow{3}{*}{\makecell{stage-1\\finetuning}} & common & 56.1 & 97.6 & 43.2 & 94.9 & 73.5 & 90.6 & 78.8 & 88.6 & 68.0 & 76.8 \\
 &  & rare & 63.5 & 100.0 & 34.6 & 87.1 & 76.7 & 84.9 & 74.8 & 84.4 & 56.8 & 73.6 \\
 &  & average & 56.9 & 97.4 & 42.4 & 94.1 & 73.9 & 90.0 & 78.4 & 88.0 & 66.9 & 76.4 \\
 
 \cmidrule{2-13}
 
 & \multirow{3}{*}{\makecell{stage-2\\classifier retraining}} & common & 56.0 & 97.5 & 47.3 & 95.0 & 72.9 & 90.0 & 78.9 & 88.5 & 67.1 
 & 77.0 \\
 
 &  & rare & 63.8 & 100.0 & 46.4 & 87.2 & 76.7 & 86.9 & 74.5 & 83.3 & 57.3 
 & 75.1 \\

 &  & average & 56.8 & 97.4 & 47.2 & 94.3 & 73.3 & 89.7 & 78.4 & 87.9 & 66.1 
 & 76.8 \\

\midrule
 
\multirow{6}{*}{\makecell{retrieved only\\(imbalanced)}} & \multirow{3}{*}{\makecell{stage-1\\finetuning}} & common & 56.2 & 84.4 & 52.5 & 30.4 & 52.4 & 90.8 & 76.0 & 78.1 & 62.5 & 64.8 \\
 &  & rare & 15.0 & 54.4 & 10.2 & 0.0 & 61.1 & 85.5 & 73.3 & 51.8 & 46.4 & 44.2 \\
 &  & average & 52.1 & 81.6 & 48.3 & 27.9 & 53.3 & 90.3 & 75.7 & 75.3 & 60.9 & 62.8 \\
  
 \cmidrule{2-13}
 
 & \multirow{3}{*}{\makecell{stage-2\\classifier retraining}} & common & 60.0 & 90.2 & 57.5 & 32.2 & 54.6 & 90.9 & 76.8 & 82.9 & 64.7 
 & 67.8 \\

 &  & rare & 36.9 & 77.8 & 33.7 & 0.0 & 62.8 & 86.6 & 74.2 & 66.8 & 58.8 
 & 55.3 \\
 
 &  & average & 57.7 & 88.6 & 55.1 & 29.4 & 55.4 & 90.5 & 76.6 & 81.2 & 64.1 
 & 66.5 \\
 
\midrule
 
\multirow{6}{*}{\makecell{retrieved + few-shot}} & \multirow{3}{*}{\makecell{stage-1\\ finetuning}} & common & 61.4 & 94.6 & 57.4 & 93.4 & 62.3 & 91.4 & 77.9 & 81.8 & 64.8 & 76.1 \\
 &  & rare & 49.4 & 96.8 & 26.2 & 87.5 & 69.4 & 87.4 & 75.9 & 68.8 & 52.7 & 68.2 \\
 &  & average & 60.2 & 94.7 & 54.3 & 92.8 & 63.1 & 91.0 & 77.7 & 80.3 & 63.6 & 75.3 \\
  
 \cmidrule{2-13}
 
 & \multirow{3}{*}{\makecell{stage-2\\classifier retraining}} & common & 61.6 & 95.4 & 60.6 & 93.4 & 62.8 & 91.3 & 78.0 & 84.0 & 65.7 
 & 77.0 \\
 
 &  & rare & 52.2 & 98.0 & 44.3 & 87.5 & 68.9 & 87.1 & 76.0 & 73.1 & 57.6 
 & 71.6 \\
 
 &  & average & 60.6 & 95.4 & 59.0 & 92.8 & 63.5 & 91.0 & 77.8 & 82.8 & 64.9 
 &76.4 \\

\midrule
 
\multirow{6}{*}{\makecell{retrieved + few-shot\\w/ CutMix}} & \multirow{3}{*}{\makecell{stage-1\\finetuning}} & common & 63.7 & 96.4 & 61.3 & 93.4 & 64.8 & 91.5 & 78.3 & 83.9 & 68.3 & 78.0 \\
 &  & rare & 55.8 & 100.0 & 34.7 & 83.9 & 72.2 & 89.2 & 77.4 & 78.0 & 56.1 & 71.9 \\
 &  & average & 62.9 & 96.3 & 58.7 & 92.5 & 65.6 & 91.3 & 78.2 & 83.2 & 67.1 & 77.3 \\
  
 \cmidrule{2-13}
 
 & \multirow{3}{*}{\makecell{stage-2\\classifier retraining}} & common & 64.0 & 96.4 & 63.7 & 93.7 & 65.6 & 91.9 & 78.4 & 86.1 & 68.3 
 & 78.7 \\

 &  & rare & 54.9 & 100.0 & 50.9 & 82.0 & 72.2 & 88.6 & 77.5 & 79.9 & 61.2 
 & 74.1 \\

 &  & average & 63.1 & 96.4 & 62.4 & 92.6 & 66.3 & 91.6 & 78.3 & 85.4 & 67.6 
 & 78.2 \\ 

\bottomrule
\end{tabular}
}
\end{table*}
}

{
\setlength{\tabcolsep}{0.32em} 
\begin{table*}[t]
\small
\centering
\caption{\small 
{\bf Ablation study on important components in our SWAT.}
We show the detailed performance improvements by each component for each dataset in our SWAT.
Finetuning on simply combined retrieved and few-shot data underperforms finetuning solely on few-shot data (8-shot and 16-shot, with or without CutMix), due to the large domain gap and imbalanced distribution in retrieved data. 
However, further applying CutMix and classifier retraining improves the test accuracy significantly, confirming their effectiveness in mitigating the domain gap and imbalanced distributions.
We also compare the performance of few-shot finetuning with and without CutMix data augmentation. The results indicate more few-shot data yields more improvements, likely due to stronger data augmentation.
}
\vspace{-3mm}
\label{tab:components_contribution_dataset}
\scalebox{0.9}{
\begin{tabular}{llcccc!{\vrule}ccccccccc!{\vrule}c}
\toprule
shots & method & \makecell{finetune\\model} & \makecell{retrieve\\data} &\makecell{apply\\CutMix} & \makecell{retrain\\classifier} & Semi-Aves & Flowers & Aircraft & EuroSAT & DTD & Pets & Food & Cars & ImageNet & average \\
\midrule

\multirow{6}{*}{4} & CLAP~\cite{clap24} &  &  &  &  & 34.0 & 90.1 & 28.0 & 74.7 & 63.0 & 87.0 & 76.7 & 84.9 & 64.0 & 66.9 \\
 & FTFS (ours) & \checkmark &  &  &  & 47.5 & 92.5 & 27.9 & 81.6 & 66.6 & 88.7 & 75.8 & 81.5 & 62.3 & 69.4 \\
 & FTFS (ours) & \checkmark &  & \checkmark &  & 48.0 & 92.2 & 28.8 & 81.8 & 66.7 & 89.0 & 76.1 & 82.5 & 62.4 & 69.7 \\
 &  & \checkmark & \checkmark &  &  & 54.7 & 89.7 & 50.1 & 80.2 & 56.3 & 90.7 & 76.4 & 76.9 & 61.8 & 70.8 \\
 &  & \checkmark & \checkmark & \checkmark &  & 57.9 & 90.2 & 53.8 & 83.2 & 58.7 & 91.0 & 77.2 & 79.8 & 65.2 & 73.0 \\
 & SWAT (ours) & \checkmark & \checkmark & \checkmark & \checkmark & 58.5 & 90.6 & 55.7 & 83.2 & 58.3 & 91.3 & 77.3 & 81.1 & 65.8 & 73.5 \\
\midrule

\multirow{6}{*}{8} & CLAP~\cite{clap24} &  &  &  &  & 42.9 & 92.9 & 33.6 & 77.4 & 66.4 & 87.8 & 77.5 & 86.1 & 65.6 & 70.0 \\
 & FTFS (ours) & \checkmark &  &  &  & 51.2 & 95.4 & 33.1 & 90.3 & 71.0 & 89.3 & 76.0 & 83.5 & 64.4 & 72.7 \\
 & FTFS (ours) & \checkmark &  & \checkmark &  & 52.3 & 95.2 & 35.4 & 89.4 & 70.6 & 89.6 & 77.0 & 85.3 & 64.8 & 73.3 \\
 &  & \checkmark & \checkmark &  &  & 57.3 & 91.9 & 52.4 & 87.0 & 59.2 & 91.1 & 76.8 & 78.9 & 62.5 & 73.0 \\
 &  & \checkmark & \checkmark & \checkmark &  & 60.6 & 93.7 & 55.7 & 89.1 & 61.8 & 90.8 & 77.6 & 81.3 & 65.8 & 75.2 \\
 & SWAT (ours) & \checkmark & \checkmark & \checkmark & \checkmark & 60.8 & 94.1 & 59.1 & 89.2 & 62.6 & 90.8 & 77.5 & 83.5 & 66.6 & 76.0 \\
\midrule

\multirow{6}{*}{16} & CLAP~\cite{clap24} &  &  &  &  & 49.2 & 94.8 & 39.1 & 81.7 & 69.9 & 88.4 & 78.5 & 87.8 & 67.1 & 72.9 \\
 & FTFS (ours) & \checkmark &  &  &  & 55.3 & 97.0 & 37.0 & 94.0 & 73.3 & 89.5 & 77.1 & 85.7 & 66.7 & 75.1 \\
 & FTFS (ours) & \checkmark &  & \checkmark &  & 56.5 & 97.1 & 42.7 & 94.3 & 73.4 & 89.6 & 78.2 & 87.8 & 66.9 & 76.3 \\
 &  & \checkmark & \checkmark &  &  & 60.2 & 94.7 & 54.3 & 92.8 & 63.1 & 91.0 & 77.7 & 80.3 & 63.6 & 75.3 \\
 &  & \checkmark & \checkmark & \checkmark &  & 62.9 & 96.3 & 58.7 & 92.5 & 65.6 & 91.3 & 78.2 & 83.2 & 67.1 & 77.3 \\
 & SWAT (ours) & \checkmark & \checkmark & \checkmark & \checkmark & 63.1 & 96.4 & 62.4 & 92.6 & 66.3 & 91.6 & 78.3 & 85.4 & 67.6 & 78.2 \\ \bottomrule
\end{tabular}
}
\end{table*}
}

{
\setlength{\tabcolsep}{0.9em} 
\begin{table*}[t]
\small
\centering
\caption{\small {\bf Impact of retrieval size (number of images per class) on the performance of SWAT.} 
We show the performance of SWAT on each dataset using different numbers of retrieved images.
We highlight the best number in \textbf{bold} and \underline{underline} the second best. 
Importantly, we find that retrieving 10 images per class works best for Flowers, EuroSAT, DTD, and Cars datasets. This is probably because LAION-400M contains limited images that match these downstream concepts and simply retrieving more will include more noisy images and more imbalanced distributions, which hurt the training performance.
We list the performance of the previous state-of-the-art few-shot recognition method CLAP~\cite{clap24} in the table for comparison with our SWAT.
}
\vspace{-3mm}
\label{tab:ablate_retr_size_dataset}
\scalebox{0.9}{
\begin{tabular}{crccccccccc!{\vrule}c}
\toprule
shots & Retrieval size & Semi-Aves & Flowers & Aircraft & EuroSAT & DTD & Pets & Food & Cars & ImageNet & average \\
\midrule
\multirow{6}{*}{4} & CLAP~\cite{clap24} & 34.0 & 90.1 & 28.0 & 74.7 & 63.0 & 87.0 & 76.7 & 84.9 & 64.0 & 66.9 \\

& 10 & 52.4 & \textbf{91.8} & 37.0 & \textbf{84.7} & \textbf{63.5} & 89.3 & 75.9 & \textbf{83.5} & 64.8 & 71.4 \\
 & 100 & 57.4 & 90.7 & 47.0 & 82.1 & \underline{62.1} & 89.9 & 76.6 & \textbf{83.5} & \textbf{66.1} & 72.8 \\
 & 300 & \textbf{58.7} & \underline{91.4} & 54.1 & 82.2 & 59.3 & 91.1 & \underline{77.1} & \underline{81.7} & \underline{65.9} & \underline{73.5} \\
 & 500 & \underline{58.5} & 90.6 & \underline{55.7} & 83.4 & 58.3 & \underline{91.3} & \textbf{77.3} & 81.1 & 65.8 & \textbf{73.6} \\
 & 1,000 & 58.3 & 89.6 & \textbf{58.1} & \underline{84.1} & 57.7 & \textbf{91.4} & 76.2 & 81.1 & 65.2 & \underline{73.5} \\

\midrule

\multirow{6}{*}{8} & CLAP~\cite{clap24} & 42.9 & 92.9 & 33.6 & 77.4 & 66.4 & 87.8 & 77.5 & 86.1 & 65.6 & 70.0 \\
& 10 & 55.7 & \textbf{95.2} & 42.2 & \textbf{90.0} & \textbf{69.1} & 89.4 & 76.9 & \textbf{86.8} & 65.8 & 74.6 \\
 & 100 & 59.2 & \underline{94.6} & 49.9 & 88.6 & \underline{65.2} & 90.2 & \underline{77.2} & \underline{85.3} & \underline{67.0} & 75.2 \\
 & 300 & 60.6 & 94.3 & 56.5 & \underline{89.3} & 63.1 & 90.9 & \textbf{77.6} & 83.9 & \textbf{67.3} & \underline{75.9} \\
 & 500 & \textbf{61.3} & 94.1 & \underline{59.1} & 88.7 & 62.6 & \textbf{91.5} & \textbf{77.6} & 83.5 & 66.6 & \textbf{76.1} \\
 & 1,000 & \underline{60.9} & 92.9 & \textbf{60.6} & 88.9 & 59.8 & \underline{91.4} & 76.7 & 83.6 & 66.2 & 75.7 \\

\midrule
 
\multirow{6}{*}{16} & CLAP~\cite{clap24} & 49.2 & 94.8 & 39.1 & 81.7 & 69.9 & 88.4 & 78.5 & 87.8 & 67.1 & 72.9 \\
& 10 & 58.4 & \textbf{97.0} & 48.6 & \textbf{94.0} & \textbf{72.9} & 89.6 & \underline{78.5} & \textbf{88.6} & 66.9 & 77.2 \\
 & 100 & 61.8 & \underline{96.8} & 54.5 & \underline{93.4} & \underline{69.4} & 90.2 & \textbf{78.6} & \underline{87.1} & \textbf{67.9} & 77.7 \\
 & 300 & \underline{63.2} & \underline{96.8} & 60.8 & 93.1 & 67.0 & 91.3 & \textbf{78.6} & 86.0 & \underline{67.8} & \textbf{78.3} \\
 & 500 & 63.1 & 96.4 & \underline{62.4} & 92.9 & 66.3 & \underline{91.6} & 78.3 & 85.4 & 67.6 & \underline{78.2} \\
 & 1,000 & \textbf{63.6} & 96.4 & \textbf{64.2} & 93.0 & 63.0 & \textbf{91.8} & 77.2 & 85.6 & 67.2 & 78.0 \\
\bottomrule
\end{tabular}
}
\label{tab:ablation-retr}
\end{table*}
}

\begin{figure}[t]
  \centering
  \includegraphics[width=0.47\textwidth, clip=true,trim = 0mm 0mm 0mm 0mm]{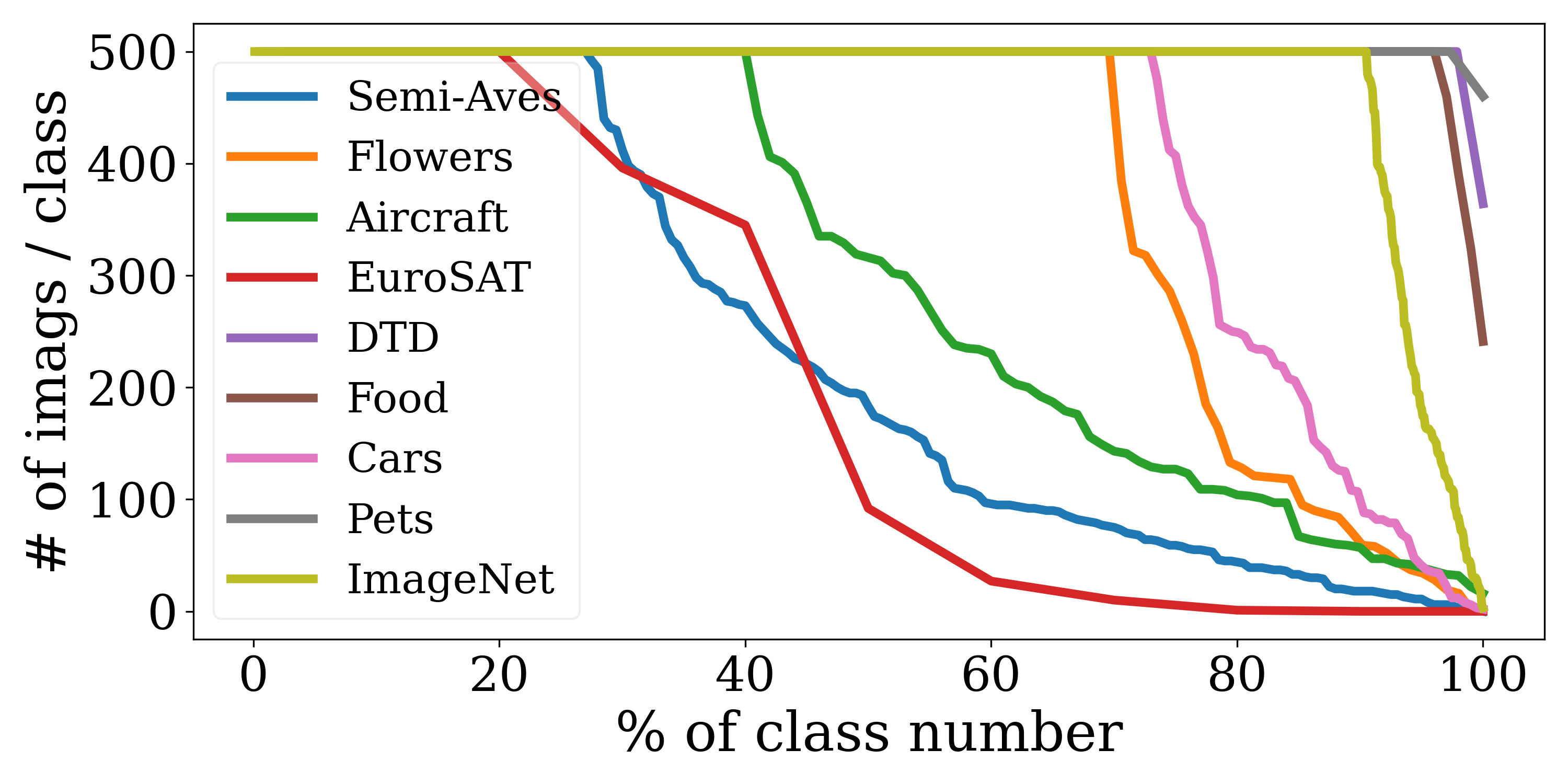}
  \caption{\small
  {\bf Retrieved data follows imbalanced distribution for all nine datasets.} 
   The retrieved data for ImageNet, Food, DTD, and Pets datasets are less imbalanced than other datasets, likely because the concepts from these datasets naturally appear more frequently on the Internet \cite{parashar2024neglected}.
  }
  \label{fig:imbalanced_all}
% \vspace{-2mm}
\end{figure}

\section{Analysis of Retrieved Data}
\label{sec:retrieved-stats}

\begin{figure*}[t]
    \centering
    \small
    % \vspace{-3mm}
    \includegraphics[width=0.97\textwidth, clip=true,trim = 5mm 0mm 5mm 0mm]{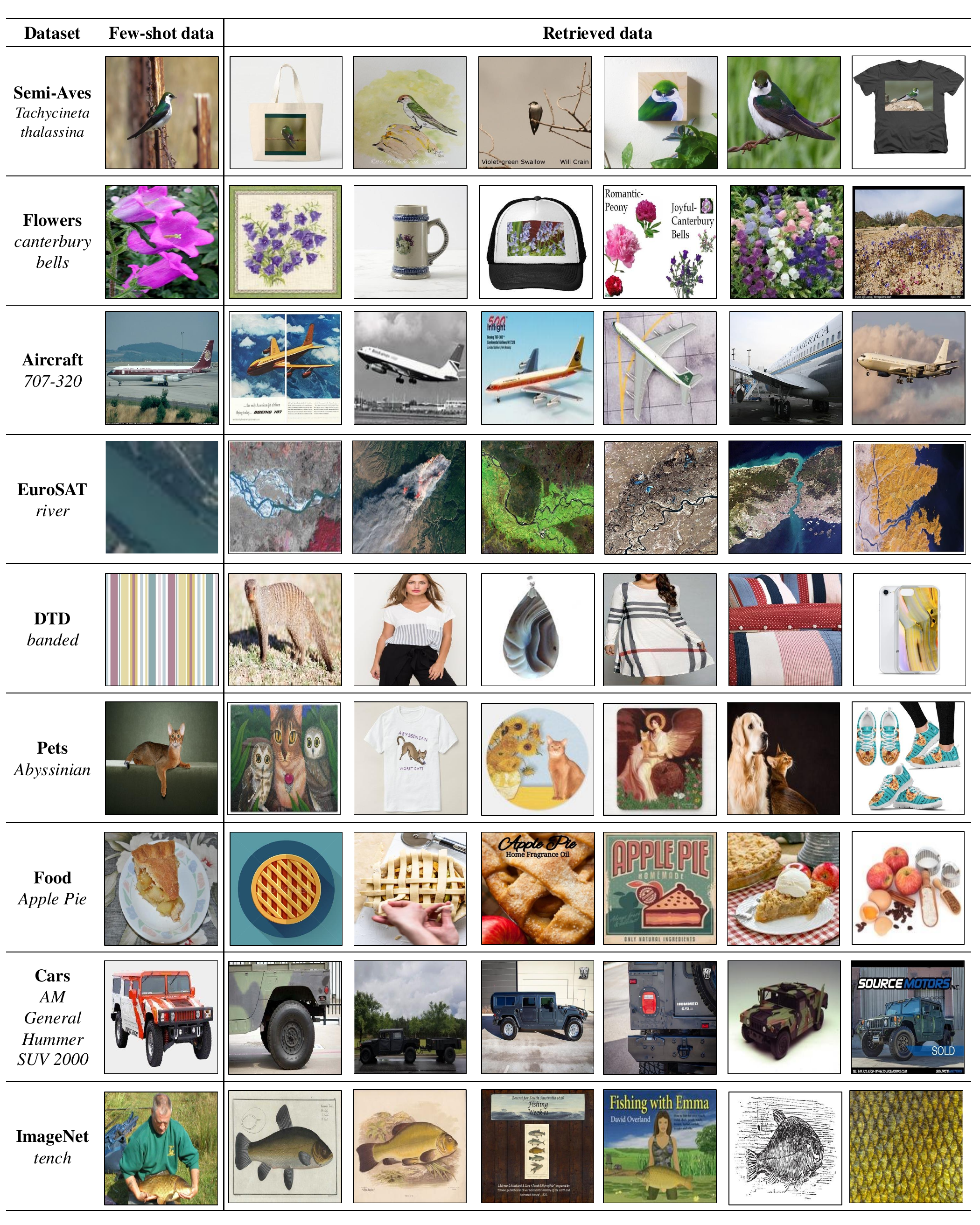}
    
    \vspace{-3mm}
    \caption{\small {\bf Comparison of downstream few-shot data with retrieved pretraining images (from LAION-400M~\cite{laion400m}) for nine fine-grained datasets}. 
    We present more examples of retrieved images for randomly selected classes from each dataset. 
    Compared to downstream few-shot images, the retrieved data exhibits diverse styles, backgrounds, resolutions, and even semantics, demonstrating significant domain gaps.  
    }
% \vspace{-3mm}
\label{fig:retrived_imgs_more}
\end{figure*}

{\bf Imbalances of Retrieved Data.}
We show the imbalanced distribution of retrieved data for all nine datasets in Fig.~\ref{fig:imbalanced_all}. 
We report the total number of retrieved images per dataset with increasing retrieval size (images per class) in Table~\ref{tab:total_retr}. 
With increasing retrieval size, the total number of retrieved images increases less significantly due to the limited presentation of many downstream concepts in the pretraining datasets (e.g. LAION~\cite{laion400m,laion5b}).
To address this issue, we suggest future work to retrieve relevant images from diverse data sources, e.g. other datasets or the Internet~\cite{li2023internet}.
Fig.~\ref{fig:retrived_imgs_more} shows more examples of retrieved images for each dataset.

\begin{figure*}[t]
    \centering
    \small
    \vspace{-3mm}
    \includegraphics[width=0.9\linewidth, clip=true,trim = 0mm 0mm 0mm 0mm]{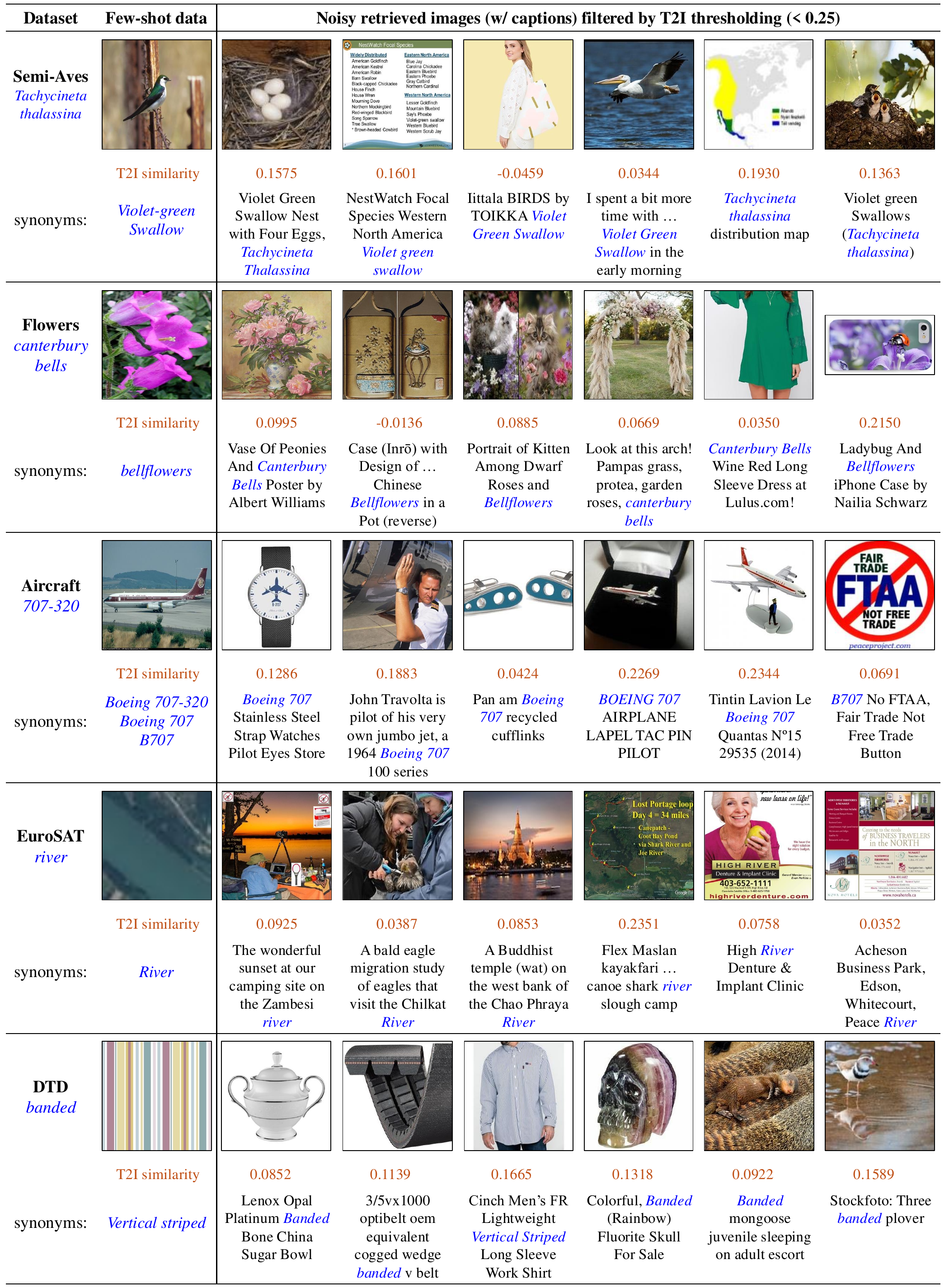}
    \vspace{-3mm}
    \caption{\small {\bf Examples of noisy retrieved images (from LAION-400M~\cite{laion400m}) filtered by T2I thresholding.}
    We show that string-matching-based retrieval (by searching image captions that contain any \textcolor{blue}{\textit{concept synonyms}}) can retrieve noisy images that could compromise the learning of downstream concepts, e.g., the bird eggs or the distribution map of bird species (first row).
    Using text-to-image (T2I) filtering helps remove such noisy images and improve the performance of SWAT (Table~\ref{tab:retrieve_methods}). We choose a T2I threshold of 0.25 for our experiment, similar to that used in the curation of LAION~\cite{laion400m, laion5b}.
    We highlight the \textcolor{brown}{T2I cosine similarity} and \textcolor{blue}{concept synonyms} for each image.
    }
% \vspace{-3mm}
\label{fig:filtered_imgs}
\end{figure*}

{\bf Impact of Retrieval Sizes.}
Additionally, we compare SWAT's performance on different retrieval sizes in Table~\ref{tab:ablate_retr_size_dataset}. Results show that SWAT saturates at 500 images per class for 4-shot and 8-shot cases and at 300 for 16-shot. Notably, for Flowers, EuroSAT, DTD, and Cars, retrieving only 10 images per class yields the best results, likely due to improved data balance and the exclusion of noisy images (Fig.~\ref{fig:filtered_imgs}). Future work can study how to enhance the balance and quality of retrieved data.

{
\setlength{\tabcolsep}{0.85em} 
\begin{table*}[t]
\small
\centering
\caption{\small {\bf Total number of retrieved images for each dataset under different retrieval sizes.} 
With a larger retrieval size (number of retrieved images per class), we observe a diminished increase in the total number of retrieved images. This is because many downstream concepts have limited presence in the pretraining set (LAION-400M~\cite{laion400m, laion5b}). See the imbalanced distribution of each dataset in Fig.~\ref{fig:imbalanced_all}.
}
\vspace{-2mm}
\label{tab:total_retr}
\begin{tabular}{rrrrrrrrrrrr}
\toprule
images / class & Semi-Aves & Flowers & Aircraft & EuroSAT & DTD & Pets & Food & Cars & ImageNet \\
\midrule
10 & 1,940 & 1,002 & 1,000 & 71 & 470 & 370 & 1,010 & 1,939 & 9,989 \\
100 & 15,687 & 9,376 & 9,120 & 530 & 4,700 & 3,700 & 10,100 & 18,494 & 98,753 \\
300 & 34,685 & 25,140 & 21,774 & 1,330 & 14,100 & 11,100 & 30,241 & 51,251 & 288,532 \\
500 & 47,006 & 39,465 & 30,429 & 1,871 & 23,364 & 18,460 & 49,914 & 80,648 & 471,876 \\
1,000 & 67,418 & 71,332 & 44,519 & 2,387 & 45,978 & 36,105 & 96,697 & 147,568 & 901,902 \\
\bottomrule
\end{tabular}
\end{table*}
}

\section{Code and Instructions}
\label{sec:Demo-code}

We release open-source Python code at \url{https://github.com/tian1327/SWAT}.

{\bf Requirements}.
Running our code requires some common packages.
We installed Python and most packages through Anaconda. A few other packages might not be installed automatically, such as clip, open\_clip\_torch, img2dataset, torchvision, and PyTorch, which are required to run our code. We provide detailed instructions for building the environment in file {\tt ENV.md}. Below are the versions of Python and PyTorch used in our work. 
\begin{itemize}
\item Python version: 3.8.19
\item PyTorch version: 2.0.1
\end{itemize}
We suggest assigning $>$50GB storage space and $>$5GB GPU RAM to reproduce our experiments.

{\bf License}.
We release open-source code under the MIT License to foster future research in this field.

{\bf Instructions.}
We provided detailed step-by-step instructions for running our code in the following markdown files.

\begin{itemize}
\item
\begin{verbatim}ENV.md\end{verbatim}
Create a conda environment and install the required packages.

\item
\begin{verbatim}DATASETS.md\end{verbatim}
We provide detailed steps for setting up the benchmarking datasets and sampling few-shot data from the official training sets with three random seeds.

\item
\begin{verbatim}RETRIEVAL.md\end{verbatim}
We provide step-by-step instructions on how to use string-matching~\cite{parashar2024neglected} to retrieve relevant images from OpenCLIP's pretraining dataset LAION-400M~\cite{laion400m, laion5b}. Examples of different ranking and filtering methods for selecting the images that are most relevant to downstream concepts are also provided.

\item
\begin{verbatim}README.md\end{verbatim}
We provide instructions on how to run the provided code for few-shot finetuning (FSFT) and SWAT. In addition, we provide guidelines on how to reproduce the baseline methods Cross-Modal Linear Probing~\cite{lin2023multimodality} and CLAP~\cite{clap24}.

\end{itemize}

\end{document}